%% file: main.tex
\documentclass[%
  5p,
  sort&compress,
  number.
  final,
]{elsarticle} 

\biboptions{longnamesfirst,semicolon}

\usepackage{lineno,hyperref}
\modulolinenumbers[5]
\usepackage[utf8]{inputenc}
\usepackage{graphicx}
\usepackage{subcaption}
\def\github{https://github.com/marccoru/crop-type-mapping}

\input{preamble.tex}

\journal{Journal of Photogrammetry and Remote Sensing}

\begin{document}

\begin{frontmatter}

\title{Self-attention for raw optical Satellite Time Series Classification}

\author{Marc Rußwurm and Marco Körner}
\address{Chair of Remote Sensing Technology,\\ Department of Aerospace and Geodesy,\\ Technical University of Munich \\ Arcisstraße 21, 80333 Munich, Germany}

\begin{abstract}
	{
		The amount of available Earth observation data has increased dramatically in recent years.
		Efficiently making use of the entire body of information is a current challenge in remote sensing; it demands lightweight problem-agnostic models that do not require region- or problem-specific expert knowledge.
		End-to-end trained deep learning models can make use of raw sensory data by learning feature extraction and classification in one step, solely from data.
		Still, many methods proposed in remote sensing research require implicit feature extraction through data preprocessing or explicit design of features.
		
		In this work, we compare recent deep learning models on crop type classification on raw and preprocessed Sentinel 2 data.
		{We concentrate on the common neural network architectures for time series, \ie 1D-convolutions, recurrence, and the novel self-attention architecture.}
		Our central findings are that data preprocessing still increased the overall classification performance for all models while the choice of model was less crucial.
		Self-attention and recurrent neural networks, by their architecture, outperformed convolutional neural networks on raw satellite time series.
		We explore this by a feature importance analysis based on gradient backpropagation that exploits the differentiable nature of deep learning models.
		Further, we qualitatively show how self-attention scores focus selectively on a few classification-relevant observations. 
	}
\end{abstract}


\end{frontmatter}


\section{Introduction}

{Today, medium resolution Earth observation data is often provided free-of-charge; this has been the case ever since the Landsat program, jointly operated by the United States Geological Survey (USGS) and the National Aeronautics and Space Administration (NASA), set a precedent by adopting an open data policy in 2008 \cite{woodcock2008free,wulder2019current}. 
In recent years, the Copernicus program of the European Space Agency (ESA) joined this initiative with a growing family of Sentinel satellites that capture data from the Earth's surface at increasing temporal intervals.

} For instance, the Sentinel 2 multispectral satellites acquire data at up to \SI{10}{\meter} resolution in \num{13} spectral bands every two to five days.
{
Acquiring surface reflectance data at a high temporal frequency is crucial for vegetation-related remote sensing applications, such as crop type mapping \cite{Reed1994,odenweller1984} or forest monitoring~\cite{wulder2004national,herold2019role,reiche2016combining}.
This data abundance, however, leads to computational and organizational challenges and demands methods that are problem- and region-agnostic enough to utilize all available temporal data on large spatial scales.
That these \emph{challenges of Big Geo Data} are not solved yet can be seen in the low exploitation ratio of Sentinel 2 data in 2018.  
}
A {large body} of \SI{7.76}{\tebi\byte}\footnote{$\SI{1}{\tebi\byte}=2^{40}\si{\byte}$} Sentinel 2 data was published on a daily average.
Still, only \SI{7.6}{\percent} of all published images in 2018%
\footnote{62,317,329 published, 4,737,253 downloaded Sentinel 2 products in 2018} 
have actually been downloaded~\citep{Copernicus2019}.
Accordingly, \num{12} out of \num{13} published images remain unused.
Similar figures can be drawn for the Sentinel 1, 3, and 5 missions.
This is a clear sign that, despite claims made by academia and industry, methods and principles of \emph{big data analytics} are rarely applied to their full potential in the field of Earth observation.
The reasons for this low exploitation ratio are manifold.
For instance, visual inspection of satellite images is still often the first step of data acquisition.
Further, intermediate results are often required to be visually interpretable to allow for review by domain experts.
Another reason is that preprocessing steps, like atmospheric correction or manual or automatic cloud filtering, are {common} in remote sensing. 
In principle, deep learning mechanisms seem well-suited to approximating preprocessing-like mechanisms by jointly learning feature extraction and classification within one neural network topology using gradient backpropagation.
Developing models that do not strictly require extensive data preprocessing is likely to be a key contribution to utilizing available published satellite data efficiently in the future.
Within the scope of this objective, we assess three deep learning mechanisms on four deep learning models on selectively available preprocessed and readily available raw satellite data.

Summarizing, the contribution of this work is twofold:
\begin{enumerate}
	\item We first present a large-scale evaluation of deep learning models on preprocessed and raw satellite data using three mechanisms---\ie convolution, recurrence, and self-attention---for crop type identification.
	\item We further provide quantitative and qualitative analysis of self-attention in the context of commonly used deep learning models for satellite time series classification. 
\end{enumerate}

\section{Related Work}

Since satellite data is inherently multitemporal, this temporal dimension has been a topical focus of vegetation-related applications for a long time~\citep{odenweller1984, Reed1994}.
The limited availability of multitemporal satellite data and the excessive computational demand for temporal stacks of large-scale satellite images have impeded broad exploitation of the temporal dimension.
Methods have relied on separate feature extraction, \ie calculation of vegetation indices~\cite{Foerster2012, conrad2010, conrad2014, Hao15, Barragan2011} with temporal statistics~\cite{Foerster2012,Hao15,zhong2019deep}, and classification mechanisms, \eg with Random Forest Classifiers~\cite{Hao15,Valero2016,conrad2014} or Support Vector Machines~\cite{Kumar15,zheng2015support,devadas2012support}, as summarized by~\cite{Uensalan11:RLU,shao2012comparison}.
Similarly, {the TimeSat software}~\cite{jonsson2004timesat, eklundh2016timesat} fits non-symmetric Gaussian curves to satellite time series. 
The parameters of these curves, \ie the steepest ascent and descent, and their times of occurrence are assumed to coincide with key phenological characteristics, such as the onset of greenness or the date of senescence.
This allows for detailed phenological analyses~\cite{white2009intercomparison, olsson2005recent} and can be used as a distinctive feature for further classification~\cite{jia2014land, singha2016object}.
{
The \emph{Continuous Change Detection and Classification (CCDC)} \cite{zhu2014continuous} algorithm first identifies \enquote{clear} observations by prior cloud filtering with FMask \cite{zhu2012object, zhu2015improvement} on atmospherically corrected images. The FMask \cite{zhu2012object} algorithm itself is a functionally designed decision tree that exploits the physical properties of Landsat bands to identify cloud coverage and was later extended to more satellites \cite{zhu2015improvement}. The CCDC algorithm then explicitly models inter- and intra-annual seasonality by a sum of periodic functions for each pixel fitted with \emph{Robust Iteratively Reweighted Least Squares (RIRLS)} \cite{street1988note,dumouchel1989integrating} to training data. The parameters of the periodic functions can be used as features for a subsequent classification. 
Similarly, the periodic functions can be used to predict future reluctance values and compared to measured data for change detection.

The \emph{Breaks for Additive Seasonal and Trend (BFAST)} algorithm~\cite{verbesselt2010detecting} similarly decomposes a satellite time series into piecewise seasonal and linear components where the remainder can be used to detect anomalies and the periodicity can be used as a feature for classification.
The LandTrendr \cite{kennedy2010detecting} aims at removing high-frequency components from a Landsat time series by finding a combination of successively simpler linear models, defined by vertices, that still explain the data sufficiently well. A variety of parameters can be adjusted to tune the model to specific regions and data through the number of allowed spikes (robustness to clouds) or the number of segments (model complexity).
These methods have been designed to require data preprocessing, \ie atmospheric correction of cloud filtering, and some degree of direct supervision, \ie through the choice of parameters, or they are tailored towards specific types of data, \eg Landsat for CCDC or MODIS for BFAST.
}

In contrast, artificial neural networks~\cite{mcculloch1943logical, cowan1990neural} aim at learning feature extraction and classification with a single dynamic model solely from the provided data with minimal supervision beyond the provided data. 
First, success was achieved for the problem of recognition of handwritten digits~\cite{lecun1998gradient} using multilayer neural networks trained with gradient descent.
However, comparatively recent advances in parallel processing and data availability~\cite{deng2009imagenet, russakovsky2015imagenet} were necessary to outperform common feature extraction and classification pipelines in computer vision~\cite{simonyan2014vgg,krizhevsky2012imagenet,he2016deep,long2015fully, szegedy2017inception} and natural language processing~\cite{vaswani2017attention,devlin2018bert,radford2019language} on a large scale.
Following these developments, monotemporal Earth observation approaches have adapted 2D convolutional neural networks from the domain of computer vision with great success~\cite{marmanis2015deep,volpi2016dense,sherrah2016fully,audebert2016semantic}.
{
	Rotation-invariant feature extraction \cite{cheng2016} for scene classification \cite{cheng2017} and metric learning \cite{cheng2018} has been similarly tested for remote sensing image classification.
}
Methods from multitemporal Earth observation~\cite{russwurm2017temporal,Jia2017b,Lyu2016,Lichao2018,benedetti2018m3fusion,interdonato2019duplo} adapted models from natural language processing. 
Here, recurrent neural networks~\cite{Werbos1990}, such as Long Short-Term Memory (LSTM)~\cite{Hochreiter97:LST} or Gated Recurrent Units (GRU)~\cite{chung2014empirical}, were commonly used in encode-decoder architectures\cite{sutskever2014sequence} for generative prediction of words.
In Earth observation, the encoder model was utilized for change detection~\cite{Lyu2016,Lichao2018} and land cover~\cite{jia2014land}, as well as for crop type identification~\cite{russwurm2017temporal,interdonato2019duplo,garnot2019time,sharma2018land}.
To utilize both spatial and temporal features from time series, combinations of convolutional layers and recurrent layers~\cite{benedetti2018m3fusion, Lichao2018} or convolutional-recurrent networks~\cite{russwurm2018multi} have been explored and comprehensively compared~\cite{garnot2019time}.
These recurrent neural network encoders can be augmented by soft-attention mechanisms, originally developed for machine translation~\cite{bahdanau2016end}, as tested in~\cite{interdonato2019duplo}.
Recently, lightweight time-convolutional feed-forward neural networks have shown to be a powerful alternative~\cite{fawaz2019deep} and have achieved promising results in crop type identification~\cite{pelletier2019temporal}.
While the state-of-the-art in remote sensing remains focused on convolutional and recurrent architectures, \emph{self-attention}~\cite{vaswani2017attention}, in combination with pre-training~\cite{devlin2018bert,radford2019language}, has started dominating the state-of-the-art in natural language processing.
In the remote sensing context, self-attention has {barely been explored in the canon of existing work \citep{garnot2019satellite}} on comparative analysis of recurrence and convolutional mechanisms.

\section{Method}
\label{sec:method}

In this section, we introduce the notation used throughout this work, provide background on convolution and recurrence, and address self-attention before employing these mechanisms in four neural network topologies in \cref{sec:models}.

A deep learning model $f_\Mweight: \Set{X} \mapsto \Set{Y}$ approximates a mapping from an input domain $\Set{X}$ to a target domain $\Set{Y}$. 
These deep models are implemented by cascaded neural network layers that form a problem-agnostic, nonlinear differentiable function.
The parameters $\Mweight$ are determined by 
minimizing
an objective function $\mathcal{L}(\V{y},f_{\Mweight}(\M{X}))$ that quantifies the dissimilarity of ground truth labels $\V{y} \in \Set{Y}$ and the model predictions $\yhat = f_{\Mweight}(\M{X})$.
This objective function is minimized iteratively via mini-batch stochastic gradient descent 
\begin{align}
\Mweight_{t+1} &\leftarrow \Mweight_{t} - \lr \V{s}_t, & 
\text{ with } \V{s}_t &= \frac{1}{N} \sum_{i=1}^N \frac{\partial}{\partial\Mweight} \mathcal{L}(\V{y}_i, f_{\Mweight_t}(\V{X}_i)) \quad.
\label{eq:gradientdescent}
\end{align}
The gradients  $\frac{\partial}{\partial\Mweight} \mathcal{L}$ are averaged over a batch of size $N$, while the learning rate $\lr$ determines the step size.
This optimization scheme guarantees that the chosen parameters $\Mweight$ are optimal for the observed dataset and objective function.

Deep learning for time series classification aims at learning the mapping $\yhat = f_\Mweight(\M{X})$ from an input time series $\M{X} \in \Set{X}^T = \R^{T \times D} = (\V{x}_0,\V{x}_1,\dots,\V{x}_T)$ of individual measurements $\V{x}_t \in \R^D$ of $D$ features to one of $C$ classes.
These classes are represented by the \emph{one-hot} target vector $\V{y} \in \{0,1\}^{C}$ where $y_i \in \{0,1\}$ indicates the boolean class membership.
We denote the approximation of the target vector as $\yhat \in [0,1]^C$.
A neural network is a nonlinear and differentiable function 
$f_{\Mweight}=f^{(1)}_{\Mweight_1} \circ f^{(2)}_{\Mweight_2} \circ \dots  \circ f^{(L)}_{\Mweight_L}$ 
of $L$ cascaded layers $f^{(l)}_{\Mweight_l}$.
Each layer $f^{(l)}_{\Mweight_l}: \M{H}^{(l)} \xmapsto{\Mweight_l} \M{H}^{(l+1)}$ consists of a linear transformation 
$\Mweight_l$ mapping an input representation $\M{H}^{l} \in \R^{T \times D_h}$ to corresponding hidden features $\M{H}^{l+1} \in \R^{T \times D_h}$ that encode increasingly higher-level features. 
In time series classification, one of the following mechanisms is used to implement this nonlinear mapping.

\subsection{Fully Connected Layers}

A dense or fully connected layer $\V{h}_t = f^\text{fc}_{\Mweight}(\V{x}_t) \in \R^{D_h}$ is applied independently at time instance $t$ of a series. 
Each vector $\V{x}_t \in \R^D$ is transformed into \\
$\V{h}_t =  f^\text{fc}_{\{\Mweight_\text{l},\Mweight_\text{b}\}}(\V{x}_t) = \phi\left(\Mweight_\text{l}\T \V{x}_t + \Vweight_\text{b}\right)$ 
by a linear transformation $\Mweight_\text{l} \in \R^{D \times D_h}$ with optional bias translation~
$\Vweight_\text{b} \in \R^{D_h}$ followed by a nonlinear activation function \\ $\phi\left(\cdot\right) \in \{\tanh,\sigma,\text{ReLU},\ldots\}$.

\subsection{Recurrent Layers}

Recurrent neural networks \citep{Rumelhart1986} extend fully connected layers by contextual information from previous time steps 
$t-1$
This results in a transformation \\ $\V{h}_t=f^\text{rnn}_{\Mweight}(\V{x}_t, \V{h}_{t-1}) = \phi\left(\Mweight_x\T \V{x}_t + \Mweight_h\T \V{h}_{t-1} + \Vweight_\text{b}\right)$.
This formulation is prone to \emph{vanishing} and \emph{exploding} gradients through time \citep{Hochreiter97:LST, jozefowicz2015empirical, Werbos1990, Bengio1994} that inhibit the extraction of features from long-term temporal contexts.
While the effect of exploding gradients could be controlled through gradient clipping, vanishing gradients have been addressed by the introduction of additional gates.
This led to \emph{LSTM} recurrent networks \citep{Hochreiter97:LST} that introduced four internal gates $\V{f}_t = f^\text{rnn}_{\Mweight_f}(\V{x}_t, \V{h}_{t-1})$, $\V{i}_t=f^\text{rnn}_{\Mweight_i}(\V{x}_t, \V{h}_{t-1})$, $\V{g}_t=f^\text{rnn}_{\Mweight_g}(\V{x}_t, \V{h}_{t-1})$, $\V{o}_t=f^\text{rnn}_{\Mweight_o}(\V{x}_t, \V{h}_{t-1})$.
An additional \emph{cell state} matrix $\V{c}_t$ was introduced as a container for the long-time temporal context.
In each iteration, this cell state $\V{c}_t = \V{f}_t\odot\V{c}_{t-1} + \V{i}_t \odot \V{g}_t$ is updated by element-wise multiplications $\odot$ with the values of 
the forget, input, and modulation gates $\V{f}_t$, $\V{i}_t$, and $\V{g}_t$, respectively.
{If the forget gate evaluates values close to zero, features in the prior cell state $\V{c}_{t-1}$ will be suppressed (forgotten). The modulation and input gates control the writing of new features from the input to the cell state $\V{c}_t$.}
The cell output $\V{h}_t = \V{o}_t \odot \tanh(\V{c}_t)$ is calculated using the result from the output gate $\V{o}_t$ and the new cell state $\V {c}_t$.
Overall, these transformations yield the LSTM update $f^\text{lstm}_{\Mweight}: \V{h}_t, \V{c}_t \leftarrow \V{x}_t, \V{h}_{t-1}, \V{c}_{t-1}$.
An alternative to LSTMs are \emph{GRUs}~\cite{chung2014empirical}, which follow the same principle of additional gates but are parameterized with fewer gates and weights. 
Despite computational benefits, empirical evaluations did not show a significant difference in performance between these parameterizations~\cite{jozefowicz2015empirical}.

\subsection{1D Convolutional Layers}

Convolutional layers extract features by correlating the input signal $\M{X} \in \R^{T \times D}$ with a set of $D_\text{h}$ convolutional kernels $\Mweight = (\Mweight_0,\Mweight_1,\dots,\Mweight_{D_\text{h}})$ with $\Mweight_d \in \R^{K \times D}$ employing a convolutional operation $\M{H} = \phi(\M{X}\ast\Mweight)$ followed by an element-wise nonlinear activation function $\phi$. 
The size of the convolutional kernel, $K$, determines the receptive field of each layer.
In contrast to recurrent layers, convolutional layers extract features from a fixed temporal neighborhood.
The receptive field increases through the number of layers. 
Convolutional layers perform well in the field of computer vision where features from the local neighborhood of an image are extracted with small local kernels. 
In a temporal context, larger kernel sizes are used.
There, the learned kernel resembles a correlation between learned patterns in the time domain \citep{IsmailFawaz2019inceptionTime, hatami2018classification} and recent applications in the field of remote sensing \citep{pelletier2019temporal}. 

\subsection{Self-Attention Layers}
\label{sec:self-attention}

\begin{figure}
	\centering
	
	\subcaptionbox
	{matrix multiplications in the calculation of attention scores $\M{A}$}
	[.45\textwidth][t]
	{\centering\input{images/attentionmatrices.tikz}}%
	\hfill
	\subcaptionbox
	{relationship between values and outputs as bipartite graph}
	[.45\textwidth][t]
	{\centering\input{images/attention_graph.tikz}}%
	
	\caption{%
		Schematic self-attention operation using $\Tin = \Tout = 4$, $\Din = \Dout=2$ and $\Dh = 4$. 
		The hidden output $\M{H} \in \R^{4 \times 2}$ results from matrix multiplication with attention scores $\M{A} \in [0,1]^{4 \times 4}$ with a time series $\M{V} \in R^{4 \times 2}$. 
		These attention scores themselves are calculated by a matrix multiplication between key $\M{K} \in \R^{4\times4}$ and query matrices $\M{Q} \in \R^{4\times4}$. 
		In self-attention, all key, query, and value matrices originate from an input time series $\M{X}$ transformed via a linear mapping $\M{X}\Mweight$.}
	\label{fig:attention}
\end{figure}

Attention mechanisms \citep{bahdanau2016end} allow a neural network to extract features from observations at specific time steps of an input time series of \emph{values} $\M{V}=(\V{v}_0,\V{v}_1,\dots,\V{v}_\Tin) \in \R^{\Tin \times \Din}$.
This is realized by a weighted sum 
$\V{h}_i = \sum_{t=0}^{T_\text{in}} \alpha_t \V{v}_t = \V{\alpha}\T\M{V}$ 
of attention scores $\V{\alpha} \in [0,1]^\Tin, \sum_i \alpha_i = 1$.
Extending this to a matrix multiplication 
\begin{align}
\M{H} = \M{A}\T\M{V}
\end{align}
with attention matrix $\M{A} = (\V{a}_0,\dots,\V{a}_\Tout)$ and \\ $\M{H} = (\V{h}_0,\dots,\V{h}_\Tout) \in \R^{\Tout \times \Dout }$ generates $\Tout$ output vectors in parallel.
To illustrate these operations and to prepare for numerical results later in \cref{sec:exp:qualitative:self-attention}, we show a visual representation of these operations in \cref{fig:attention}. 
Attention scores themselves are the result of a second matrix multiplication $\M{A}=\M{Q}\M{K}\T$ in \cref{eq:attention} of 
\emph{key} and \emph{query} matrices $\M{K} \in \R^{\Tin\times\Dh}$ and $\M{Q} \in \R^{\Tout \times \Dh}$, respectively.
This results in the generic formulation of attention 
\begin{align}
\M{H} = \underbrace{\text{softmax}\left({\M{Q}\M{K}\T}\right)\T}_{\M{A}\T} \M{V}
\label{eq:attention}
\end{align}
that involves a softmax function to normalize $\M{A}$. 
In self-attention \citep{vaswani2017attention} 
\begin{align}
\label{eq:attentioncomplete}
\M{H}=f_{\{\Mweight_K,\Mweight_Q,\Mweight_V\}}(\M{X}) = 
\text{softmax}\left(
\frac{ { \Mweight_Q\T \M{X} } { \Mweight_K\T \M{X}\T} }
{\sqrt{\Dh}}\right) 
{\Mweight_V\T \M{X}},
\end{align}
this formulation is adopted with unified dimensions $\Din=\Dout=\Dh$ and $\Tin=\Tout$ where the keys $\M{K} = \Mweight_K\T \M{X}$, queries $\M{Q} = \Mweight_Q\T \M{X}$, and values $\M{V} = \Mweight_V\T \M{X}$ matrices originate from the same input matrix $\M{X}$ transformed by three linear transformations.
Following the original work \cite{vaswani2017attention}, the attention scores are scaled by $\sqrt{ \Dh }$ for better gradient backpropagation.
Today, such self-attention mechanisms are the core of state-of-the-art natural language processing models \citep{devlin2018bert, radford2019language} and will be experimentally analyzed later in \cref{sec:exp:qualitative:self-attention,sec:exp:qualitative:tsne}. 
{
	\subsection{Soft-Attention}
	\label{sec:soft-attention}
	
	Soft-attention~\citep{britz2017efficient}, as implemented in the DuPLO model \cite{interdonato2019duplo}, is a special case of \cref{eq:attention} where values $\M{V}=\M{X}$ and queries $\M{Q} = \tan\left(\M{\Theta}_\alpha\M{X}\right)$ are obtained from a linear transformation from the input tensor $\M{X}$. 
	In contrast to self-attention, the keys $\M{K}=\M{\Theta}_K$ are a fixed weight matrix that is learned by gradient descent. 
	Following the definitions of \cite{interdonato2019duplo} and modified from \cref{eq:attentioncomplete}, soft-attention yields 
	\begin{align}
	\label{eq:softattention}
	\V{h}=f_{\{\Mweight_K,\Mweight_A\}}(\M{X}) = 
	\underbrace{\text{softmax}\left(\tan\left(\Mweight_A\T \M{X}\right) \Mweight_K\right)\T}_{\V{\alpha}\T} 
	\M{X}
	\end{align}
	where a nonlinearity $\phi$ is implemented as tangents $\tan$ without scaling factor $D_h$.
	We compare self-attention with simplified soft-attention in a dedicated experiment in \cref{sec:attentioncomparison}.
	
}

\section{Models}
\label{sec:models}

The previous section provided an overview of neural network layers used for temporal feature extraction and introduced temporal convolution, recurrence, and self-attention.
In this section, we describe four neural network topologies, each of them using one of these layer mechanisms, which will be evaluated experimentally in the following\footnote{The models are available at \url{\github/tree/master/src/models}}.

\subsection{Recurrent Long Short-Term Memory Neural Network}

\begin{figure*}
	\begin{subfigure}[t]{.245\textwidth}
		\centering\input{images/models/lstm.tikz}
		\caption{\rnn (recurrence)}
		\label{fig:models:rnn}
	\end{subfigure}
	\hfill
	\begin{subfigure}[t]{.22\textwidth}
		\centering\input{images/models/transformer.tikz}
		\caption{Transformer (self-attention)}
		\label{fig:models:transformer}
	\end{subfigure}
	\hfill
	\begin{subfigure}[t]{.29\textwidth}
		\centering\input{images/models/msresnet.tikz}
		\caption{MS-ResNet (convolution)}
		\label{fig:models:msrenset}
	\end{subfigure}
	\hfill
	\begin{subfigure}[t]{.2\textwidth}
		\centering\input{images/models/tempcnn.tikz}
		\caption{TempCNN (convolution)}
		\label{fig:models:tempcnn}
	\end{subfigure}
	\caption{A schematic overview of the four network topologies used in this study.}
	\label{fig:models}
\end{figure*}
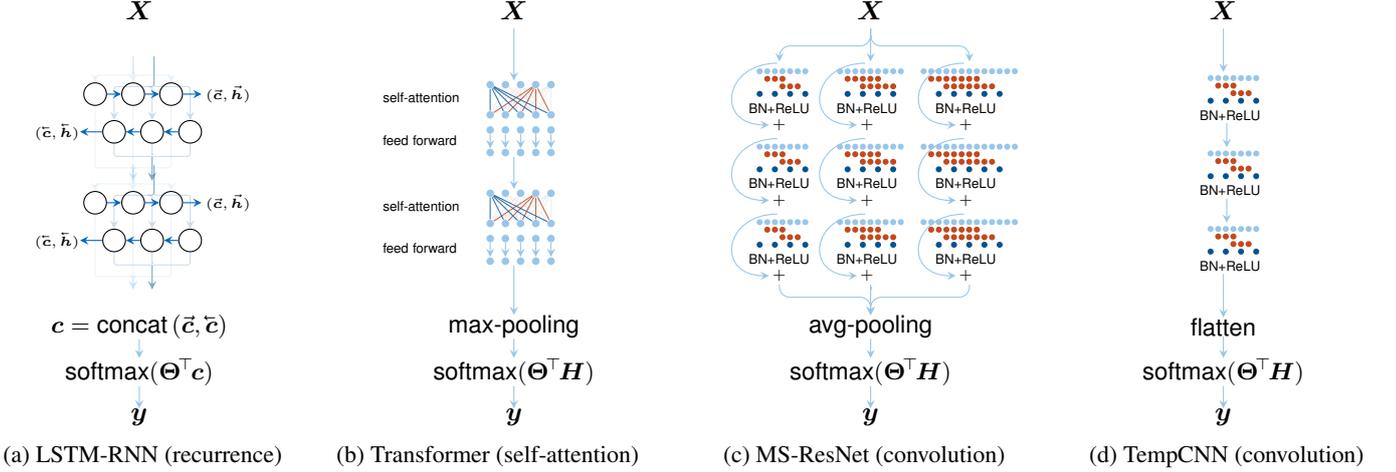

We designed a multilayer \emph{Recurrent Neural Network (RNN)} \citep{Rumelhart1986} with \emph{LSTM}\citep{Hochreiter97:LST} cell architecture, similar to our previous work \citep{russwurm2017temporal}. 
This network encodes a satellite time series to increasingly higher-level $D$-dimensional representations through $N$ cascaded bidirectional LSTM layers. 
The hidden states are initialized with zero-valued vectors.
The last hidden states from forward and backward passes, $\vec{\V{c}}$ and $\cev{\V{c}}$, are concatenated and produce final classification scores via a dense output layer with softmax activation function.
This network architecture is illustrated in \cref{fig:models:rnn}.
We tuned the conditionality of hidden states $\hiddendims \in \SetDef{2^4,2^5,\dots,2^8}$, the number of recurrent layers $\numlayers \in \SetDef{1,2,\dots,7}$, and applied dropout $\dropout \in [0,1]$ between the recurrent layers at training time.

\subsection{Transformer}

For the self-attention model, we adopted the encoder architecture of the self-attention \transformer network~\cite{vaswani2017attention}, as illustrated in \cref{fig:models:transformer}.
We discarded the step of word embedding, as the satellite time series data lives in a continuous space of spectral reflectance values. 
Following Vaswani \etal \citep{vaswani2017attention}, we added a positional encoding to the time series, since self-attention cannot utilize the sequential correlation of time series.
Subsequently, the time series with positional encoding is transformed into higher-level $D$-dimensional feature representations through $L$ \transformer blocks.
Each block encodes features through a multihead self-attention mechanism followed by multiple dense layers applied to each time instance independently.
Skip connections are introduced between the layers, and layer normalization is used throughout the model. 
We refer to \cite{vaswani2017attention} for a detailed description of the layer topology.
The representation of the last layer $\M{H} \in \R^{H \times T}$ is then reduced to $\V{h} \in \R^{D}$ by a global maximum pooling through the time dimension. 
This reduced representation is then projected to scores for each class by a final fully connected layer with a softmax activation function. 

For this architecture, we tuned the dimensionality of the hidden states $\hiddendims \in \{2^4,2^5,\dots, 2^8\}$, the number of self-attention layers $\numlayers \in \{1,2, \dots,8\}$, and the number of self-attention heads $\numheads \in \{1,2,\dots,8\}$ that determine the number of self-attention mechanisms applied in parallel.

\subsection{Multiscale Residual Networks}

\emph{Residual Networks (ResNets)} \citep{he2016deep} are a common backbone architecture in computer vision.
They consist of residual blocks $f^{\text{cbn}}_\Mweight = f^\text{c}_\Mweight \circ f^\text{b} \circ f^\text{r}$ of 
convolution layers $f_\Mweight^\text{c}$ and batch normalization layers $f^\text{b}$ \cite{ioffe2015batch},  
followed by a \emph{Rectified Linear Unit (ReLU)} activation function $f^\text{r}$.
Characteristically, in each block $\M{H} = f^\text{cbn}_\Mweight(\M{X}) + \M{X}$, the encoded feature representation $f^{\text{cbn}}_\Mweight(\M{X})$ is added to the identity mapping of the input. 
This forms residual skip connections where each residual layer numerically adds higher-level features to the forward propagated representation. 
These skip connections aid gradient backpropagation through the network and allow the training of very deep models.

These convolutional residual connections have been {adapted} to time series classification~\citep{wang2017time} where 2D convolutions through the spatial dimensions are replaced by 1D convolutions through time.
Since convolutional layers, by design, extract features from the local (temporal) neighborhood, architectures that process time series at multiple scales \cite{cui2016multi} have shown good results on time series benchmark datasets \cite{dau2018ucr,uea}.
In this work, we utilize the architecture proposed by Wang \etal \citep{wang2017time}\footnote{Original implementation available at \url{https://github.com/geekfeiw/Multi-Scale-1D-ResNet}} where an input time series $\M{X} \in \R^{(T=512) \times D_\text{in}}$ is processed in three separate streams, each of which has three residual blocks with increasing convolutional kernel sizes $K^\text{c} \in \{3,5,7\}$.
Hence, each stream extracts features at a different temporal scale, as drawn in \cref{fig:models:msrenset}.
The representation after each stream is then average-pooled with pooling kernel sizes $K^\text{p} \in \{16,11,6\}$, yielding three feature representations
$\V{h}^{K^c=3} , \V{h}^{K^{c}=5} , \V{h}^{K^{c}=7} \in \R^{D=256}$,
and concatenated to a common representation $\M{h} \in \R^{D=768}$. 
A final fully connected layer with softmax activation function reduces this vector to individual activation scores per class.
Note that the design of this network requires a fixed sequence length of $T=512$. 
To obtain this sequence length, we interpolated the satellite time series by the nearest neighbor method.
Since the architecture is precisely defined in the original implementation, we solely tune the number of convolutional kernels $\hiddendims \in \{2^4,2^5,\dots, 2^8\}$ for this architecture.

\subsection{TempCNN}

Recently, a Temporal Convolutional Neural Network (TempCNN) \citep{pelletier2019temporal} has been proposed and evaluated specifically for the purpose of crop type mapping.
It is a comparatively lightweight architecture of three sequential 1D convolutional layers followed by batch normalization, a ReLU activation function, and dropout.
The encoded features are flattened and passed to a final fully connected layer with batch normalization, ReLU activation function, and dropout, as shown in \cref{fig:models:tempcnn}.
These features are then projected to scores per class using a final fully connected layer and softmax.

The tuneable hyperparameters of this model are 
the number of convolutional kernels $\hiddendims \in \SetDef{2^4,2^5,\dots, 2^8}$ that determine the dimensionality of hidden states, 
the respective kernel sizes $K \in \{3,5,7\}$,
and the dropout probability $\dropout \in [0,1]$. 

{
	\subsection{DuPLO}
	\label{sec:duplo}
	
	The \emph{DUal view Point deep Learning architecture for time series classificatiOn (DuPLO)} \cite{interdonato2019duplo} model is a complex deep learning model designed for crop type classification from sequences of small satellite images of five by five pixels. Since we do not use sequences of images in this work, we treated the time series as single-pixel images.
	It consists of two streams. A three-layer CNN stream uses 2D convolutions to aggregate spatial features independently of time.
	The second stream implements a 2D-CNN encoder and monodirectional RNN layer implemented by a Gated Recurrent Unit (GRU)~\cite{chung2014empirical} for temporal characteristics. 
	The recurrent neural network features $\M{H} \in \R^{H \times T} \mapsto \V{y} \in \R^H$ are then aggregated via a simplified {soft-attention} mechanism~\citep{britz2017efficient}, described in \cref{sec:soft-attention}.
	According to the authors, this model requires a special training procedure through additional weighted auxiliary losses from additional dense layers for each stream to simplify gradient propagation through.  
}

\subsection{Random Forests}

As shallow learning baseline, we tested a \emph{Random Forest (RF)} model using the \texttt{scikit-learn} framework \cite{pedregosa2011scikit}. 
{
	We augmented the raw reflectance input features by additional spectral indices, since feature extraction is usually required for a good performance. 
	Hence, we added the features 
	\emph{Normalized Difference Vegetation Index (NDVI)}, 
	\emph{Normalized Difference Water Index (NDWI)}, 
	\emph{Brightness Index (BI)}, 
	\emph{Inverted Red-Edge Chlorophyll Index (IRECI)}, and
	\emph{Enhanced Vegetation Index (EVI)} 
	to the time series data.
	Overall, the random forest model receives a flattened vector of all spectral values for each time. This results in a raw feature vector $\V{x}_\text{raw} \in \R^{1260}$ (70 times each 13 spectral bands and 5 spectral indices) and $\V{x}_\text{pre} \in \R^{345}$ (23 times each 10 spectral bands and 5 spectral indices).
	Also note that we did not obtain satisfactory results on the RF using the entire body of the dataset, likely due to class imbalance of the labels.
	Hence, we chose to sample a more uniform subset of up to 500 time series per class to train and test the RF model.
	We argue that this would rather advantage RF in the classification results (especially for the class-wise accuracies).
}

We tuned 
the number of trees $\SetDef{200,400,\dots,2000}$, 
the number of features to be considered at every split $\SetDef{\text{auto}, \text{sqrt}}$, 
the maximum depth of the trees $\SetDef{10,20,\dots,110}$, 
the minimum number of samples required to split a node $\SetDef{2,3,\dots,10}$, 
the minimum number of samples required at each leaf node $\SetDef{1,2,3,4}$, and 
whether or not to use bootstrapping $\SetDef{\text{true}, \text{false}}$. 

\section{Training Details}

We trained the \rnn, \msresnet, and \tempcnn models via mini-batch stochastic gradient descent with gradients scaled using the Adam optimizer \cite{kingma2014adam} with momentum parameters $\beta_1 = 0.9$ and $\beta_2 = 0.98$. 
The learning rate and weight decay parameters were sampled from log-uniform distributions over $[\num{e-8},\num{e-1}]$ and $[\num{e-12},\num{e-1}]$, respectively, and determined via hyperparameter tuning, as described in \cref{sec:tune}. 
Following Vaswani \etal \cite{vaswani2017attention}, we employed a learning rate scheduler for the \transformer model where the learning rate is first increased linearly for $N_\text{warmup} \in \{\num{e1},\num{e2},\num{e3}\}$ gradient steps and then decayed exponentially.
We stopped training early when the loss did not decrease over an average of the last 10 epochs. 
This condition needed to have been true for five epochs in a row to stop the training process.

\section{Model Selection}
\label{sec:tune}

The model architectures and training procedures depend on a variety of hyperparameters that may vary based on the objective and the dataset.
We described the tuneable hyperparameters for each model in the previous sections. 
These were, for the deep learning models, 
the dimensionality of the hidden vectors, 
the number of layers, 
the number of self-attention heads, 
the convolutional kernel sizes, 
the dropout probability, 
the number of \transformer warmup steps, 
the learning rate, 
and the weight decay rate.
To determine the optimal set of parameters, we sampled candidate hyperparameters from a hyperparameter space, trained on a subset of the training partition, and evaluated the performance on a validation set.
The partitioning scheme and spatial separation of the datasets are described in \cref{sec:data}.
Here, we first sampled hyperparameters from the search space at random for the first \num{34} models and recorded the validation performance measured in terms of the kappa metric \cite{cohen1960}.
Based on these initial points, a kernel density estimate scaled by the validation performance of the model determined a probability distribution that was used to sample the next best set of hyperparameters, following the protocol suggested by Bergstra \etal \citep{bergstra2013hyperopt} and implemented using their \textsc{HyperOpt} framework.
We combined this with an \emph{asynchronous successive halving} strategy \citep{li2018massively}.
This algorithm splits the \num{60} training epochs into four brackets with a grace period of \num{10} epochs.
After each bracket, only the best-performing half of the models continued training, while the other runs were stopped at the end of each bracket.
Both of these were implemented using the \textsc{Ray-Tune} framework \citep{liaw2018tune} that allows hyperparameter training of four models in parallel per GPU on a \textsc{Nvidia-DGX1}.
For each model architecture, we evaluated on \num{300} sets of hyperparameters for both preprocessed and raw datasets having resulted in \num{2400} model evaluations.

To reduce tuning time and the class imbalance in the data, we used only up to \num{500} samples per class. 
{
	The architecture of the \duplo model was precisely defined in \cite{interdonato2019duplo} with detailed specifications for hidden dimensions, learning procedure and learning rate. Hence, we employed the model directly as specified.
}

\section{Data}
\label{sec:data}

\begin{figure*}[t!]
	\subcaptionbox
	{%
		The three areas of interest in Bavaria, Hollfeld, Krumbach, and Bavarian Forest are located in different regional environments indicated by the elevation map. 
		We show the field parcels colored by crop type, along with the random train test split for each of the regions.%
		\label{fig:aoi:regions}%
	}
	[\linewidth][c]
	{%
		\tikzstyle{map} = [draw, rounded corners, fill=white]
		\input{images/areaofinterest.tikz}
	}%
	
	\begin{subfigure}[t]{.6\linewidth}
		\centering
		{\includegraphics[width=\linewidth]{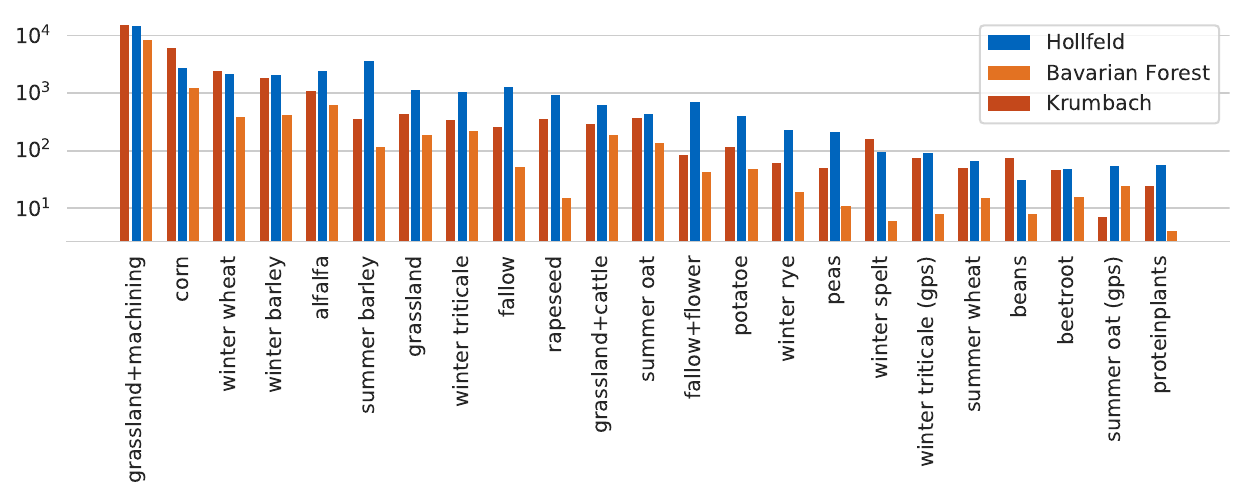}}%
		\caption{Class frequencies per region on a 23 class partition}
		\label{fig:aoi:hist23}
	\end{subfigure}%
	\hfill%
	\begin{subfigure}[t]{.4\linewidth}
		\centering
		{\includegraphics[width=\linewidth]{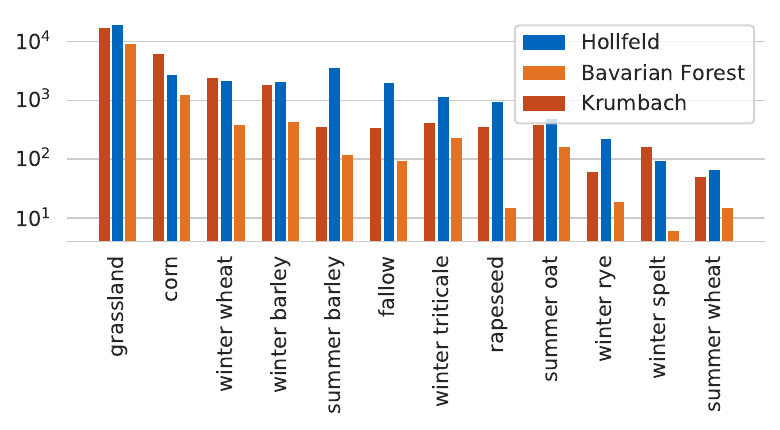}}%
		\caption{Class frequencies per region on a 12 class partition}
		\label{fig:aoi:hist12}
	\end{subfigure}%
	
	\caption{Test regions for labels and satellite time series}
	\label{fig:aoi}
\end{figure*}

We evaluated the mechanisms described in \cref{sec:method} that were implemented in the models of \cref{sec:models} on the task of crop type identification.
Crop type identification is a subtask of land cover and land use classification, where the model has to extract classification-relevant features and learn a discriminative decision function to separate the classes of vegetation. 
We analyze the extracted features later in \cref{sec:exp:qualitative:tsne}.
Vegetation life cycle events, known as phenology, provide a distinctive temporal signal to identify types of vegetation using a limited set of spectral channels.
This makes the temporal signal a key source of relevant features when learning a discriminative model with which to differentiate various types of vegetation and thus it is well-suited to testing the mechanisms and models of this work.

We focus on three spatially separate regions in Bavaria, as shown in \cref{fig:aoi}.
These regions are \emph{Hollfeld} in Upper Franconia, \emph{Krumbach} in Swabia, and a northeastern portion of the \emph{Bavarian Forest}.
These regions are located at a distance of approximately \SI{100}{\kilo\meter} from each other. 
While the climate conditions are comparatively similar, different elevations and soil conditions favor differences in the distribution of cultivated crops, as can be observed from the class distribution histograms in \cref{fig:aoi:hist23,fig:aoi:hist12}.

The label data for this study originates from a joint cooperation project with the \emph{Bavarian State Ministry of Food, Agriculture and Forestry (StMELF)} and the German remote sensing company \emph{GAF AG}.
This enabled us to obtain two Sentinel 2 time series datasets from the same field parcels. 
These are one dataset with raw top-of-atmosphere Sentinel 2 observations acquired with minimal effort from Google Earth Engine, as described in \cref{sec:raw}, and one preprocessed time series dataset provided by GAF AG which can be considered prototypical for the industry standard, as outlined in \cref{sec:pre}.

To obtain spatially separate partitions \citep{schratz2018performance} for \emph{training} of model parameters using gradient descent, \emph{validation} of hyperparameter sets, and final \emph{evaluation} of the model, we divided the three regions further into rectangular blocks of 4500m $\times$ 4500m with a \SI{500}{\meter} margin between blocks, as shown in the train-test split in \cref{fig:aoi:regions}.
These blocks were randomly assigned to training, validation, and evaluation partitions obeying a 4:1:1 ratio {which resulted in 94, 21, and 29 blocks, respectively}. 
We decided on such a block-wise spatial separation in order to enforce independence of the dataset partitions without implicit overfitting, as experimentally evaluated and observed in previous work \cite{russwurm2017temporal} and as recommended for geospatial data by further studies \cite{brenning2012spatial,schratz2018performance} focusing on the implicit bias of spatial auto-correlation. 

\subsection{Crop Type Labels}

The \emph{Common Agricultural Policy} of the European Union subsidizes farmers based on the types of crop they cultivate.
Each member country is required to gather geographical information on the geometry of the field parcels and the types of crop. 
This information is provided through obligatory surveys, as part of the subsidy application process, completed directly by the farmers.
National agencies monitor the correctness either by gathering control samples \textit{in-situ} or by utilizing remote sensing and Earth observation technology.

The crop label categories provided for this study were provided by StMELF.
They follow a \emph{long-tailed class distribution} with more than \num{269} distinct categories. 
Here, the most common \num{15}, \num{26}, and \num{62} categories cover \SI{90}{\percent}, \SI{95}{\percent}, and \SI{99}{\percent} of the field parcels, respectively.
In cooperation with StMELF and GAF AG, a set of land use and land cover categories was aggregated and selected with respect to the aims and objectives of the ministry.
From this aggregation, we selected two labeled datasets:
The first one contains \num{23} classes, as shown in \cref{fig:aoi:hist23}, and resembles the land use of the parcels. 
These categories cover, for instance, multiple types of grassland.
This dataset is challenging to classify, since multiple categories (\eg grassland for cattle use and for machining) share similar surface reflectance features measured by the satellite.
We also aggregated these categories further into a second dataset that focuses on \num{12} land cover categories, as shown in \cref{fig:aoi:hist12}.
By evaluating models on two datasets---a 23-class land use categorization and a 12-class land cover categorization---we aimed at reporting model accuracies from two objectives of different difficulty.

\subsection{Satellite Data}
\label{sec:data:sat}

\begin{figure}
	
	\input{images/preprocessed_example.tikz}
	\input{images/raw_example.tikz}
	
	\def\secondid{27-71460294}
	
	\subcaptionbox
	{raw time series of a meadow parcel\label{fig:sat:raw}}
	[.5\textwidth][c]
	{\rawexamplemeadow}%
	\hfill%
	\subcaptionbox
	{preprocessed time series of a meadow parcel\label{fig:sat:pre}}
	[.5\textwidth][c]
	{\preexamplemeadow}%
	
	\caption{%
		{note the dates on the x axis}
		An illustration and comparison of a raw and a preprocessed Sentinel 2 time series of the same meadow field parcel. 
		Preprocessing allows for a visual interpretation. 
		The onset of growth after time step $t=5$ is clearly visible. 
		Also, several cutting events can be observed over the vegetation period. 
		The preprocessed time series, however, contains repeated values due to temporal interpolation and cloud removal. 
		In the raw time series, most information from the measured signal is retained. 
		Noise caused by \eg atmospheric effects and clouds obscures the phenological events.
	}
	\label{fig:sat}

\end{figure}
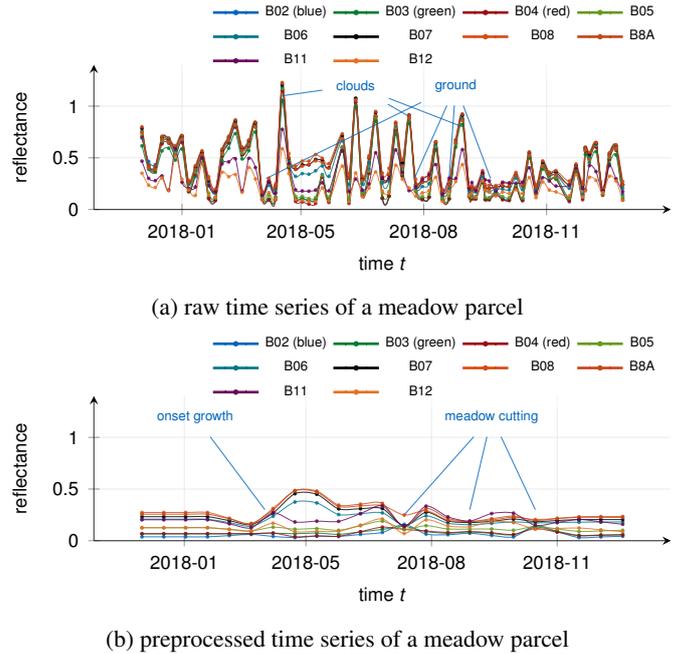

We utilized data from the optical Sentinel 2 satellite constellation that consists of two satellites orbiting the Earth in a sun-synchronous orbit on opposite tracks.
These satellites observe the same spot on the Earth's surface every two to five days, depending on the latitude. 
The data is acquired by a line-scanner capturing 13 spectral bands, ranging from 
ultraviolet wavelengths, for observing atmospheric water vapor, 
over optical and near-infrared wavelengths, sensitive to chlorophyll and photosynthesis,
up to short-wave infrared wavelengths that are sensitive to soil moisture. 
In regions where the sensor strips overlap, we observed approximately \num{140} measurements of the same point during the entire year of 2018, while on the strip centers we recorded only \num{70} observations within the same time range. 

\subsubsection{Raw Dataset}
\label{sec:raw}

For the raw Sentinel 2 dataset, we utilized the \emph{top-of-atmosphere} reflectance values at the processing level L1C.
This data was acquired from \emph{Google Earth Engine} \citep{gorelick2017google} and queried for each field parcel individually.
Pixels located within the boundaries of a field parcel were mean-aggregated into a single feature vector of \num{13} spectral bands at each time.
We show examples of a meadow parcel of the raw Sentinel 2 time series in \cref{fig:sat:raw}.
Note that cloud coverage, visible as positive peaks within the reflectance profiles, dominates the signal and makes a visual interpretation of this time series difficult.
This time series dataset is challenging to classify, but it can be acquired at minimal effort.
Hence, we benchmarked our models on this harder objective.

\subsubsection{Preprocessed Dataset}
\label{sec:pre}

For further evaluation, we had access to a second dataset that originated from the same publicly available top-of-atmosphere data products. 
In contrast to the raw dataset, this data was processed by GAF AG through their tested and operational preprocessing engine.
This process includes common preprocessing techniques such as \eg 
atmospheric correction, 
temporal selection of cloud-free observations, 
a focus on observations of the vegetative period, 
and cloud masking.
We show an example of the preprocessed dataset in \cref{fig:sat:pre} compared to the identical parcel in \cref{fig:sat:raw}.
Here, the cloudy observations have been identified and filtered by a separate cloud classification model. 
This makes this dataset easier to classify by shallow models, as distinctive phenological features, \ie onset of growth and cutting of the meadows, can be visually distinguished. 
Overall, this preprocessing pipeline can be considered prototypical for an industry standard, but it requires significant computational and design effort to generate.

\section{Experiments and Results}
\label{sec:experiments}

In this section, we experimentally compare the \transformer model, based on self-attention, to a recurrent neural network and two convolutional neural networks, as described in \cref{sec:models}.
This experimental section is structured in three parts:
First, in \cref{sec:exp:quantitative:all}, we show quantitative results on all evaluated neural network architectures.
We present results on preprocessed and raw Sentinel 2 data described in \cref{sec:data:sat} and on two sets of categories. 
One 23-class categorization evaluates model performance on land use classification while the other 12-class categorization focuses on land cover.
Next, in \cref{sec:exp:qualitative:backprop}, we analyze the ability of the models to suppress noise, \eg induced by clouds, in raw time series data by a feature importance analysis based on gradient backpropagation.  
Finally, in \cref{sec:exp:qualitative:self-attention,sec:exp:qualitative:tsne}, we focus specifically on the self-attention mechanism and analyze activation scores and internal states of the \transformer model in detail.

\subsection{Quantitative Model Evaluation}
\label{sec:exp:quantitative:all}

\begin{table*}
	\renewcommand{\arraystretch}{1}
	\setlength{\tabcolsep}{2pt}
	\scriptsize 
	
	\caption{Comparison of models on preprocessed (pre) and raw datasets on the 23-class land use and the 12-class land cover categorizations. The values reported are the mean and standard deviation of three models with the best, second-best, and third-best hyperparameter sets trained on the training and validation partitions and tested on the evaluation partition.}
	\label{tab:preraw}
	
	\subcaptionbox
	{Kappa metric 23-class dataset\label{tab:preraw:kappa}}
	[.5\linewidth][c]
	{\input{tables/preraw/kappa.tex}}%
	\subcaptionbox
	{Kappa metric 12-class dataset\label{tab:preraw2:kappa}}
	[.5\linewidth][c]
	{\input{tables/preraw2/kappa.tex}}%
	\\[\bigskipamount]
	\subcaptionbox
	{Overall accuracy metric 23-class dataset\label{tab:preraw:accuracy}}
	[.5\linewidth][c]
	{\input{tables/preraw/accuracy.tex}}%
	\subcaptionbox
	{Overall accuracy metric 12-class dataset\label{tab:preraw2:accuracy}}
	[.5\linewidth][c]
	{\input{tables/preraw2/accuracy.tex}}%
	\\[\bigskipamount]
	\subcaptionbox
	{Class-mean $f_1$ score 23-class dataset\label{tab:preraw:f1}}
	[.5\linewidth][c]
	{\input{tables/preraw/f1.tex}}%
	\subcaptionbox
	{Class-mean $f_1$ score 12-class dataset\label{tab:preraw2:f1}}
	[.5\linewidth][c]
	{\input{tables/preraw2/f1.tex}}%
	
\end{table*}

We compared the performance of the deep learning models \rnn~\citep{Hochreiter97:LST}, \transformer~\cite{vaswani2017attention}, \msresnet~\cite{wang2017time}, {the \duplo \\ model~\cite{interdonato2019duplo}}, and \tempcnn~\cite{pelletier2019temporal}, as well as an {RF} classifier as shallow baseline.
We determined the optimal hyperparameters separately for preprocessed and raw Sentinel 2 time series datasets, and for the 23-class and 12-class categorizations, as described in \cref{sec:tune}.
For each experiment, we trained and evaluated three different models with the best, second-best, and third-best hyperparameter configuration and random seeds for parameter initialization and composition of the training batches.
In \cref{tab:preraw}, we present the mean and standard deviation of the accuracy metrics from these three results. 
All models were trained and evaluated on block partitions for training and evaluation (\cf \cref{fig:aoi}) in all three regions, Hollfeld, Krumbach, and Bavarian forest. 
We evaluated these models on raw and preprocessed satellite time series on identical field parcels.

It is noteworthy that the overall accuracy measure, as presented in \cref{tab:preraw:accuracy,tab:preraw2:accuracy}, overrepresents frequent classes. 
Since the datasets used for this study show a heavily imbalanced class distribution, the accuracy of frequent classes dominates this metric. 
Nevertheless, it is an intuitive measure and a good representation of the example-wise accuracy, representative of the visual impression from observing a spatial map classification.
To account for the less frequent classes, we also report Cohen's kappa metric \citep{cohen1960} in \cref{tab:preraw:kappa,tab:preraw2:kappa}.
This is a correlation score that is frequently used in remote sensing and that normalizes the classification scores by the probability of a random chance prediction based on empirical class frequencies. 
As a further measure of performance, we report the $f_1$ score, \ie the harmonic mean of precision and recall, for each class and then average these over all classes in \cref{tab:preraw:f1,tab:preraw2:f1}.
In doing this, we ensured that all classes get equally weighted, disregarding the number of samples per category.
The class-mean $f_1$ scores are generally lower than accuracy and kappa, as we chose a large set of classes where some classes semantically overlap or have only a few examples, which makes it difficult for a data-driven neural network to learn feature extraction and decision function.
Overall, we aimed at comparing different properties of the classification models.
Hence, the overall accuracy reflects the classification accuracy per field parcel, while the $f_1$ score measures the accurate classification of all classes. 

\Cref{tab:preraw} reveals that data preprocessing had a positive effect on the accuracy, $f_1$ score, and kappa of all models. 
It seems that manual supervision during preprocessing improved the classification performance.
This better performance on preprocessed time series is, to a certain degree, expected, since this preprocessing makes a visual identification of classification-relevant events possible, as could be seen in \cref{fig:sat:pre}.
Comparing the visual examples at \cref{fig:sat} suggests that classification of the preprocessed datasets is an easier task than approximating a \\ preprocessing-like mechanism jointly together with classification in a holistic end-to-end model.
Overall, the region- and domain-specific expertise needed to design such preprocessing pipelines increases the classification performance compared to classification of the raw datasets where the deep learning models have to learn \\ preprocessing-like mechanisms solely based on the provided examples without access to model knowledge. 
This holds especially true if the number of samples is limited, if some examples are falsely labeled, or if the semantic representations of classes overlap.
Considering this, it is notable that the \rnn and \transformer models performed competitively well on raw data compared to preprocessed data.
Especially in the 12-class land cover categorization setting, the difference in performance was rather minor, speaking of $\SI{5}{\percent}$ in accuracy and $0.03$ in kappa score. 
In terms of the $f_1$ score, the \rnn variant achieved even better performance on raw data compared to preprocessed data, while it ranged behind the \transformer model.
Throughout both the 23-class dataset (\cf \cref{tab:preraw:accuracy,tab:preraw:kappa,tab:preraw:f1}) and the 12-class dataset (\cf \cref{tab:preraw2:accuracy,tab:preraw2:kappa,tab:preraw2:f1}) variants, all evaluated deep learning models performed similarly well on preprocessed data.
Interestingly, for the raw dataset partitions, the \rnn and \transformer models achieved slightly better accuracy, kappa, and $f_1$ score values compared to the \msresnet and \tempcnn variants.
In the 23-class setting, the difference is rather small with \SIrange{1}{2}{\percent} in overall accuracy and $0.01$ in kappa score, which seems not significant considering their reported variances.
The 12-class case, however, confirms this observation with a more pronounced difference of \numrange{0.03}{0.07} in kappa metric and \SIlist{5;11}{\percent} in accuracy.
In the next \cref{sec:exp:qualitative:backprop}, we will investigate this further through a feature importance analysis monitoring the backpropagation process.

When comparing the $f_1$ scores in \cref{tab:preraw:f1,tab:preraw2:f1} with the accuracy scores in \cref{tab:preraw:accuracy,tab:preraw2:accuracy} on raw data, we see that the convolutional \tempcnn model showed similar performance compared to the \msresnet model. 
The $f_1$ scores, however, show a lower score for \tempcnn. 
From these metrics, we can derive that the \tempcnn did classify the majority of field parcels accurately, but it achieved lower accuracies on some of the more infrequent class categories compared to \msresnet. 
This may be attributed to the shallower network topology. 
{
	The \duplo achieves very good classification accuracy on preprocessed satellite data, as it is on par with the other models.
	Since the model architecture is precisely defined by the authors, we could not run the model with different hyperparameter settings. Still, it outperforms the other models in some tasks on preprocessed data; see \cref{tab:preraw2:kappa,tab:preraw:kappa,tab:preraw:f1}. On the raw satellite data, it is behind the \transformer and \rnn models.
}

Interestingly, the RF baseline achieved competitive results compared to the deep learning models on preprocessed time series data.
It appears that preprocessing, as a form of feature extraction, helped the RF classifier to improve its performance significantly.
Without data preprocessing on the raw dataset, the RF baseline fell behind the deep learning models.
The kappa metric---which ranged $0.15$ worse in the 23-class land use variant and $0.1$ to $0.2$ worse in the 12-class land use variant---indicates that infrequent classes were classified less accurately by the RF classifier.
This got further confirmed by the poor $f_1$ score throughout both dataset variants, where the RF achieved worse performance on raw datasets compared to preprocessed datasets.
Overall, this stresses the necessity of feature extraction for shallow machine learning methods while demonstrating that---with sufficient manual effort in preprocessing and, thus, feature extraction---random forests can achieve competitive accuracies.

It may be concluded that the region- and domain-specific expert knowledge in data preprocessing helped all models to achieve better accuracies compared to the raw sentinel 2 time series dataset.
All models achieved similar accuracies for preprocessed datasets. 
Even the RF classifier showed a competitive performance compared to the deep learning models with similar scores in overall accuracy, albeit with slightly worse kappa and $f_1$ scores.
Both the \rnn and the \transformer models, which rely on recurrence and self-attention, achieved better accuracies on raw time series data compared to the convolutional models \msresnet and \tempcnn.
This effect was minor in the land use categorization with 23 classes and more pronounced in the land cover categorization with 12 classes, but overall it was consistent throughout this evaluation.
We will investigate the difference of recurrence and self-attention compared to convolution further in the next section.

{
	\subsection{Qualitative Model Evaluation}
	\label{sec:qualiatative}
	
	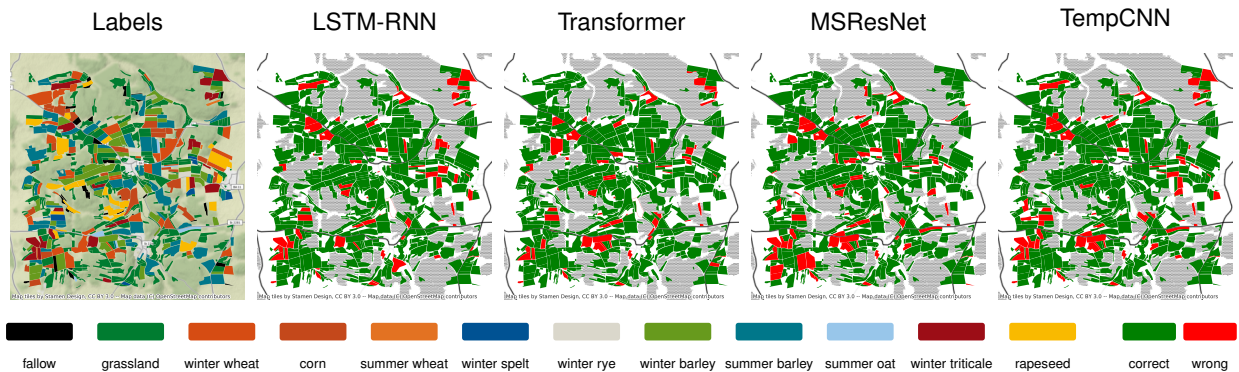
\begin{figure*}
		\centering\input{images/qualitative.tikz}
		\caption{Qualitative comparison of the classification models from region 4430000 E/W, 5528000 N/S in EPSG:31468}
		\label{fig:qualitative}
	\end{figure*}
	
	We conducted a qualitative analysis in addition to the quantitative model comparison in \cref{sec:exp:quantitative:all} and \cref{tab:preraw}.
	Here, we use the deep learning models to predict the 12 land cover classes on the raw time series dataset.
	Each time series, as in \cref{fig:sat:raw}, was obtained from the reflectance values aggregated over one field geometry, as shown in the left-most figure.
	The parcels were colored by crop type and obtained from a 4500m by 4500m block (4430000 E/W, 5528000 N/S in EPSG:31468) from the test partition.
	The figures toward the right show the fields that were correctly and wrongly classified by the respective models.
	As indicated by \cref{tab:preraw}, the classification accuracies do not vary significantly throughout the models so that no clear advantage between models is visible.

}

\subsection{Temporal Feature Importance by Gradient Backpropagation}
\label{sec:exp:qualitative:backprop}

\begin{figure*}[h]
	\input{images/backprop.tikz}
	
	\subcaptionbox
	{Example of a \emph{summer barley} time series\label{fig:backprop:barley}}
	[.5\linewidth][c]
	{\backpropexamplesummerbarley}%
	\hfill%
	\subcaptionbox
	{Example of a \emph{corn} time series\label{fig:backprop:corn}}
	[.5\linewidth][c]
	{\backpropexamplecorn}%
	
	\caption{Input feature importance analysis by gradients of the input time series $\M{X}$}
	\label{fig:backprop}
\end{figure*}
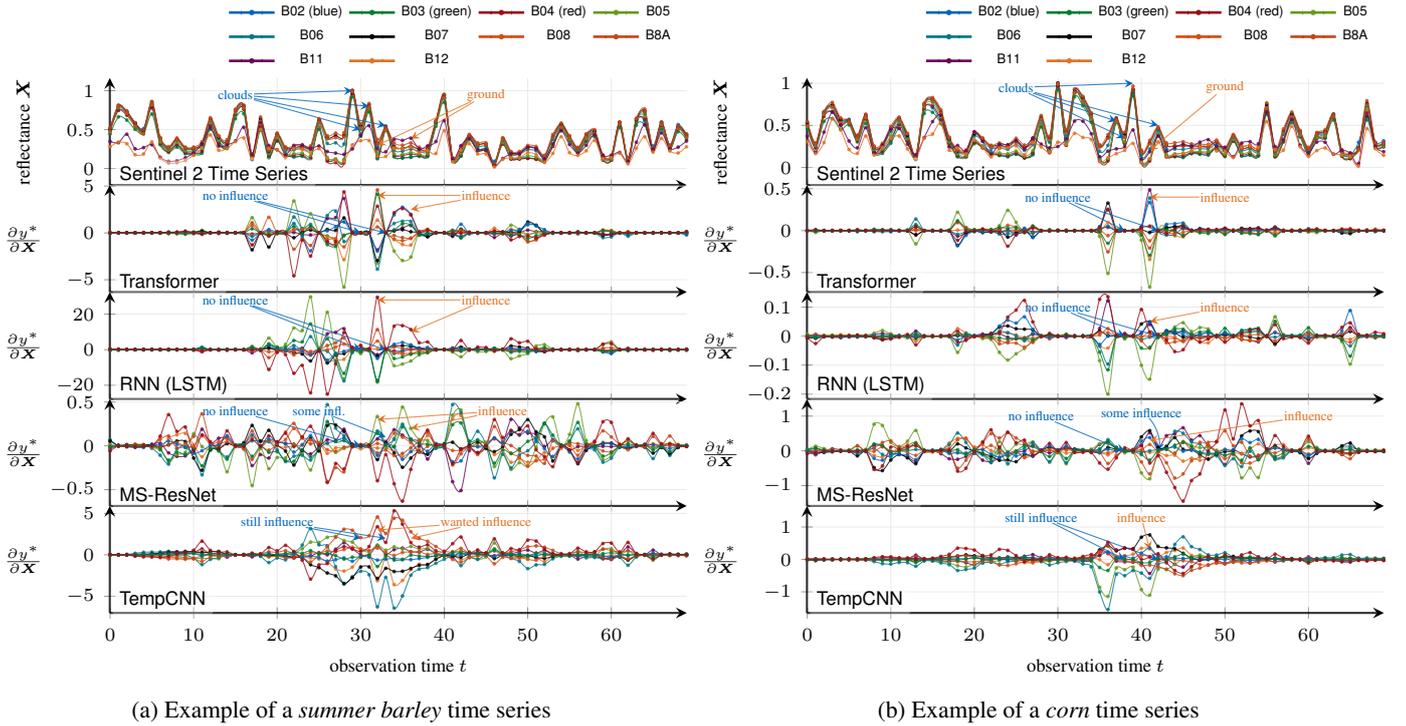

\begin{figure*}[h]
	
	\subcaptionbox
	{Self-Attention Matrix of Head 1\label{fig:attn:h1matrix}}
	[.3\textwidth][c]
	{\includegraphics[width=.3\textwidth]{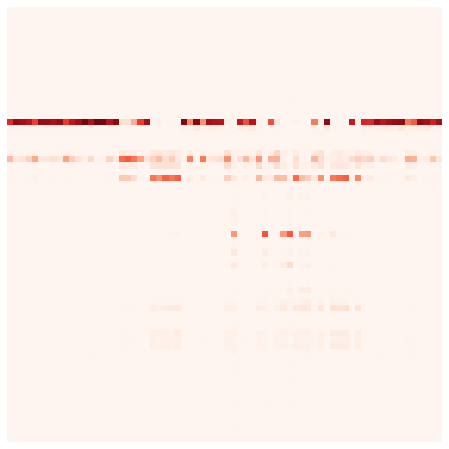}}%
	\hfill%
	\subcaptionbox
	{Head 1 as bipartite graph visualization in context of the input time series\label{fig:attn:h1graph}}
	[.7\textwidth][c]
	{
		\begin{tikzpicture}
		\node(a){\includegraphics[width=.7\textwidth]{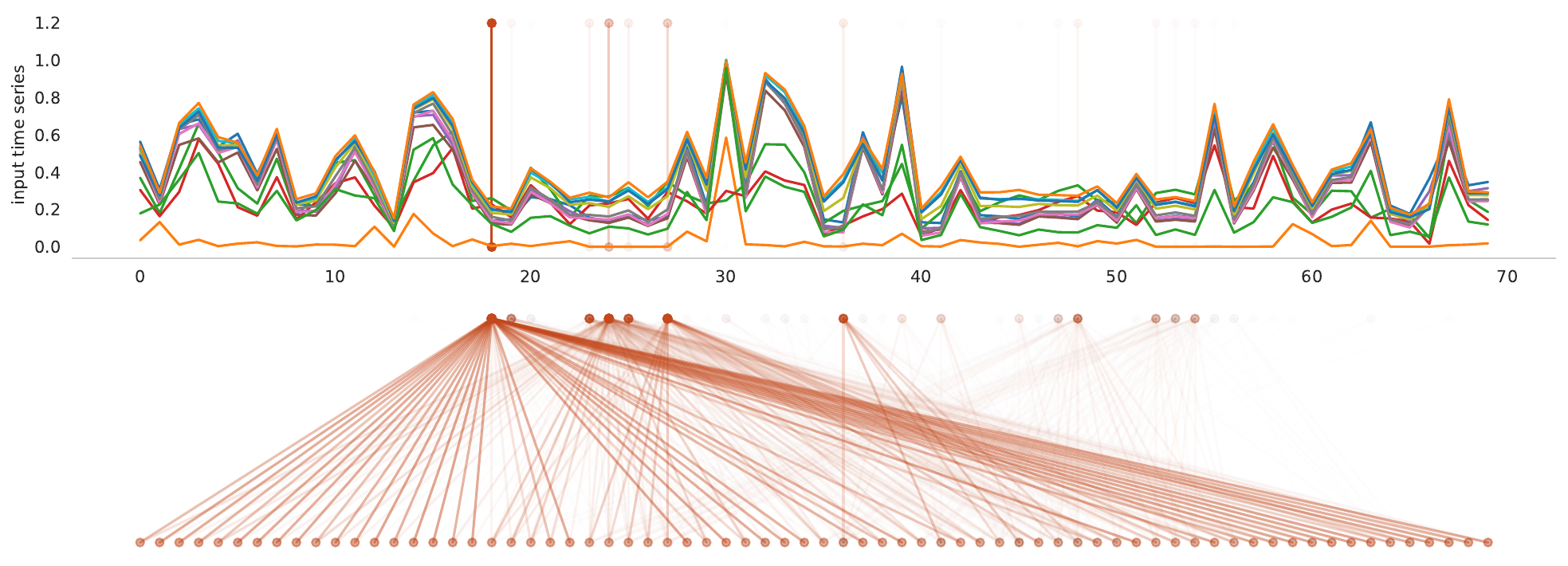}};
		\node[font=\tiny\sffamily](c) at (-2.5,1.8){clouds};
		\node[font=\tiny\sffamily](g) at (-1.5,1.8){ground};
		\draw[-stealth] (c) -- ++(-.1,-.3);
		\draw[-stealth] (g) -- ++(.2,-1);
		\draw[-stealth] (g) -- ++(-.6,-1.1);
		\end{tikzpicture}
	}%
	
	\subcaptionbox
	{Self-Attention Matrix of Head 2\label{fig:attn:h2matrix}}
	[.3\textwidth][c]
	{\includegraphics[width=.3\textwidth]{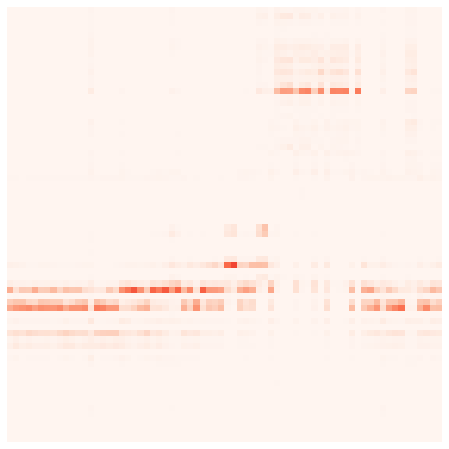}}%
	\hfill%
	\subcaptionbox
	{Head 2 as bipartite graph visualization in context of the input time series\label{fig:attn:h2graph}}
	[.7\textwidth][c]
	{\includegraphics[width=.7\textwidth]{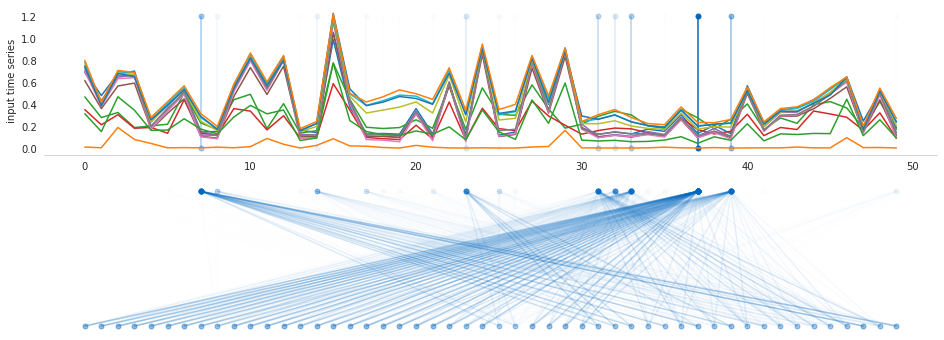}}%
	
	\caption{Visual illustration of the self-attention scores from a pre-trained \transformer self-attention model}
	\label{fig:self-attention}
\end{figure*}

In the previous section, we observed that models based on recurrence and self-attention achieved better accuracy metrics on noisy raw data compared to convolutional models.
Here, we investigate this further through a simple, yet effective, feature importance analysis based on gradient backpropagation.

Deep learning models are differentiable functions $f_\Mweight$ that approximate a mapping from an input tensor $\M{X}$ to a ground truth tensor $\V{y}$, as detailed in \cref{sec:method}. 
In the training process, gradients are backpropagated using \cref{eq:gradientdescent} in order to adjust the model weights $\Mweight$. 
Since the model $f_\Mweight$ is fully differentiable, we can use the same mechanism to backpropagate gradients further up to the input tensor $\M{X}$.  
When we start from the highest predicted score $y^\ast = \max(\yhat)$, we can propagate gradients through the entire network back to the individual elements $\V{x}_t = (x_t^\text{B01},x_t^\text{B02},x_t^\text{B03},\dots)\T$ of the input tensor 
$\M{X} = (\V{x}_0,\dots,\V{x}_T)$.
These gradients $\frac{\partial y^\ast}{\partial\M{X}}$ reflect the influence of each input data element on the current class prediction. 
Vanishing gradients indicate that any change in the input has no effect on the prediction.
Positive or negative gradients would suggest an increase or decrease of the predicted score if the input changed at these times.
We would like to emphasize that this analysis is model-agnostic, can be implemented in a few lines of code in current deep learning frameworks that utilize automatic differentiation, and does not require ground truth labels.\footnote{Our implementation is available at \github/blob/master/notebooks/FeatureImportance.ipynb}.

For this experiment, we estimated the influence of each input time step on the classification prediction for each of the evaluated networks, \ie
\rnn (recurrence) and
\transformer (self-attention), 
as well as
\msresnet and 
\tempcnn (convolution).
\Cref{fig:backprop} illustrates this by means of two separate examples, a corn parcel and a summer barley parcel.
The top figures in each show the input time series $\M{X}$ as a sequence of raw Sentinel 2 reflectances over the year of 2018.
In the raw time series, we can identify atmospheric noise and clouds as positive peaks in the data.
These are caused by the high reflectance values of clouds throughout all spectral bands. 
The presence of a cloud at a given point in time does not provide any additional information about the covered surface; thus, it should be considered irrelevant for the classification.
The following rows display the gradients $\frac{\partial y^\ast}{\partial\M{X}}$ for each of the respective models. 
These plots indicate the influence of the measurement at the particular input time in the top row on the classification prediction of the respective model.

The results of this experiment show that the gradients through the \rnn and \transformer networks were only non-zero for comparatively few observations.
This indicates that only a few key time-points were necessary to extract the classification-relevant information from the raw time series.
Only time-points where the surface was not obscured by clouds had non-zero gradients.
{Time instances with positive peaks in reflectance, likely induced by clouds, did not have influence on the classification prediction, as indicated by vanishing gradients.
	This experimentally shows that the \rnn and \transformer models have been optimized to automatically identify and suppress time instances that are not classification relevant, \eg through cloud coverage, and these were learned purely from data.}

The convolution-based model architectures \msresnet and \tempcnn show, in general, a similar behavior. 
However, these models seem to have some non-zero gradients on cloudy observations.
This is especially true for the comparatively shallow \tempcnn architecture where the cloudy observations at $t=30$ in \cref{fig:backprop:barley} and $t=38$ in \cref{fig:backprop:corn} still have influence in the classification prediction.
The \msresnet model also seems to be able to extract features from the entire temporal range.
Since the data ranges from January to December, it is expected that the particular crop class is visible only during the summer periods.
The \msresnet model seems to still include features from the winter periods.

Summarizing the above, the recurrence- and self-attention-based models \rnn and \transformer seem to extract selective temporal features, while the convolution-based models utilize the entire time series.
Also, the \rnn and \transformer models were able to transition from a time observation with high gradients to zero gradients without any time delay.
It appears that the convolution-based models \tempcnn and \msresnet require at least two subsequent observations to reduce the influence of the particular time instance.
This can be observed from \cref{fig:backprop:barley} at $t=30$ and $t=34$:
The \msresnet model could reduce the influence of the observations at $t=30$ (indicated by zero gradients) when three subsequent observations were cloudy.
However, when only one observation was cloudy, as in $t=34$, \msresnet was unable to suppress this observation completely. 
In contrast, the models \rnn and \transformer were able to suppress these cloudy observations even if they appeared in a single time instance.

We attribute this difference in the influence of input observations to the respective mechanisms for feature extraction.
Recurrent networks utilize internal gates that can control the influence of the particular time instance to a hidden memory state. 
Our previous work \citep{russwurm2018multi}, which visualized internal LSTM states, supports this hypothesis for recurrence.
Similarly, self-attention enables a model to select specific observations by assigning a large attention score to them.
In contrast, convolutions always extract features from a local neighborhood. 
Hence, it seems to be more difficult for convolutional architectures to ignore sudden appearances of irrelevant observations within a temporal sequence, as indicated by the non-zero gradients at cloudy observations.

\subsection{Qualitative Analysis of Self-Attention Scores}
\label{sec:exp:qualitative:self-attention}

In the previous section, we experimentally observed that the \transformer model was able to suppress the influence of an observation for classification-irrelevant observations, \eg clouds.
Here, we concentrate on the self-attention mechanism \\ (\cf \cref{sec:self-attention}) that was implemented in the \transformer model and analyze the attention scores on the example of the cornfield parcels from the previous experiment, as shown in \cref{fig:backprop:corn}.

The \transformer models realize multiheaded self-attention as multiple attention mechanisms in parallel.
To recall, every self-attention mechanism calculates an attention matrix $\M{A} \in [0,1]^{T \times T}$ using the softmax operation following \cref{eq:attention}.
These attention scores define the influence of an input time feature on a higher-level output time feature. 
We visualize the values of this matrix in \cref{fig:attn:h1matrix,fig:attn:h2matrix} for three attention heads from the first self-attention layer (of three).
This matrix can be seen as an adjacency matrix between input nodes and output nodes.
In \cref{fig:attn:h1graph,fig:attn:h2graph}, we alternatively show the same matrix as bipartite graphs. 
The former matrix elements are now shown as weighted directed edges between input and output nodes. 
The strength of the attention is indicated by the opacity of the edge.
Above the attention graph, we show the input time series of the example as a reference to allow for visual interpretation of the time instances.
Here, the attention scores are drawn as vertical lines corresponding to the values in $\M{A}$.

From these figures, we can directly observe that the self-attention scores focus on distinct events with each head. 
{
	While the first head appears to focus on the early-middle part of the time series, the second head extracts features from the early and late time series.}
Also, the attention scores seem to redistribute features over time. 
Hence, even though hidden feature tensors throughout the network still maintain a temporal dimension, the temporal consistency with the input time series loses context.
Consistent with the previous experiment in \cref{sec:exp:qualitative:backprop}, we observed that the attention scores did not focus on the strong positive peaks in the time series, which indicates a cloudy observation.
Hence, we can conclude that self-attention mechanisms are a key tool in suppressing the non-classification-relevant \\ cloudy observations which explain the zero-gradients of the previous experiment.

{
	\subsection{Comparison of Self-attention with Soft Attention as implemented in \duplo}
	\label{sec:attentioncomparison}
	
	\begin{figure*}[h]
		\subcaptionbox
		{}
		[.49\linewidth][c]
		{\includegraphics[width=.49\textwidth]{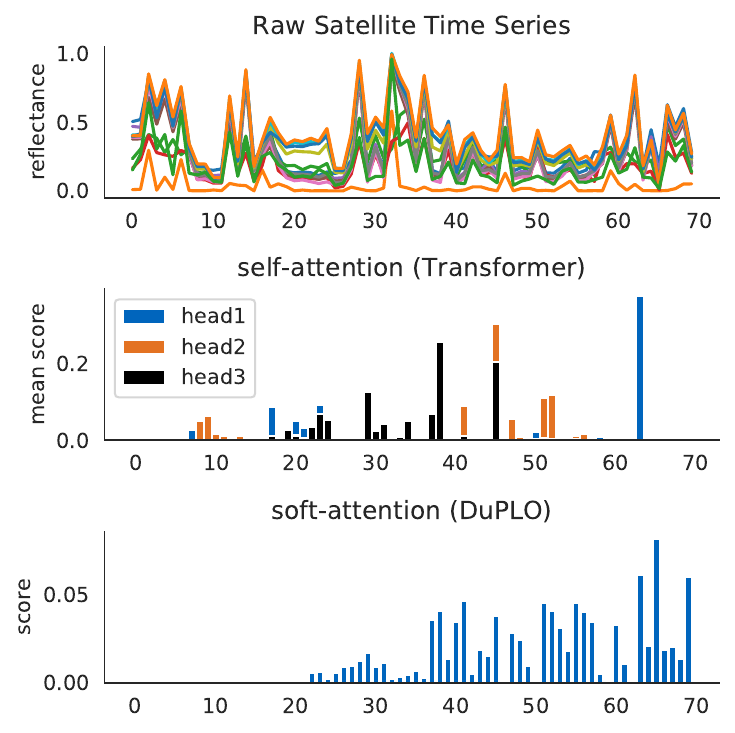}}
		\hfill
		\subcaptionbox
		{}
		[.49\linewidth][c]
		{\includegraphics[width=.49\textwidth]{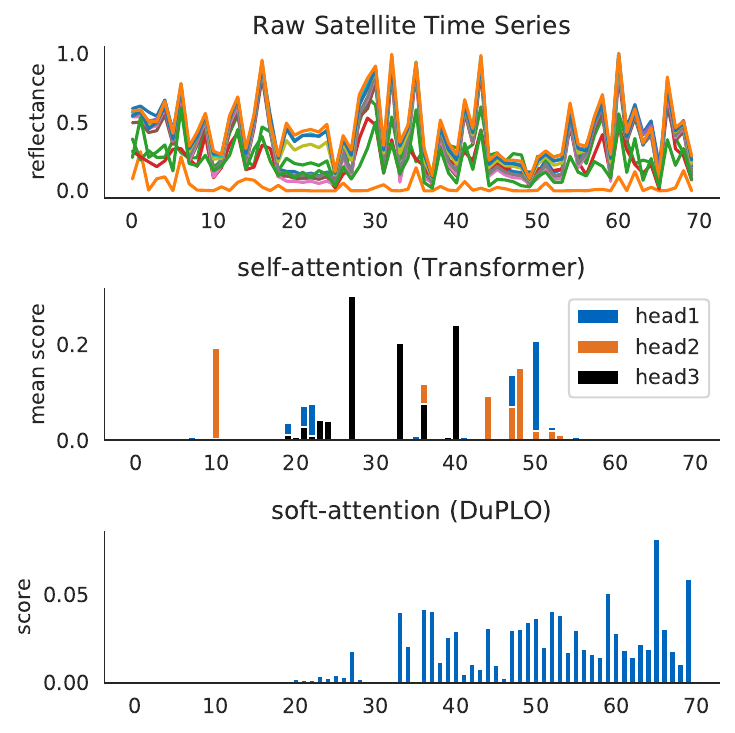}}
		\caption{Comparison of self-attention scores, as implemented in the \transformer model, with soft-attention scores used in the \duplo model}
		\label{fig:attentioncomparison}
		
	\end{figure*}
	
	In the previous experiment, we explored the self-attention mechanism, as implemented in cascaded layers of the \transformer model.
	The \duplo model\cite{interdonato2019duplo}, as described in \cref{sec:duplo}, proposes a neural network architecture that combines convolutional layers, recurrent layers, and a soft-attention mechanism, and has been employed for crop type mapping in remote sensing.
	In \cref{tab:preraw}, we observed that the \duplo architecture achieved good accuracies on preprocessed data, but it was outperformed by a purely recurrent model and the self-attention model on the raw time series data.
	This brings into question the effectiveness of soft-attention (\cref{sec:soft-attention}) compared to self-attention (\cref{sec:self-attention}), which we address in this experiment.
	
	In \cref{fig:attentioncomparison}, we draw the attention scores from two test samples shown in the top row. In the second row, we see the average self-attention scores $\V{\alpha} = \frac{1}{T_{out}}\sum_{i=0}^{T_{out}}\M{A}_{i}$ from the three attention heads of the first \transformer layer per input time. This corresponds to the average over the columns in \cref{fig:attn:h2matrix,fig:attn:h1matrix}. 
	The third row shows the soft-attention scores, as obtained from the \duplo model.
	
	Comparing the attention scores, we observe that, consistent with the previous experiment, the self-attention scores focus on a few single time instances that obtain classification-relevant features.  
	In contrast, the comparatively simpler soft-attention is not able to focus on single time instances so that multiple neighboring time instances have attention scores greater than zero.
	Note that the soft-attention scores are calculated within the recurrent branch of the \duplo model using a complex combination of convolutional layers augmented with a single monodirectional recurrent layer.
	The limitations of these attention scores may also be caused by the preceding layers.
	For instance, the monodirectional encoding of the GRU of the \duplo model likely leads to higher-level features in the latter half of the time series.
	It is likely that this effect causes the higher attention scores focusing on the later time instances where more expressive recurrent features have been extracted from the single monodirectional GRU layer.
	
	Overall, we conclude that the self-attention mechanism, due to its being designed with three dense layers, can be more expressive than the simpler soft-attention that requires only one dense layer.
}

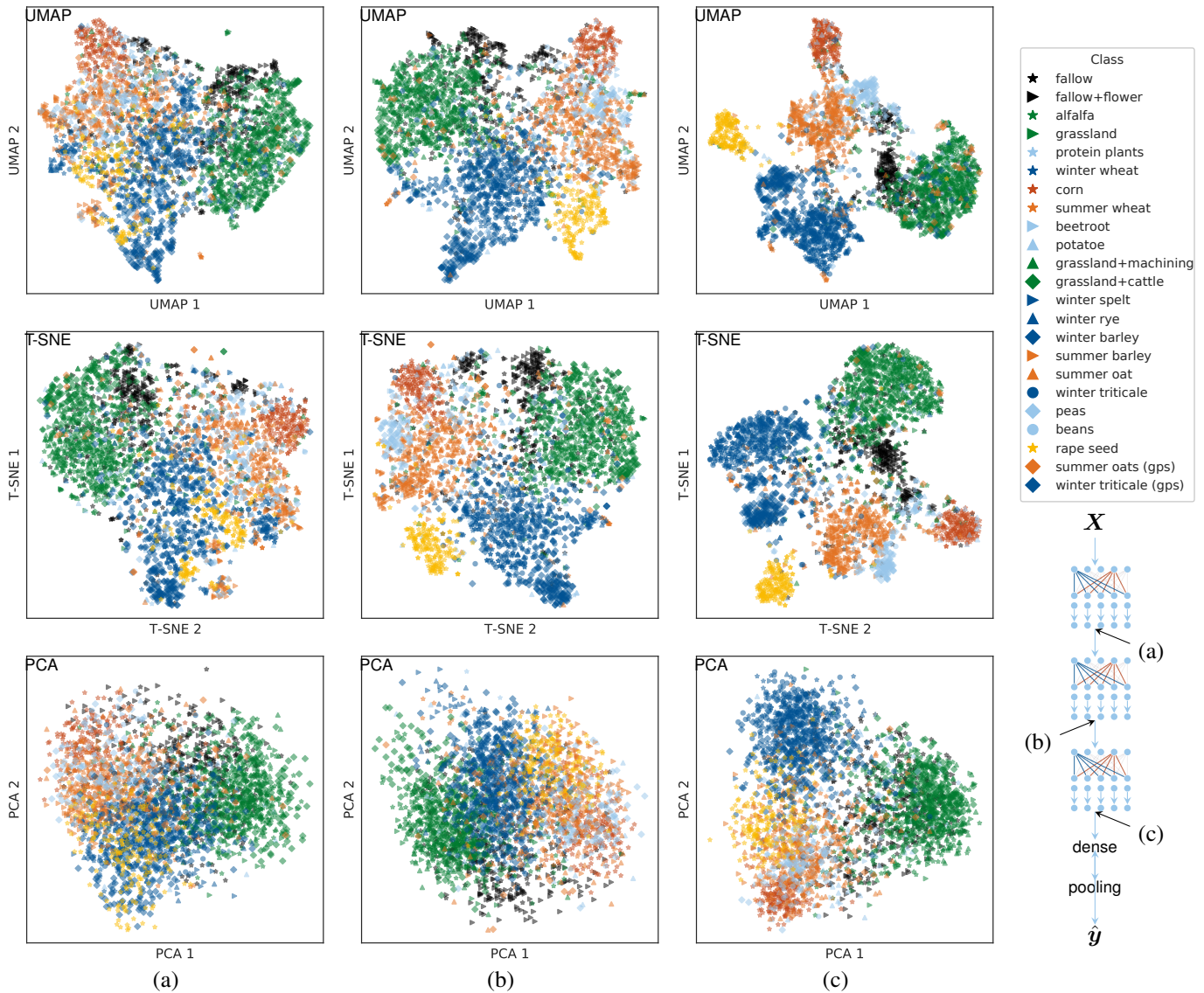
\begin{figure*}[h!]
	
	\input{images/embedding.tex}
	\caption{{Feature embedding at varying depths of the \transformer network. Each self-attention layer increases the separability of the extracted features.}}
	\label{fig:embeddings}
\end{figure*}

{
	\subsection{Feature Separability Analyses through Manifold Embeddings}
	\label{sec:exp:qualitative:tsne}

	\newcommand{\marker}[1]{\raisebox{-.1em}{\includegraphics[height=.8em]{images/embedding/markers/#1}}}
	
	Deep learning models, in general, extract features of increasing complexity throughout the cascaded layer architectures \cite{goodfellow2016deep}. 
	In this experiment, we analyzed this property by visualizing the hidden features at varying deeper layers of the \transformer architecture, as shown in \cref{fig:embeddings}. 
	{
		For this analysis, we extract $D=128$-dimensional feature vectors after encoding time series samples from the test dataset.
		We project them into a two-dimensional embedding space using three different dimensionality reduction techniques. These were \emph{t-distributed Stochastic Neighbor Embedding (t-SNE)}~\citep{maaten:tsne} which approximates the data distribution in embedding space, \emph{Uniform Manifold Approximation and Projection (UMAP)}~\citep{mcinnes2018umap} which preserves more of the global structure compared to t-SNE and was used in Mohammadimanesh et al. (2019)~\citep{mohammadimanesh2019new}, as well as a comparatively straightforward \emph{Principal Component Analysis (PCA)}.
	}
	We used a trained \transformer network with three self-attention layers and $D=128$ hidden dimensions.
	To obtain the features, we passed the entire test dataset over all regions through a pre-trained \transformer network and recorded the hidden features after each self-attention layer and after the last dense layer before max pooling and softmax activation, as visualized in \cref{fig:embeddings}.
	The original features $\M{H} \in \R^{(T=70) \times (D=128)}$ were then averaged along the time dimension to obtain $\V{H}^\prime \in \R^{D=128}$ and mapped into the final embedding space $\R^{D=2}$.
	The illustration shows the representations of the embedded 200 samples for each of the 23 land use classes.
	We augmented the final plots with a key indicating the classes' identities and a schematic picture of the network topology. 
	Since the class definitions are rather broad with, for instance, multiple categories of grassland, we grouped them by color but assigned varying marker shapes to each class.  
	Note that the embeddings rely on the extracted features without taking the actual ground truth class label into account.
	One can easily observe that the transformed features become more and more separable with each deeper network layer throughout all embedding techniques.
	This becomes apparent by the formation clusters throughout the layers.   
	Looking at the last layer, which is closest to the final prediction, we can identify distinct separable manifolds for corn (\marker{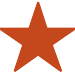}) and rapeseed (\marker{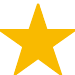}).
	This suggests that these classes are separable in the feature space.
	This observation is clearly observable for both the nonlinear t-SNE and the UMAP projections while the separability in the first two principal components in the linear PCA projection is less pronounced.

	In summary, the feature extraction capabilities learned by this self-attention network are consistent with the assumptions based on the domain knowledge of the particular classes.
	Methodologically, we observed that features from deeper cascaded network layers were distinctive and that field parcels of the same label were mapped in similar regions of the embedding space.
	We would like to stress that this type of analysis is not unique to self-attention networks and can be reproduced for many deep learning architectures that extract features in cascaded layers.
}

\section{Conclusion}
\label{sec:conclusion}

We quantitatively and qualitatively analyzed self-attention for application in multitemporal Earth observation. 
We performed a large-scale quantitative comparison in \cref{sec:exp:quantitative:all} and evaluated multiple model architectures that rely on self-attention, recurrence, and convolution, and an RF baseline.
These models were compared from multiple angles by reporting their performance on preprocessed and raw Sentinel 2 time series as well as a land use- and a land cover-oriented set of classes.
All models performed equally well on preprocessed data. Even the RF baseline achieved competitive overall accuracy. 
This indicated that the choice of model architecture was less critical when extensive data preprocessing was utilized.
{
While we concentrated on one type of preprocessed data in this work, we should emphasize that a comparison to models that include distinct model-specific preprocessing, as used in CCDC \cite{zhu2014continuous}, BFAST~\cite{verbesselt2010detecting}, LandTrendr~\cite{kennedy2010detecting}, or TimeSat~\cite{eklundh2016timesat}, will be necessary to confirm these findings on a general scale.}

Our results on raw unprocessed Sentinel 2 time series data showed that the \transformer and \rnn architectures achieved better accuracies compared to the convolutional models.  
We investigated this further by a feature importance analysis in \cref{sec:exp:qualitative:backprop} using gradients where we observed that the self-attention and recurrence mechanisms helped to suppress {information not relevant for classification} in the time series.
For the \transformer, this was realized by learning weights in the attention mechanism which enable the model to specifically focus on some observations, as could be observed in \cref{fig:attention} in \cref{sec:exp:qualitative:self-attention}.
Finally, we analyzed the \transformer layers that were able to learn increasingly separable representations of the classes in \cref{sec:exp:qualitative:tsne}.

To summarize, we observed that self-attention allowed neural networks to extract features from specific time instances with raw optical satellite time series.
We observed that self-attention and recurrence were, due to the design of their feature extraction, more robust in dealing with noise in the data and that they could better suppress cloudy observations in raw time series.

In the future, a robust classification of raw time series data without region-specific expert knowledge will be key if we are to quantitatively exploit the satellite data that is published daily. 
We hope to have contributed a step in this direction with this comprehensive study that evaluated model accuracies on preprocessed and raw data for a variety of mechanisms for time series classification.

\section{Acknowledgments}

{We would like to thank the reviewer for his persistence and the detailed and constructive critique that significantly improved the quality of this work.
}
Also, we would like to thank Sabine Erbe from the Bavarian Ministry of Agriculture for the continuous and persistent support and valuable domain knowledge as well as Markus Ziegler and Christoph Schmidt from GAF AG for their collaboration in the joint crop type classification project which provided objective, method, and preprocessed time series data for this study.
The work of Marc Rußwurm was funded by the \emph{German Federal Ministry for Economic Affairs and Energy (BMWi)} under reference \emph{50EE1908}.

\bibliography{bib/used.bib}

\end{document}

%% file: preamble.tex
\usepackage{tummath}
\usepackage{tumcolors}

\usepackage[utf8]{inputenc}
\usepackage[english]{babel}
\usepackage[]{csquotes}

\usepackage{siunitx}

\usepackage{animate}
\usepackage{glossaries}
\usepackage{graphicx}
\usepackage{nth}
\usepackage{xspace}

\PassOptionsToPackage{cmyk}{xcolors}

\usepackage{booktabs}
\usepackage{contour}

\usepackage{tikz}

\usepackage{pgfplots}
\usepgfplotslibrary{groupplots}
\usepgfplotslibrary{dateplot}

\pgfplotsset{compat=1.15} 
\usepackage{pgfmath}
\usetikzlibrary{pgfplots.dateplot}
\usepgfplotslibrary{dateplot}

\usepackage{booktabs}
\usepackage{array}
\newcolumntype{x}{l}
\newcolumntype{X}{>{}l}
\newcolumntype{v}[1]{>{\raggedright\hspace{0pt}}p{#1}}
\newcolumntype{V}[1]{>{\scriptsize\raggedright\hspace{0pt}}p{#1}}

\definecolor{fusionremovedcolor}{HTML}{00CA43}
\definecolor{fusionaddedcolor}{HTML}{FF00F7}
\definecolor{overlapcolor}{HTML}{FAC843}

\usepgfplotslibrary{external}
\tikzsetexternalprefix{tikz/}
\tikzexternaldisable

\pgfdeclarelayer{background}
\pgfdeclarelayer{foreground}
\pgfsetlayers{background,main,foreground}

\usetikzlibrary{arrows} 
\usetikzlibrary{backgrounds}
\usetikzlibrary{fit}
\usetikzlibrary{shapes}
\usetikzlibrary{calc}
\usetikzlibrary{positioning}

\usetikzlibrary{matrix}
\usepackage[nodayofweek,level]{datetime}
\usetikzlibrary{spy}

\usetikzlibrary{3d}
\tikzstyle{perspective3d}=[
x={(0.5cm,0.5cm)}, y={(1cm,0cm)}, z={(0cm,1cm)}]

\colorlet{traincolor}{tumbluelight}
\colorlet{validcolor}{tumbluedark}
\colorlet{evalcolor}{tumorange}

\colorlet{querycolor}{tumorange}
\colorlet{attentioncolor}{tumred}
\colorlet{valuecolor}{tumblue}
\colorlet{attentionoutcolor}{tumbluedark}
\colorlet{keycolor}{tumgreen}

\colorlet{forwardcolor}{tumblue}
\colorlet{backwardcolor}{tumorange}

\colorlet{activationcolor}{tumblue}
\colorlet{gridcolor}{tumgraylight}
\colorlet{contextonecolor}{tumorange}
\colorlet{contexttwocolor}{tumorange!80}
\colorlet{contextthreecolor}{tumorange!60}
\colorlet{contextfourcolor}{tumorange!40}

\colorlet{tensorcolor}{forwardcolor}

\colorlet{classcolor}{tumivory}
\colorlet{encodercolor}{tumblue}
\colorlet{encodercolor}{tumred}

\newcommand*{\etal}{%
  \@ifnextchar{.}
    {\textit{et\,al}}
    {\@ifnextchar{,}
      {\textit{et\,al}.}
      {\textit{et\,al}.\,}}%
}
\makeatother

\newcommand{\ie}{\textit{i{.}e{.}}, }
\newcommand{\eg}{\textit{e{.}g{.}}, }
\newcommand{\cf}{\textit{cf{.}}\,}

\newcommand{\mathleft}{\@fleqntrue\@mathmargin0pt}
\newcommand{\mathcenter}{\@fleqnfalse}

\usepackage{times}
\usepackage{epsfig}
\usepackage{graphicx}
\usepackage{amsmath}
\usepackage{amssymb}
\usepackage[utf8]{inputenc}
\usepackage{booktabs}
\usepackage{subcaption}

\usepackage{cancel}

\usepackage[capitalize]{cleveref}

\newcommand{\yhat}{\hat{\V{y}}}

\newcommand{\Mweight}{ {\M{\Theta}} }
\newcommand{\Vweight}{ {\V{\theta}} }
\newcommand{\lr}{\ensuremath{\mu}}

\newcommand{\Tin}{\ensuremath{{ T_\text{in} }}}
\newcommand{\Tout}{\ensuremath{{ T_\text{out} }}}
\newcommand{\Din}{\ensuremath{{ D_\text{in} }}}
\newcommand{\Dout}{\ensuremath{{ D_\text{out} }}}
\newcommand{\Dh}{\ensuremath{{ D_\text{h} }}}

\newcommand{\hiddendims}{\ensuremath{D_h}}
\newcommand{\dropout}{\ensuremath{p_\text{drop}}}
\newcommand{\numlayers}{\ensuremath{L}}
\newcommand{\numheads}{\ensuremath{H}}

\newcommand{\rnn}{LSTM-RNN\xspace}
\newcommand{\transformer}{Transformer\xspace}
\newcommand{\msresnet}{MS-ResNet\xspace}
\newcommand{\tempcnn}{TempCNN\xspace}
\newcommand{\duplo}{DuPLO\xspace}

\newcommand{\new}[1]{#1} 

\usepackage{mathtools}

\definecolor{evalcolor}{HTML}{3F3F3F}
\definecolor{traincolor}{HTML}{B98951}
\definecolor{validcolor}{HTML}{3F4BBE}

\colorlet{colortrain}{tumblue}
\colorlet{colorinfer}{tumblack}

\colorlet{earlinesscolor}{tumblue}
\colorlet{accuracycolor}{tumorange}

\colorlet{stdcolor}{tumbluelight}
\colorlet{mediancolor}{tumorange}
\colorlet{meancolor}{tumblue}

\colorlet{b1color}{tumaubergine}
\colorlet{b9color}{tumblack}
\colorlet{b10color}{tumblack}

\colorlet{b2color}{tumblue}
\colorlet{b3color}{tumgreen}
\colorlet{b4color}{tumdarkred}

\colorlet{b5color}{tumlimegreen}
\colorlet{b6color}{tumturquoise}
\colorlet{b7color}{tumblack}
\colorlet{b8color}{tumredorange}
\colorlet{b8Acolor}{tumred}

\colorlet{b11color}{tumaubergine}
\colorlet{b12color}{tumorange}

\colorlet{epsilon0color}{tumorange}
\colorlet{epsilon1color}{tumblue}
\colorlet{epsilon10color}{tumblack}

\colorlet{meadowcolor}{tumbluemedium}
\colorlet{wbarleycolor}{tumbluedark}
\colorlet{corncolor}{tumorange}
\colorlet{wheatcolor}{tumgreen}
\colorlet{sbarleycolor}{tumred}
\colorlet{clovercolor}{tumturquoise}
\colorlet{triticalecolor}{tumsand}

\tikzstyle{rnn}=[draw,circle, inner sep=.1em]
\tikzstyle{norm}=[rounded corners,draw]
\tikzstyle{annot}=[rounded corners, fill=tumblue!20]
\tikzstyle{infer}=[-stealth, shorten >=.0em, shorten <=.0em, colorinfer]
\tikzstyle{loss}=[fill=tumblue!10, rounded corners, font=\small]
\tikzstyle{grad}=[colortrain]

\tikzstyle{test} = [thick]
\tikzstyle{train} = [thin, dotted]

\usepackage[inline]{enumitem}
\setenumerate{label=\roman*), itemsep=3pt, topsep=3pt, font=\itshape}

\usepackage{setspace}
\usepackage{eqparbox, etoolbox}

\newlist{rdescription}{description}{1}

\AtBeginEnvironment{rdescription}{%
	\setlist[rdescription]{leftmargin =\dimexpr\eqboxwidth{Des}+\labelsep}}%

\usepackage[eulergreek]{sansmath}
\pgfplotsset{
	y tick label style={/pgf/number format/.cd,%
		scaled y ticks = false,
		set thousands separator={},
		fixed},
	x tick label style={/pgf/number format/.cd,%
		scaled x ticks = false,
		set decimal separator={,},
		fixed},
	tick label style = {font=\sansmath\sffamily\scriptsize},
	every axis label = {
		font=\sansmath\sffamily\scriptsize},
	every axis/.append style={
		axis lines=left, 
		enlargelimits, 
		thick},
	legend style = {font=\sansmath\sffamily, draw=none, rounded corners, fill opacity=.5, text opacity=1},
	label style = {font=\sansmath\sffamily\scriptsize},
	grid style={line width=.1pt, draw=gray!10},
	major grid style={line width=.2pt,draw=tumgraylight},
	grid=both,
}

\tikzstyle{circ} = [circle, draw=white, fill=tumblue, inner sep=1pt]

\tikzstyle{proba} = [circle, draw=tumgray, inner sep=2.5pt, fill=tumorange]

\tikzstyle{tsmark} = [mark=|,mark size=2pt]

\definecolor{s1}{RGB}{228, 26, 28}
\definecolor{s2}{RGB}{55, 126, 184}
\definecolor{s3}{RGB}{77, 175, 74}
\definecolor{s4}{RGB}{152, 78, 163}
\definecolor{s5}{RGB}{255, 127, 0}
\pgfplotscreateplotcyclelist{featurecolorlist}{
	s1,every mark/.append style={fill=s1},mark=*\\
	s2,every mark/.append style={fill=s2},mark=*\\
	s3,every mark/.append style={fill=s3},mark=*\\
	s4,every mark/.append style={fill=s4},mark=*\\
	s5,every mark/.append style={fill=s5},mark=*\\
}

\newcommand{\rawtimeseriestwo}[1]{
	
	\begin{tikzpicture}[baseline=-2em, inner sep=0]
	
	\begin{axis}[
	thin,
	width=5.5cm,
	hide axis,
	enlargelimits=0,
	height=3cm,
	ymin=0, ymax=1.4,
	draw opacity=.8,
	smooth=0.01
	]

	\addplot[b11color, mark=*,mark size=.5pt] table [x=t, y=B11, col sep=comma, forget plot] {images/example/#1};
	\addplot[b12color, mark=*,mark size=.5pt] table [x=t, y=B12, col sep=comma] {images/example/#1};
	
	\addplot[b5color, mark=*,mark size=.5pt] table [x=t, y=B05, col sep=comma, forget plot] {images/example/#1};
	\addplot[b6color, mark=*,mark size=.5pt] table [x=t, y=B06, col sep=comma, forget plot] {images/example/#1};
	\addplot[b7color, mark=*,mark size=.5pt] table [x=t, y=B07, col sep=comma, forget plot] {images/example/#1};
	\addplot[b8color, mark=*,mark size=.5pt] table [x=t, y=B08, col sep=comma, forget plot] {images/example/#1};
	\addplot[b8Acolor, mark=*,mark size=.5pt] table [x=t, y=B8A, col sep=comma] {images/example/#1};
	
	\addplot[b2color, mark=*,mark size=.5pt] table [x=t, y=B02, col sep=comma, forget plot] {images/example/#1};
	\addplot[b3color, mark=*,mark size=.5pt] table [x=t, y=B03, col sep=comma, forget plot] {images/example/#1};
	\addplot[b4color, mark=*,mark size=.5pt] table [x=t, y=B04, col sep=comma] {images/example/#1};
	
	\end{axis}
	
	\end{tikzpicture}	
}

\newcommand{\cev}[1]{\reflectbox{\ensuremath{\vec{\reflectbox{\ensuremath{#1}}}}}}

%% file: images/attentionmatrices.tikz
\newcommand{\attnquery}{%
	
		\begin{tikzpicture}[scale=0.4]
		\node[draw=querycolor, circle, fill=querycolor, fill opacity=.2, text opacity=1, font=\small, inner sep=.2em](a) at (0, 0){};
		\node[draw=querycolor, circle, fill=querycolor, fill opacity=.2, text opacity=1, font=\small, inner sep=.2em](a) at (0, 1){};
		\node[draw=querycolor, circle, fill=querycolor, fill opacity=.2, text opacity=1, font=\small, inner sep=.2em](a) at (0, 2){};
		\node[draw=querycolor, circle, fill=querycolor, fill opacity=.2, text opacity=1, font=\small, inner sep=.2em](a) at (0, 3){};
		
		\node[draw=querycolor, circle, fill=querycolor, fill opacity=.1, text opacity=1, font=\small, inner sep=.2em](b) at (1, 0){};
		\node[draw=querycolor, circle, fill=querycolor, fill opacity=.1, text opacity=1, font=\small, inner sep=.2em](b) at (1, 1){};
		\node[draw=querycolor, circle, fill=querycolor, fill opacity=.1, text opacity=1, font=\small, inner sep=.2em](b) at (1, 2){};
		\node[draw=querycolor, circle, fill=querycolor, fill opacity=.1, text opacity=1, font=\small, inner sep=.2em](b) at (1, 3){};
		
		\node[draw=querycolor, circle, fill=querycolor, fill opacity=.1, text opacity=1, font=\small, inner sep=.2em](b) at (2, 0){};
		\node[draw=querycolor, circle, fill=querycolor, fill opacity=.1, text opacity=1, font=\small, inner sep=.2em](b) at (2, 1){};
		\node[draw=querycolor, circle, fill=querycolor, fill opacity=.1, text opacity=1, font=\small, inner sep=.2em](b) at (2, 2){};
		\node[draw=querycolor, circle, fill=querycolor, fill opacity=.1, text opacity=1, font=\small, inner sep=.2em](b) at (2, 3){};
		
		\node[draw=querycolor, circle, fill=querycolor, fill opacity=.1, text opacity=1, font=\small, inner sep=.2em](b) at (3, 0){};
		\node[draw=querycolor, circle, fill=querycolor, fill opacity=.1, text opacity=1, font=\small, inner sep=.2em](b) at (3, 1){};
		\node[draw=querycolor, circle, fill=querycolor, fill opacity=.1, text opacity=1, font=\small, inner sep=.2em](b) at (3, 2){};
		\node[draw=querycolor, circle, fill=querycolor, fill opacity=.1, text opacity=1, font=\small, inner sep=.2em](b) at (3, 3){};
		\end{tikzpicture}
	
}

\newcommand{\attention}{%
	
		\begin{tikzpicture}[scale=0.4]
		\node[draw=attentioncolor, circle, fill=attentioncolor, fill opacity=.4, text opacity=1, font=\small, inner sep=.2em](a) at (1,0){};
		\node[draw=attentioncolor, circle, fill=attentioncolor, fill opacity=.8, text opacity=1, font=\small, inner sep=.2em](b) at (2,0){};
		\node[draw=attentioncolor, circle, fill=attentioncolor, fill opacity=.2, text opacity=1, font=\small, inner sep=.2em](c) at (3,0){};
		\node[draw=attentioncolor, circle, fill=attentioncolor, fill opacity=.6, text opacity=1, font=\small, inner sep=.2em](d) at (4,0){};
		
		\node[draw=attentioncolor, circle, fill=attentioncolor, fill opacity=.4, text opacity=1, font=\small, inner sep=.2em](a) at (1,1){};
		\node[draw=attentioncolor, circle, fill=attentioncolor, fill opacity=.8, text opacity=1, font=\small, inner sep=.2em](b) at (2,1){};
		\node[draw=attentioncolor, circle, fill=attentioncolor, fill opacity=.2, text opacity=1, font=\small, inner sep=.2em](c) at (3,1){};
		\node[draw=attentioncolor, circle, fill=attentioncolor, fill opacity=.6, text opacity=1, font=\small, inner sep=.2em](d) at (4,1){};
		
		\node[draw=attentioncolor, circle, fill=attentioncolor, fill opacity=.4, text opacity=1, font=\small, inner sep=.2em](a) at (1,2){};
		\node[draw=attentioncolor, circle, fill=attentioncolor, fill opacity=.8, text opacity=1, font=\small, inner sep=.2em](b) at (2,2){};
		\node[draw=attentioncolor, circle, fill=attentioncolor, fill opacity=.2, text opacity=1, font=\small, inner sep=.2em](c) at (3,2){};
		\node[draw=attentioncolor, circle, fill=attentioncolor, fill opacity=.6, text opacity=1, font=\small, inner sep=.2em](d) at (4,2){};
		
		\node[draw=attentioncolor, circle, fill=attentioncolor, fill opacity=.4, text opacity=1, font=\small, inner sep=.2em](a) at (1,3){};
		\node[draw=attentioncolor, circle, fill=attentioncolor, fill opacity=.8, text opacity=1, font=\small, inner sep=.2em](b) at (2,3){};
		\node[draw=attentioncolor, circle, fill=attentioncolor, fill opacity=.2, text opacity=1, font=\small, inner sep=.2em](c) at (3,3){};
		\node[draw=attentioncolor, circle, fill=attentioncolor, fill opacity=.6, text opacity=1, font=\small, inner sep=.2em](d) at (4,3){};
		\end{tikzpicture}
	
}

\newcommand{\attnv}{%
		\begin{tikzpicture}[scale=0.4]
		\node[draw=valuecolor, circle, fill=valuecolor, fill opacity=.2, text opacity=1, font=\small, inner sep=.2em](a) at (0, 1){};
		\node[draw=valuecolor, circle, fill=valuecolor, fill opacity=.1, text opacity=1, font=\small, inner sep=.2em](b) at (0, 2){};
		\node[draw=valuecolor, circle, fill=valuecolor, fill opacity=.3, text opacity=1, font=\small, inner sep=.2em](c) at (0, 3){};
		\node[draw=valuecolor, circle, fill=valuecolor, fill opacity=.4, text opacity=1, font=\small, inner sep=.2em](d) at (0, 4){};
		
		\node[draw=valuecolor, circle, fill=valuecolor, fill opacity=.2, text opacity=1, font=\small, inner sep=.2em](a) at (1, 1){};
		\node[draw=valuecolor, circle, fill=valuecolor, fill opacity=.1, text opacity=1, font=\small, inner sep=.2em](b) at (1, 2){};
		\node[draw=valuecolor, circle, fill=valuecolor, fill opacity=.3, text opacity=1, font=\small, inner sep=.2em](c) at (1, 3){};
		\node[draw=valuecolor, circle, fill=valuecolor, fill opacity=.4, text opacity=1, font=\small, inner sep=.2em](d) at (1, 4){};
		\end{tikzpicture}
}

\newcommand{\attnout}{%

		\begin{tikzpicture}[scale=0.4]
		\node[draw=tumblack, circle, fill=attentionoutcolor, text=white, text opacity=1, font=\small, inner sep=.2em](d) at (0,0){};
		\node[draw=tumblack, circle, fill=attentionoutcolor, text=white, text opacity=1, font=\small, inner sep=.2em](d) at (0,1){};
		\node[draw=tumblack, circle, fill=attentionoutcolor, text=white, text opacity=1, font=\small, inner sep=.2em](d) at (0,2){};
		\node[draw=tumblack, circle, fill=attentionoutcolor, text=white, text opacity=1, font=\small, inner sep=.2em](d) at (0,3){};
		
		\node[draw=tumblack, circle, fill=attentionoutcolor, text=white, text opacity=1, font=\small, inner sep=.2em](d) at (1,0){};
		\node[draw=tumblack, circle, fill=attentionoutcolor, text=white, text opacity=1, font=\small, inner sep=.2em](d) at (1,1){};
		\node[draw=tumblack, circle, fill=attentionoutcolor, text=white, text opacity=1, font=\small, inner sep=.2em](d) at (1,2){};
		\node[draw=tumblack, circle, fill=attentionoutcolor, text=white, text opacity=1, font=\small, inner sep=.2em](d) at (1,3){};
		\end{tikzpicture}
}

\newcommand{\attnkey}{

		\begin{tikzpicture}[scale=0.4]
		\node[draw=keycolor, circle, fill=keycolor, fill opacity=.2, text opacity=1, font=\small, inner sep=.2em](a) at (0, 0){};
		\node[draw=keycolor, circle, fill=keycolor, fill opacity=.1, text opacity=1, font=\small, inner sep=.2em](b) at (1, 0){};
		\node[draw=keycolor, circle, fill=keycolor, fill opacity=.3, text opacity=1, font=\small, inner sep=.2em](c) at (2, 0){};
		\node[draw=keycolor, circle, fill=keycolor, fill opacity=.4, text opacity=1, font=\small, inner sep=.2em](d) at (3, 0){};
		
		\node[draw=keycolor, circle, fill=keycolor, fill opacity=.2, text opacity=1, font=\small, inner sep=.2em](a) at (0, 1){};
		\node[draw=keycolor, circle, fill=keycolor, fill opacity=.1, text opacity=1, font=\small, inner sep=.2em](b) at (1, 1){};
		\node[draw=keycolor, circle, fill=keycolor, fill opacity=.3, text opacity=1, font=\small, inner sep=.2em](c) at (2, 1){};
		\node[draw=keycolor, circle, fill=keycolor, fill opacity=.4, text opacity=1, font=\small, inner sep=.2em](d) at (3, 1){};
		
		\node[draw=keycolor, circle, fill=keycolor, fill opacity=.2, text opacity=1, font=\small, inner sep=.2em](a) at (0, 2){};
		\node[draw=keycolor, circle, fill=keycolor, fill opacity=.1, text opacity=1, font=\small, inner sep=.2em](b) at (1, 2){};
		\node[draw=keycolor, circle, fill=keycolor, fill opacity=.3, text opacity=1, font=\small, inner sep=.2em](c) at (2, 2){};
		\node[draw=keycolor, circle, fill=keycolor, fill opacity=.4, text opacity=1, font=\small, inner sep=.2em](d) at (3, 2){};
		
		\node[draw=keycolor, circle, fill=keycolor, fill opacity=.2, text opacity=1, font=\small, inner sep=.2em](a) at (0, 3){};
		\node[draw=keycolor, circle, fill=keycolor, fill opacity=.1, text opacity=1, font=\small, inner sep=.2em](b) at (1, 3){};
		\node[draw=keycolor, circle, fill=keycolor, fill opacity=.3, text opacity=1, font=\small, inner sep=.2em](c) at (2, 3){};
		\node[draw=keycolor, circle, fill=keycolor, fill opacity=.4, text opacity=1, font=\small, inner sep=.2em](d) at (3, 3){};
		\end{tikzpicture}
	
}

\tikzsetnextfilename{attentionmatrix}
\begin{tikzpicture}[node distance=.2em]

\node[label={below:${\M{A}\T}$}, draw=attentioncolor, rounded corners](alpha){\attention};
\node[right=of alpha, label={right:$\M{H}$}](out){\attnout};
\node[above=of out, label={above:$\M{V}$}, draw=valuecolor, rounded corners](v){\attnv};

\node[above=of alpha, label={above:$\M{K}$}, draw=keycolor, rounded corners](k){\attnkey};
\node[left=of alpha, label={below:$\M{Q}\T$}, draw=querycolor, rounded corners](q){\attnquery};

\node[fill=white, xshift=3.5em, text=black, fill opacity=0.75, text opacity=1, rounded corners, inner sep=0, font=\footnotesize] at (v){$\Mweight_V\T\M{X}$};
\node[fill=white, text=black, fill opacity=0.75,inner sep=0, text opacity=1, rounded corners, font=\footnotesize] at (k){$\Mweight_K\T\M{X}$};
\node[fill=white, text=black, fill opacity=0.75,inner sep=0, text opacity=1, rounded corners, font=\footnotesize] at (q){$\Mweight_Q\T\M{X}$};

\end{tikzpicture}

%% file: images/attention_graph.tikz
\tikzsetnextfilename{attentiongraph}
\begin{tikzpicture}[yscale=3]

\node[draw=valuecolor, circle, fill=valuecolor, fill opacity=.2, text opacity=1, font=\small, inner sep=.2em](d) at (1,.35){};
\node[draw=valuecolor, circle, fill=valuecolor, fill opacity=.1, text opacity=1, font=\small, inner sep=.2em](e) at (2,.05){};
\node[draw=valuecolor, circle, fill=valuecolor, fill opacity=.3, text opacity=1, font=\small, inner sep=.2em](f) at (3,.42){};
\node[draw=valuecolor, circle, fill=valuecolor, fill opacity=.4, text opacity=1, font=\small, inner sep=.2em](g) at (4,.25){};
\draw (d) -- (e) -- (f) -- (g);

\node[draw=valuecolor, circle, fill=valuecolor, fill opacity=.2, text opacity=1, font=\small, inner sep=.2em](a) at (1,.3){};
\node[draw=valuecolor, circle, fill=valuecolor, fill opacity=.1, text opacity=1, font=\small, inner sep=.2em](b) at (2,.1){};
\node[draw=valuecolor, circle, fill=valuecolor, fill opacity=.3, text opacity=1, font=\small, inner sep=.2em](c) at (3,.3){};
\node[draw=valuecolor, circle, fill=valuecolor, fill opacity=.4, text opacity=1, font=\small, inner sep=.2em](d) at (4,.4){};

\draw (a) -- (b) -- (c) -- (d);

\node[draw=white, circle, fill=attentionoutcolor, text=white, fill opacity=1, text opacity=1, font=\small, inner sep=.2em](outba) at (1.05,-.48) {};
\node[draw=white, circle, fill=attentionoutcolor, text=white, fill opacity=1, text opacity=1, font=\small, inner sep=.2em](outbb) at (2.05,-.48) {};
\node[draw=white, circle, fill=attentionoutcolor, text=white, fill opacity=1, text opacity=1, font=\small, inner sep=.2em](outbc) at (3.05,-.48) {};
\node[draw=white, circle, fill=attentionoutcolor, text=white, fill opacity=1, text opacity=1, font=\small, inner sep=.2em](outbd) at (4.05,-.48) {};
	
\node[draw=white, circle, fill=attentionoutcolor, text=white, fill opacity=1, text opacity=1, font=\small, inner sep=.2em](outa) at (1,-.5) {};
\node[draw=white, circle, fill=attentionoutcolor, text=white, fill opacity=1, text opacity=1, font=\small, inner sep=.2em](outb) at (2,-.5) {};
\node[draw=white, circle, fill=attentionoutcolor, text=white, fill opacity=1, text opacity=1, font=\small, inner sep=.2em](outc) at (3,-.5) {};
\node[draw=white, circle, fill=attentionoutcolor, text=white, fill opacity=1, text opacity=1, font=\small, inner sep=.2em](outd) at (4,-.5) {};

\node[text=valuecolor, xshift=-2em, yshift=-1em] at (a){$\M{V}:$};
\node[text=attentionoutcolor, xshift=-2em] at (outa){$\M{H}:$};
\node[text=attentioncolor, xshift=-2em, yshift=3.5em] at (outa){$\M{A}:$};

\draw[-stealth, draw=tumred, opacity=.4, line width=.4] (a) -- (outa);
\draw[-stealth, draw=tumred, opacity=.8, line width=.8] (b) -- (outa);
\draw[-stealth, draw=tumred, opacity=.2, line width=.2] (c) -- (outa);
\draw[-stealth, draw=tumred, opacity=.6, line width=.6] (d) -- (outa);

\draw[-stealth, draw=tumred, opacity=.4, line width=.4] (a) -- (outb);
\draw[-stealth, draw=tumred, opacity=.8, line width=.8] (b) -- (outb);
\draw[-stealth, draw=tumred, opacity=.2, line width=.2] (c) -- (outb);
\draw[-stealth, draw=tumred, opacity=.6, line width=.6] (d) -- (outb);

\draw[-stealth, draw=tumred, opacity=.4, line width=.4] (a) -- (outc);
\draw[-stealth, draw=tumred, opacity=.8, line width=.8] (b) -- (outc);
\draw[-stealth, draw=tumred, opacity=.2, line width=.2] (c) -- (outc);
\draw[-stealth, draw=tumred, opacity=.6, line width=.6] (d) -- (outc);

\draw[-stealth, draw=tumred, opacity=.4, line width=.4] (a) -- (outd);
\draw[-stealth, draw=tumred, opacity=.8, line width=.8] (b) -- (outd);
\draw[-stealth, draw=tumred, opacity=.2, line width=.2] (c) -- (outd);
\draw[-stealth, draw=tumred, opacity=.6, line width=.6] (d) -- (outd);

\end{tikzpicture}

%% file: images/models/lstm.tikz
	\tikzstyle{conn} = [-stealth, draw=tumblue, rounded corners=1pt]
	\tikzstyle{fconn} = [-stealth, draw=tumbluelight, rounded corners=1pt, opacity=.2]
	\tikzstyle{bconn} = [-stealth, draw=tumbluedark, rounded corners=1pt, opacity=.2]
	\tikzstyle{cell} = [circle, fill=white, draw, inner sep=3pt]
	\tikzstyle{txtlabel} = [font=\tiny]
	
	\newcommand{\connect}[2]{
		\draw[conn] (#1) |- ($ (#1)!0.5!(#2) $) -| (#2);
	}
	
	\newcommand{\lstm}{

	\tikzset{external/export next=false}
	\begin{tikzpicture}[xscale=.5, yscale=-.5]
		\coordinate(in) at (1,-1);
		\coordinate(bin) at (1.25,-1);

		\node[cell](f0) at (0,0){};
		\node[cell](f1) at (1,0){};
		\node[cell](f2) at (2,0){};
		\node[txtlabel](f3) at (3.5,0){$(\vec{\V{c}},\vec{\V{h}})$};
		
		\coordinate(out) at (1,2.3);
		\coordinate(bout) at (1.5,2.3);
		
		\draw[conn] (f0) -- (f1);
		\draw[conn] (f1) -- (f2);
		\draw[conn] (f2) -- ++(.8,0);

		\foreach \n in {f0,f1,f2} {\draw[fconn] (in) |- ($ (in)!0.5!(\n) $) -| (\n);}
		\foreach \n in {f0,f1,f2} {\draw[fconn] (\n) |- ($ (\n)!0.85!(out) $) -| (out);}
		
		\node[cell](b0) at (0.5,1){};
		\node[cell](b1) at (1.5,1){};
		\node[cell](b2) at (2.5,1){};
		\node[txtlabel](b3) at (-1,1){$(\cev{\V{c}},\cev{\V{h}})$};
		
		\foreach \n in {b0,b1,b2} {\draw[bconn] (bin)++(.3,0) |- ($ (bin)!0.4!(\n) $) -| (\n);}
		\foreach \n in {b0,b1,b2} {\draw[bconn] (\n) |- ($ (\n)!0.5!(bout) $) -| (bout);}
		
		\draw[conn] (b2) -- (b1);
		\draw[conn] (b1) -- (b0);
		\draw[conn] (b0) -- (b3);
		
	\end{tikzpicture}
	}
	\tikzset{external/export next=false}
	\begin{tikzpicture}[yscale=-1.2]
		\node(x) at (0,0){\M{X}};
		\node[inner sep=0](l1) at (0,1.2){\lstm};
		\node[inner sep=0](l2) at (0,2.4){\lstm};
		\node[font=\small\sffamily, inner sep=0](a) at (0,3.5){$\V{c} = \text{concat}\left(\vec{\V{c}},\cev{\V{c}}\right)$}; 
		\node[inner sep=0,font=\small\sffamily](s) at (0,4){softmax$(\Mweight\T\V{c})$};
		\node[inner sep=0](y) at (0,4.5){\V{y}};
		
		\draw[-stealth, tumbluelight] (a) -- (s);
		\draw[-stealth, tumbluelight] (s) -- (y);
	\end{tikzpicture}

%% file: images/models/transformer.tikz
	\tikzstyle{conn} = [-stealth, draw=tumblue, rounded corners=1pt]
	\tikzstyle{n} = [circle, inner sep=1pt, fill=tumbluelight]
	\tikzstyle{s} = [tumred]
	\tikzstyle{txtlabel} = [font=\tiny]

	\newcommand{\attn}{
	\tikzset{external/export next=false}
	\begin{tikzpicture}[yscale=-.4, xscale=.2]
		\node[n](a1) at (-2,0){};
		\node[n](a2) at (-1,0){};
		\node[n](a3) at (0,0){};
		\node[n](a4) at (1,0){};
		\node[n](a5) at (2,0){};
		
		\node[n](b1) at (-2,1){};
		\node[n](b2) at (-1,1){};
		\node[n](b3) at (0,1){};
		\node[n](b4) at (1,1){};
		\node[n](b5) at (2,1){};
		
		\draw[tumred,opacity=.05] (a1) -- (b1);
		\draw[tumred,opacity=.05] (a1) -- (b2);
		\draw[tumred,opacity=.05] (a1) -- (b3);
		\draw[tumred,opacity=.05] (a1) -- (b4);
		\draw[tumred,opacity=.05] (a1) -- (b5);
		\draw[tumred,opacity=.05] (a2) -- (b1);
		\draw[tumred,opacity=.05] (a2) -- (b2);
		\draw[tumred,opacity=.05] (a2) -- (b3);
		\draw[tumred,opacity=.05] (a2) -- (b4);
		\draw[tumred,opacity=.05] (a2) -- (b5);
		\draw[tumred,opacity=.05] (a3) -- (b1);
		\draw[tumred,opacity=.05] (a3) -- (b2);
		\draw[tumred,opacity=.05] (a3) -- (b3);
		\draw[tumred,opacity=.05] (a3) -- (b4);
		\draw[tumred,opacity=.05] (a3) -- (b5);
		
		\draw[tumred,opacity=.8] (a4) -- (b1);
		\draw[tumred,opacity=.8] (a4) -- (b2);
		\draw[tumred,opacity=.8] (a4) -- (b3);
		\draw[tumred,opacity=.8] (a4) -- (b4);
		\draw[tumred,opacity=.8] (a4) -- (b5);
		
		\draw[tumred,opacity=.05] (a5) -- (b1);
		\draw[tumred,opacity=.05] (a5) -- (b2);
		\draw[tumred,opacity=.05] (a5) -- (b3);
		\draw[tumred,opacity=.05] (a5) -- (b4);
		\draw[tumred,opacity=.05] (a5) -- (b5);
		
		\draw[tumbluedark,opacity=.8] (a1) -- (b1);
		\draw[tumbluedark,opacity=.8] (a1) -- (b2);
		\draw[tumbluedark,opacity=.8] (a1) -- (b3);
		\draw[tumbluedark,opacity=.8] (a1) -- (b4);
		\draw[tumbluedark,opacity=.8] (a1) -- (b5);
		
		\draw[tumbluedark,opacity=.05] (a2) -- (b1);
		\draw[tumbluedark,opacity=.05] (a2) -- (b2);
		\draw[tumbluedark,opacity=.05] (a2) -- (b3);
		\draw[tumbluedark,opacity=.05] (a2) -- (b4);
		\draw[tumbluedark,opacity=.05] (a2) -- (b5);
		\draw[tumbluedark,opacity=.05] (a3) -- (b1);
		\draw[tumbluedark,opacity=.05] (a3) -- (b2);
		\draw[tumbluedark,opacity=.05] (a3) -- (b3);
		\draw[tumbluedark,opacity=.05] (a3) -- (b4);
		\draw[tumbluedark,opacity=.05] (a3) -- (b5);
		\draw[tumbluedark,opacity=.05] (a4) -- (b1);
		\draw[tumbluedark,opacity=.05] (a4) -- (b2);
		\draw[tumbluedark,opacity=.05] (a4) -- (b3);
		\draw[tumbluedark,opacity=.05] (a4) -- (b4);
		\draw[tumbluedark,opacity=.05] (a4) -- (b5);
		\draw[tumbluedark,opacity=.05] (a5) -- (b1);
		\draw[tumbluedark,opacity=.05] (a5) -- (b2);
		\draw[tumbluedark,opacity=.05] (a5) -- (b3);
		\draw[tumbluedark,opacity=.05] (a5) -- (b4);
		\draw[tumbluedark,opacity=.05] (a5) -- (b5);
		
		
	\end{tikzpicture}
	}

	\newcommand{\ff}{
		\tikzset{external/export next=false}
		\begin{tikzpicture}[yscale=-.3, xscale=.2]
		\node[n](a1) at (-2,0){};
		\node[n](a2) at (-1,0){};
		\node[n](a3) at (0,0){};
		\node[n](a4) at (1,0){};
		\node[n](a5) at (2,0){};
		
		\node[n](b1) at (-2,1){};
		\node[n](b2) at (-1,1){};
		\node[n](b3) at (0,1){};
		\node[n](b4) at (1,1){};
		\node[n](b5) at (2,1){};
		
		\draw[tumbluelight,-stealth] (a1) -- (b1);
		\draw[tumbluelight,-stealth] (a2) -- (b2);
		\draw[tumbluelight,-stealth] (a3) -- (b3);
		\draw[tumbluelight,-stealth] (a4) -- (b4);
		\draw[tumbluelight,-stealth] (a5) -- (b5);
		
		\end{tikzpicture}
	}
	
	\newcommand{\block}{
		
	\tikzset{external/export next=false}
	\begin{tikzpicture}[xscale=.5, yscale=-.55]
		\node[inner sep=0](attn) at (0,0){\attn};
		\node[inner sep=0](ff) at (0,1){\ff};
	\end{tikzpicture}
	}
	\tikzset{external/export next=false}
	\begin{tikzpicture}[yscale=-1.2]
	\node(x) at (0,0){\M{X}};{}
	\node[inner sep=0, label={[font=\tiny\sffamily,text width=3em]left:self-attention \\\vspace{2em} feed forward}](l1) at (0,1.2){\block};
	\node[inner sep=0, label={[font=\tiny\sffamily,text width=3em]left:self-attention \\\vspace{2em} feed forward}](l2) at (0,2.4){\block};
	\node[font=\small\sffamily, inner sep=0](a) at (0,3.5){max-pooling}; 
	\node[inner sep=0,font=\small\sffamily](s) at (0,4){softmax$(\Mweight\T\M{H})$};
	\node[inner sep=0](y) at (0,4.5){\V{y}};
	
	\draw[-stealth, tumbluelight] (x) -- (l1);
	\draw[-stealth, tumbluelight] (l1) -- (l2);
	\draw[-stealth, tumbluelight] (l2) -- (a);
	\draw[-stealth, tumbluelight] (a) -- (s);
	\draw[-stealth, tumbluelight] (s) -- (y);
	\end{tikzpicture}

%% file: images/models/msresnet.tikz
	\tikzstyle{n} = [circle, inner sep=.8pt, fill=tumbluelight]
	\tikzstyle{k} = [circle, inner sep=.8pt, fill=tumred]
	\tikzstyle{o} = [circle, inner sep=.8pt, fill=tumbluedark]
	\tikzstyle{conv} = [draw=tumorange,opacity=.5]
	\tikzstyle{oconv} = [draw=tumbluedark,opacity=.5]
	
	\newcommand{\convthree}{
		\tikzset{external/export next=false}
		\begin{tikzpicture}[yscale=-.1, xscale=.1]
			\node[n](v0) at (0,0){};
			\node[n](v1) at (1,0){};
			\node[n](v2) at (2,0){};
			\node[n](v3) at (3,0){};
			\node[n](v4) at (4,0){};
			\node[n](v5) at (5,0){};
			\node[n](v6) at (6,0){};
			
			\node[k](k1) at (1,1){};
			\node[k](k2) at (2,1){};
			\node[k](k3) at (3,1){};
			
			\node[k](kk1) at (3,2){};
			\node[k](kk2) at (4,2){};
			\node[k](kk3) at (5,2){};
			
			
			\node[o](o0) at (0,3){};
			\node[o](o2) at (2,3){};
			\node[o](o4) at (4,3){};
			\node[o](o6) at (6,3){};
			
%
%
			
		\end{tikzpicture}
	}

	\newcommand{\convfive}{
		\tikzset{external/export next=false}
		\begin{tikzpicture}[yscale=-.1, xscale=.1]
		\node[n](v0) at (-1,0){};
		\node[n](v0) at (0,0){};
		\node[n](v1) at (1,0){};
		\node[n](v2) at (2,0){};
		\node[n](v3) at (3,0){};
		\node[n](v4) at (4,0){};
		\node[n](v5) at (5,0){};
		\node[n](v6) at (6,0){};
		\node[n](v7) at (7,0){};
		
		\node[k](k0) at (0,1){};
		\node[k](k1) at (1,1){};
		\node[k](k2) at (2,1){};
		\node[k](k3) at (3,1){};
		\node[k](k4) at (4,1){};
		
		\node[k](kk0) at (2,2){};
		\node[k](kk1) at (3,2){};
		\node[k](kk2) at (4,2){};
		\node[k](kk3) at (5,2){};
		\node[k](kk4) at (6,2){};
		
		
		\node[o](o0) at (0,3){};
		\node[o](o2) at (2,3){};
		\node[o](o4) at (4,3){};
		\node[o](o6) at (6,3){};
		
%
%
		
		\end{tikzpicture}
	}

	\newcommand{\convseven}{
		\tikzset{external/export next=false}
		\begin{tikzpicture}[yscale=-.1, xscale=.1]
		\node[n](v0) at (-2,0){};
		\node[n](v0) at (-1,0){};
		\node[n](v0) at (0,0){};
		\node[n](v1) at (1,0){};
		\node[n](v2) at (2,0){};
		\node[n](v3) at (3,0){};
		\node[n](v4) at (4,0){};
		\node[n](v5) at (5,0){};
		\node[n](v6) at (6,0){};
		\node[n](v7) at (7,0){};
		\node[n](v8) at (8,0){};
		\node[n](v9) at (9,0){};
		\node[n](v9) at (10,0){};
		
		\node[k](k6) at (-1,1){};
		\node[k](k0) at (0,1){};
		\node[k](k1) at (1,1){};
		\node[k](k2) at (2,1){};
		\node[k](k3) at (3,1){};
		\node[k](k4) at (4,1){};
		\node[k](k5) at (5,1){};
		
		\node[k](kk6) at (1,2){};
		\node[k](kk0) at (2,2){};
		\node[k](kk1) at (3,2){};
		\node[k](kk2) at (4,2){};
		\node[k](kk3) at (5,2){};
		\node[k](kk4) at (6,2){};
		\node[k](kk5) at (7,2){};

		
		\node[o](o0) at (0,3){};
		\node[o](o2) at (2,3){};
		\node[o](o4) at (4,3){};
		\node[o](o6) at (6,3){};
		\node[o](o8) at (8,3){};
		
		%
		%
		
		\end{tikzpicture}
	}

	\newcommand{\block}[1]{
		\tikzset{external/export next=false}
		\begin{tikzpicture}[node distance=0]
			\node(in){};
			\node[below=of in, inner sep=1pt](conv){#1};
			\node[below=of conv,font=\sffamily\tiny, inner sep=1pt](bn){BN+ReLU};
			\node[below=of bn,font=\sffamily\tiny, inner sep=1pt](p){$+$};
			\draw[-stealth, tumbluelight] (in) to[out=180, in=180, looseness=2.2] (p);
		\end{tikzpicture}
	}

	\newcommand{\stream}[1]{
		\tikzset{external/export next=false}
		\begin{tikzpicture}[yscale=-1]
			\node at (0,0){\block{#1}};
			\node at (0,1){\block{#1}};
			\node at (0,2){\block{#1}};
		\end{tikzpicture}
	}

	\tikzset{external/export next=false}
	\begin{tikzpicture}[yscale=-1.2]
		\node(x) at (0,0){\M{X}};{}
		
		\node[inner sep=0](n1) at (-1.5,1.8) {\stream{\convthree}};
		\node[inner sep=0](n2) at (-.3,1.8) {\stream{\convfive}};
		\node[inner sep=0](n3) at (1.1,1.8) {\stream{\convseven}};
		
		\node[font=\small\sffamily, inner sep=0](a) at (0,3.5){avg-pooling}; 
		\node[inner sep=0,font=\small\sffamily](s) at (0,4){softmax$(\Mweight\T\M{H})$};
		\node[inner sep=0](y) at (0,4.5){\V{y}};
		
		\coordinate(n1in) at ($ (n1.north)+(.3,0) $);
		\coordinate(n2in) at ($ (n2.north)+(.3,0) $);
		\coordinate(n3in) at ($ (n3.north)+(.2,0) $);
		
		\coordinate(n1out) at ($ (n1.south)+(.3,0) $);
		\coordinate(n2out) at ($ (n2.south)+(.3,0) $);
		\coordinate(n3out) at ($ (n3.south)+(.2,0) $);
		
		\draw[-stealth, tumbluelight, rounded corners] (x) |- ($ (x)!0.7!(n1in) $) -| (n1in);
		\draw[-stealth, tumbluelight, rounded corners] (x) |- ($ (x)!0.7!(n2in) $) -| (n2in);
		\draw[-stealth, tumbluelight, rounded corners] (x) |- ($ (x)!0.7!(n3in) $) -| (n3in);
		\draw[-stealth, tumbluelight, rounded corners] (n1out) |- ($ (n1out)!0.3!(a) $) -| (a);
		\draw[-stealth, tumbluelight, rounded corners] (n2out) |- ($ (n2out)!0.3!(a) $) -| (a);
		\draw[-stealth, tumbluelight, rounded corners] (n3out) |- ($ (n3out)!0.3!(a) $) -| (a);
		
		\draw[-stealth, tumbluelight] (a) -- (s);
		\draw[-stealth, tumbluelight] (s) -- (y);
	\end{tikzpicture}

%% file: images/models/tempcnn.tikz
	\tikzstyle{n} = [circle, inner sep=.8pt, fill=tumbluelight]
	\tikzstyle{k} = [circle, inner sep=.8pt, fill=tumred]
	\tikzstyle{o} = [circle, inner sep=.8pt, fill=tumbluedark]
	\tikzstyle{conv} = [draw=tumorange,opacity=.5]
	\tikzstyle{oconv} = [draw=tumbluedark,opacity=.5]
	
	\newcommand{\convthree}{
		\tikzset{external/export next=false}
		\begin{tikzpicture}[yscale=-.1, xscale=.1]
			\node[n](v0) at (0,0){};
			\node[n](v1) at (1,0){};
			\node[n](v2) at (2,0){};
			\node[n](v3) at (3,0){};
			\node[n](v4) at (4,0){};
			\node[n](v5) at (5,0){};
			\node[n](v6) at (6,0){};
			
			\node[k](k1) at (1,1){};
			\node[k](k2) at (2,1){};
			\node[k](k3) at (3,1){};
			
			\node[k](kk1) at (3,2){};
			\node[k](kk2) at (4,2){};
			\node[k](kk3) at (5,2){};
			
			
			\node[o](o0) at (0,3){};
			\node[o](o2) at (2,3){};
			\node[o](o4) at (4,3){};
			\node[o](o6) at (6,3){};
			
%
%
			
		\end{tikzpicture}
	}

	\newcommand{\convbnrelu}{
		\tikzset{external/export next=false}
		\begin{tikzpicture}[node distance=0]
			\node[below=of in, inner sep=1pt](conv){\convthree};
			\node[below=of conv,font=\sffamily\tiny, inner sep=1pt](bn){BN+ReLU};
		\end{tikzpicture}
	}

	\newcommand{\stream}[1]{
		\tikzset{external/export next=false}
		\begin{tikzpicture}[yscale=-1]
			\node(1) at (0,0){\convbnrelu};
			\node(2) at (0,1){\convbnrelu};
			\node(3) at (0,2){\convbnrelu};
			\draw[-stealth, tumbluelight] (1) -- (2);
			\draw[-stealth, tumbluelight] (2) -- (3);
		\end{tikzpicture}
	}

	\tikzset{external/export next=false}
	\begin{tikzpicture}[yscale=-1.2]
		\node(x) at (0,0){\M{X}};{}
		
		\node[inner sep=0](n) at (0,1.8) {\stream{\convseven}};
		
		\node[font=\small\sffamily, inner sep=0](a) at (0,3.5){flatten}; 
		\node[inner sep=0,font=\small\sffamily](s) at (0,4){softmax$(\Mweight\T\M{H})$};
		\node[inner sep=0](y) at (0,4.5){\V{y}};
		
		\draw[-stealth, tumbluelight] (x) -- (n);
		\draw[-stealth, tumbluelight] (n) -- (a);
		\draw[-stealth, tumbluelight] (a) -- (s);
		\draw[-stealth, tumbluelight] (s) -- (y);
	\end{tikzpicture}

%% file: images/areaofinterest.tikz
\tikzsetnextfilename{areaofinterest}
\begin{tikzpicture}[node distance=.5em]

\newcommand{\submap}[2]{
	\tikzset{external/export next=false}
	\begin{tikzpicture}
	\node[map, label={[font=\tiny\sffamily,anchor=north east,shift={(holl.south east)}]train-test split}](holl){#1};
	\node[xshift=-.5cm, yshift=-.5cm, map, label={[font=\tiny\sffamily,anchor=south west,shift={(hollp.north west)}]parcels}](hollp) at (holl){#2};	
	\end{tikzpicture}
}

\node[](map){\includegraphics[width=7cm]{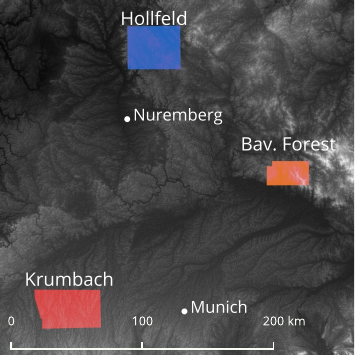}};
\node[left=of map, yshift=2.5cm, rounded corners, label={[font=\sffamily,anchor=south east,shift={(hollmap.north east)}]Hollfeld}, draw=tumblue, fill=tumblue!10](hollmap){\submap{\includegraphics[height=1.5cm]{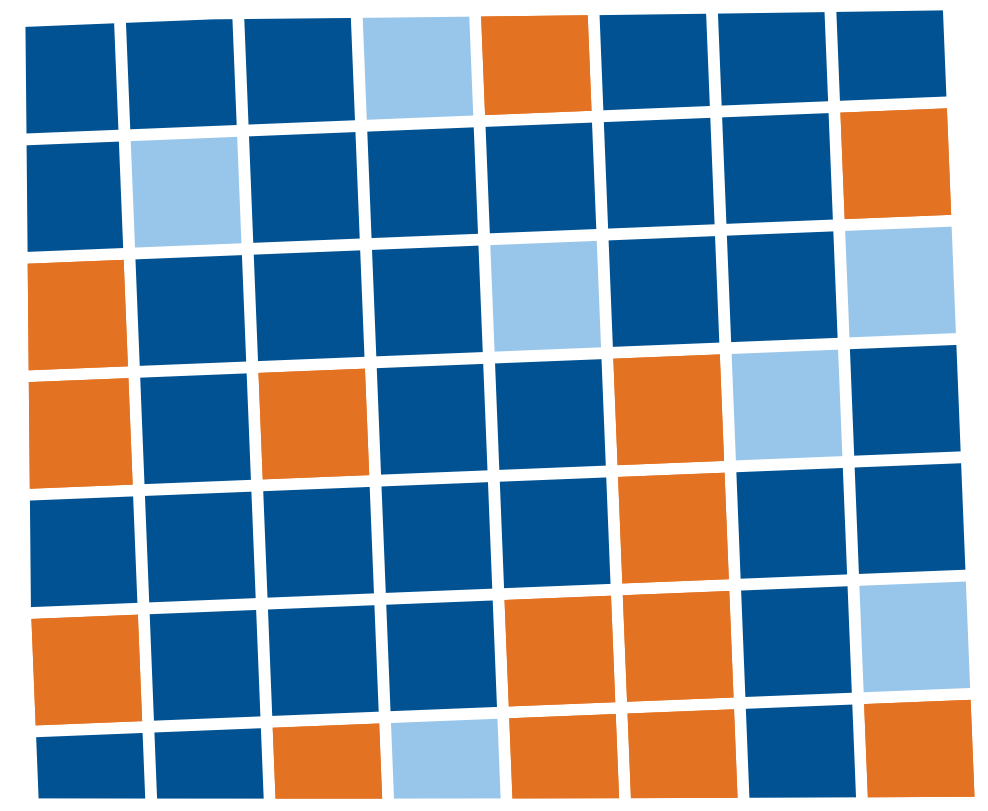}}{\includegraphics[height=1.5cm]{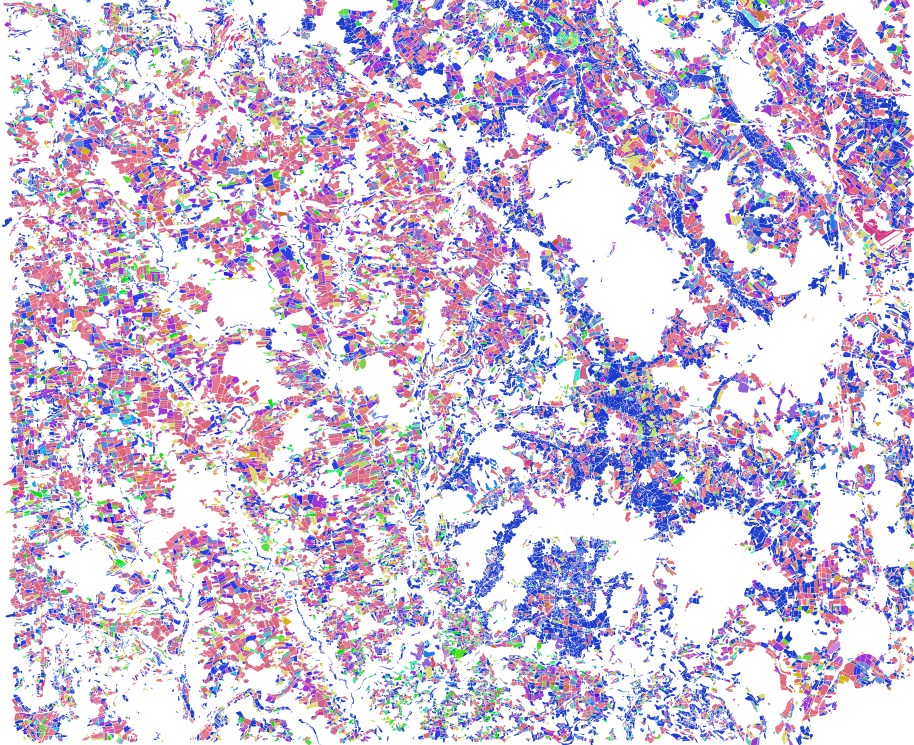}}};
	
	\node[left=of map, yshift=-.5cm, rounded corners, , label={[font=\sffamily,anchor=south east,shift={(krummap.north east)}]Krumbach}, draw=tumred, fill=tumred!10](krummap){\submap{\includegraphics[height=1.5cm]{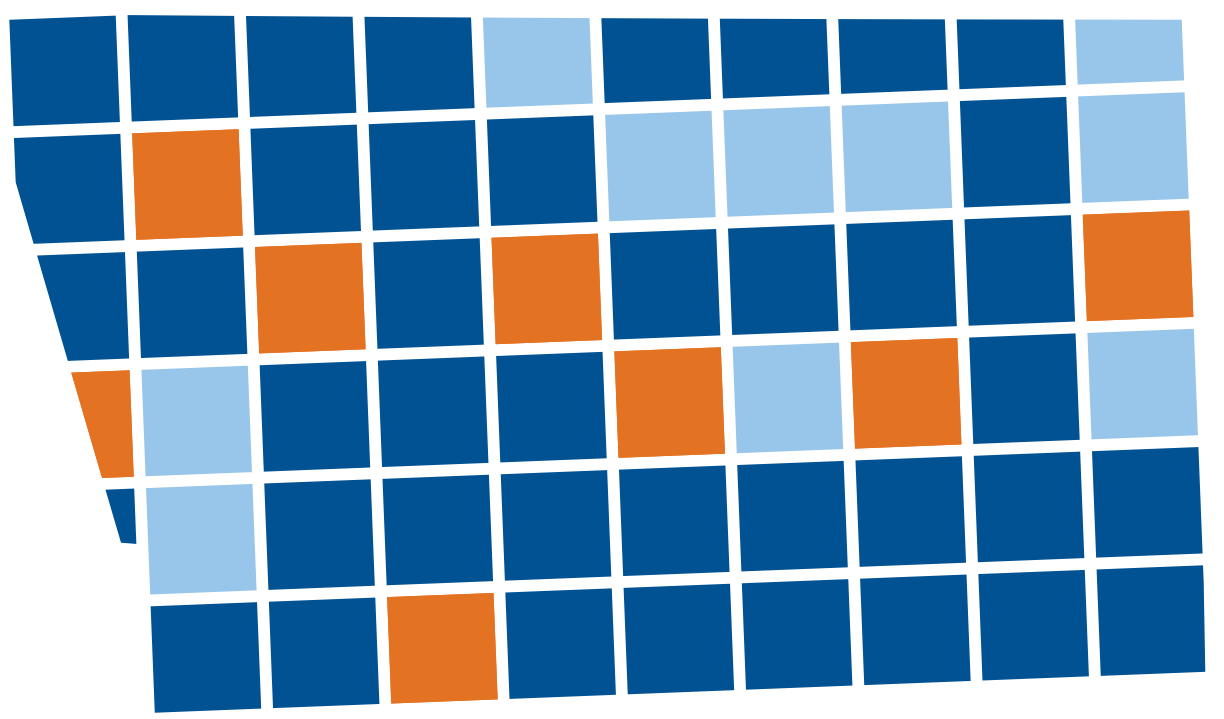}}{\includegraphics[height=1.5cm]{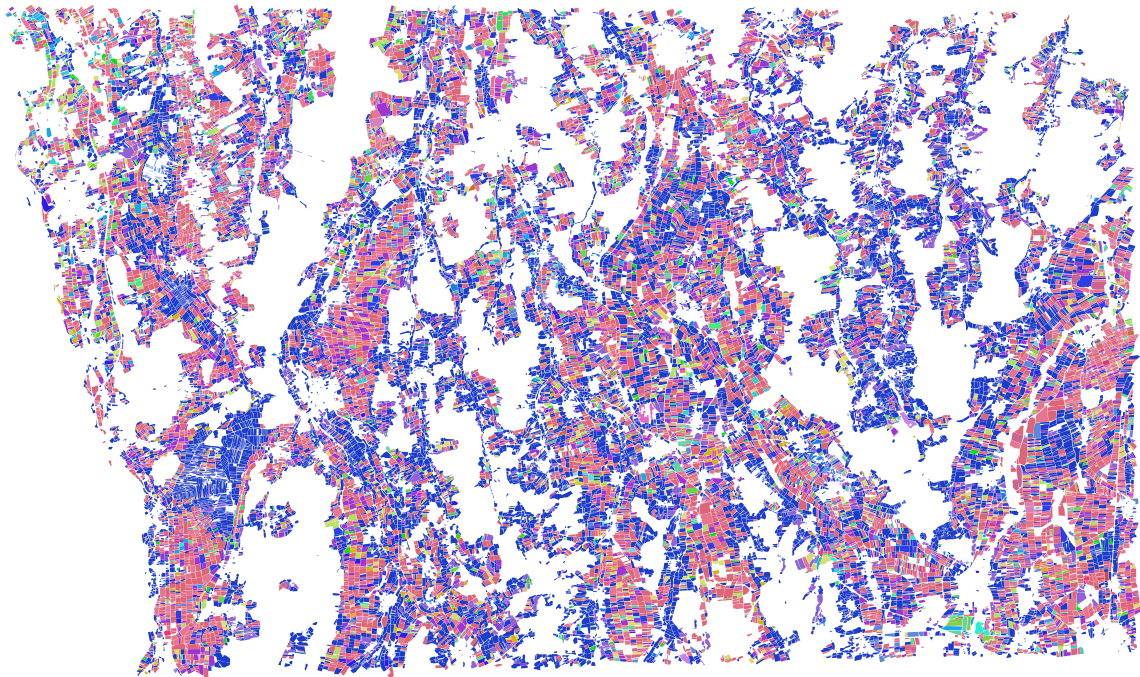}}};
	
	\node[left=of map, yshift=-3.5cm, rounded corners, label={[font=\sffamily,anchor=south east,shift={(nowamap.north east)}]Bavarian Forest}, draw=tumorange, fill=tumorange!10](nowamap){\submap{\includegraphics[height=1.5cm]{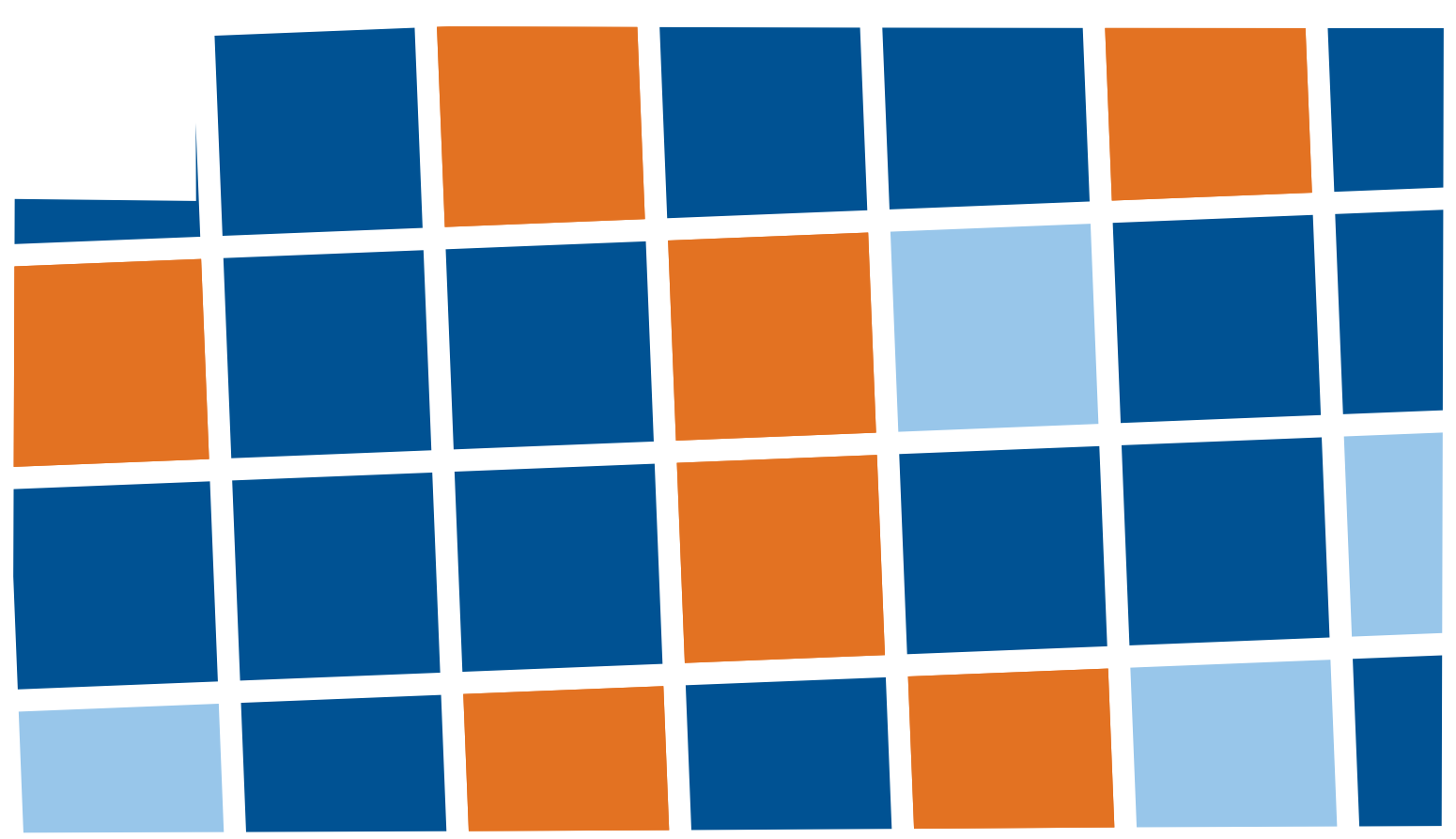}}{\includegraphics[height=1.5cm]{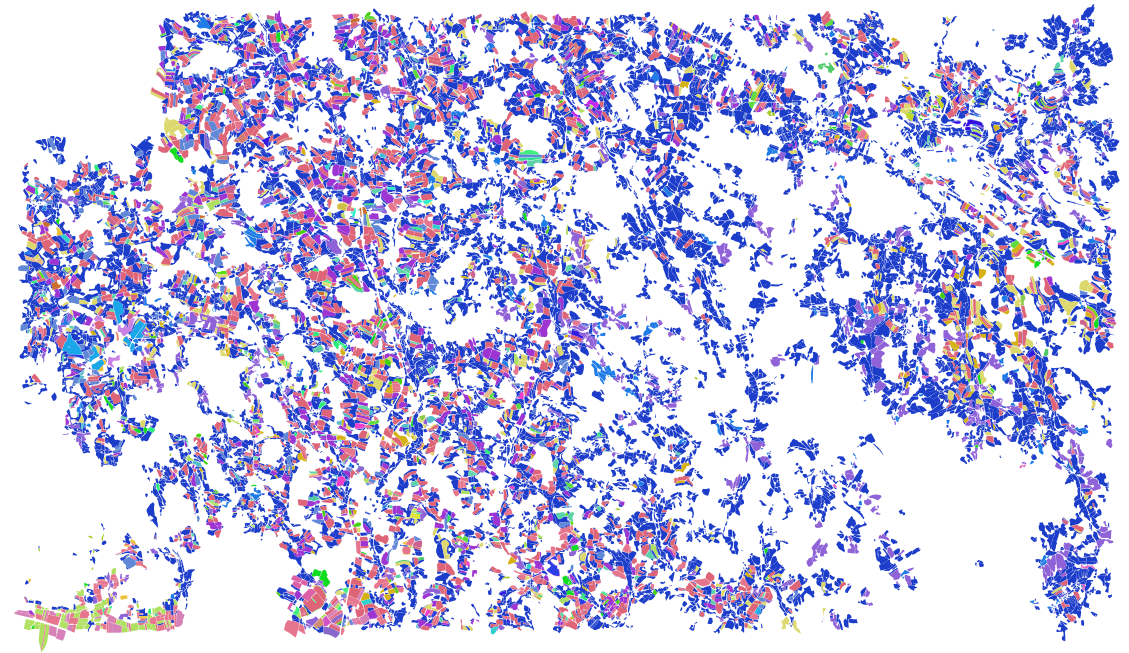}}};
	
	\node [below=.2em of map, shading = axis,rectangle, left color=white, right color=black,shading angle=90, anchor=north, minimum width=7cm, minimum height=4mm, draw=black](cbar){};
	\node[yshift=-1em, font=\small, anchor=west, inner sep=0] at (cbar.south west){650m};
	\node[yshift=-1em, font=\small, inner sep=0] at (cbar.south){450m};
	\node[yshift=-1em, font=\small, anchor=east, inner sep=0] at (cbar.south east){250m};
	
\end{tikzpicture}

%% file: images/preprocessed_example.tikz
\newcommand{\preexamplemeadow}{

\tikzsetnextfilename{preexamplemeadow}
\begin{tikzpicture}
		\begin{axis}[
		thin,
		width=.5\textwidth,
		enlarge x limits=0.1,
		enlarge y limits=0,
		date coordinates in=x,
		height=3.5cm,
		ymin=0, ymax=1.4,
		draw opacity=.8,
		xtick={2018-02-01,2018-05-01,2018-08-01,2018-11-01},
		xticklabels={2018-01,2018-05,2018-08,2018-11},
		smooth=0.01,
		xlabel={time $t$},
		ylabel={reflectance},
		grid style={line width=.1pt, draw=gray!10},
		legend style={at={(1,1.5)},thick, line width=2pt, draw opacity=1, font=\tiny\sffamily},
		legend image post style={line width =1pt},
		legend columns=4,
		y label style={at={(-0.1,0.5)}},
		]
		
		\def\sample{images/example/used/12-71456800.csv}

		\addplot[b2color, mark=*,mark size=.5pt] table [x=doa, y=B02, col sep=comma] {\sample};
		\addplot[b3color, mark=*,mark size=.5pt] table [x=doa, y=B03, col sep=comma] {\sample};
		\addplot[b4color, mark=*,mark size=.5pt] table [x=doa, y=B04, col sep=comma] {\sample};
		
		\addplot[b5color, mark=*,mark size=.5pt] table [x=doa, y=B05, col sep=comma] {\sample};
		\addplot[b6color, mark=*,mark size=.5pt] table [x=doa, y=B06, col sep=comma] {\sample};
		\addplot[b7color, mark=*,mark size=.5pt] table [x=doa, y=B07, col sep=comma] {\sample};
		\addplot[b8color, mark=*,mark size=.5pt] table [x=doa, y=B08, col sep=comma] {\sample};
		\addplot[b8Acolor, mark=*,mark size=.5pt] table [x=doa, y=B8A, col sep=comma] {\sample};
		
		\addplot[b11color, mark=*,mark size=.5pt] table [x=doa, y=B11, col sep=comma] {\sample};
		\addplot[b12color, mark=*,mark size=.5pt] table [x=doa, y=B12, col sep=comma] {\sample};

		\node[inner sep=.5em, font=\sffamily\tiny, text=tumblue](annotgrowth) at (axis cs:2018-02-09
		,1.2){onset growth};
		\coordinate(g1) at (axis cs:2018-04-01
		,.3);
		\draw[thin, draw=tumblue] (annotgrowth) -- (g1);
		
		\node[inner sep=.5em, font=\sffamily\tiny, text=tumblue](annotcut) at (axis cs:2018-09-14
		,1.2){meadow cutting};
		\coordinate(c1) at (axis cs:2018-07-12
		,.25);
		\draw[thin, draw=tumblue] (annotcut) -- (c1);
		\coordinate(c2) at (axis cs:2018-08-29
		,.3);
		\draw[thin, draw=tumblue] (annotcut) -- (c2);
		\coordinate(c3) at (axis cs:2018-10-16
		,.3);
		\draw[thin, draw=tumblue] (annotcut) -- (c3);
		\legend{B02 (blue),B03 (green),B04 (red),B05,B06,B07,B08,B8A,B11,B12}
		

		\end{axis}
		
\end{tikzpicture}	
}

\newcommand{\preexamplecorn}{

\tikzsetnextfilename{preexamplecorn}
\begin{tikzpicture}

\begin{axis}[
thin,
width=\textwidth,
enlarge x limits=0.1,
enlarge y limits=0,
height=3.5cm,
ymin=0, ymax=1.4,
draw opacity=.8,
smooth=0.01,
xlabel={time $t$},
ylabel={reflectance},
grid style={line width=.1pt, draw=gray!10}
]

\addplot[b11color, mark=*,mark size=.5pt] table [x=doa, y=B11, col sep=comma, forget plot] {images/example/\secondid.csv};
\addplot[b12color, mark=*,mark size=.5pt] table [x=doa, y=B12, col sep=comma] {images/example/\secondid.csv};

\addplot[b5color, mark=*,mark size=.5pt] table [x=doa, y=B05, col sep=comma, forget plot] {images/example/\secondid.csv};
\addplot[b6color, mark=*,mark size=.5pt] table [x=doa, y=B06, col sep=comma, forget plot] {images/example/\secondid.csv};
\addplot[b7color, mark=*,mark size=.5pt] table [x=doa, y=B07, col sep=comma, forget plot] {images/example/\secondid.csv};
\addplot[b8color, mark=*,mark size=.5pt] table [x=doa, y=B08, col sep=comma, forget plot] {images/example/\secondid.csv};
\addplot[b8Acolor, mark=*,mark size=.5pt] table [x=doa, y=B8A, col sep=comma] {images/example/\secondid.csv};

\addplot[b2color, mark=*,mark size=.5pt] table [x=doa, y=B02, col sep=comma, forget plot] {images/example/\secondid.csv};
\addplot[b3color, mark=*,mark size=.5pt] table [x=doa, y=B03, col sep=comma, forget plot] {images/example/\secondid.csv};
\addplot[b4color, mark=*,mark size=.5pt] table [x=doa, y=B04, col sep=comma] {images/example/\secondid.csv};

%


\end{axis}

\end{tikzpicture}	

}

%% file: images/raw_example.tikz
\newcommand{\rawexamplemeadow}{

\tikzsetnextfilename{rawexamplemeadow}
		\begin{tikzpicture}
		\def\sample{images/example/used/71456800_raw.csv}

		\begin{axis}[
			thin,
			width=.5\textwidth,
			enlarge x limits=0.1,
			enlarge y limits=0,
			height=3.5cm,
			date coordinates in=x,
			ymin=0, ymax=1.4,
			draw opacity=.8,
			xtick={2018-02-01,2018-05-01,2018-08-01,2018-11-01},
			xticklabels={2018-01,2018-05,2018-08,2018-11},
			smooth=0.01,
			xlabel={time $t$},
			ylabel={reflectance},
			grid style={line width=.1pt, draw=gray!10},
			legend style={at={(1,1.5)},thick, line width=2pt, draw opacity=1, font=\tiny\sffamily},
			legend image post style={line width =1pt},
			legend columns=4,
			y label style={at={(-0.1,0.5)}},
			]
			
			\addplot[b2color, mark=*,mark size=.5pt] table [x=doa, y=B2, col sep=comma] {\sample};
			\addplot[b3color, mark=*,mark size=.5pt] table [x=doa, y=B3, col sep=comma] {\sample};
			\addplot[b4color, mark=*,mark size=.5pt] table [x=doa, y=B4, col sep=comma] {\sample};

			
			\addplot[b5color, mark=*,mark size=.5pt] table [x=doa, y=B5, col sep=comma] {\sample};
			\addplot[b6color, mark=*,mark size=.5pt] table [x=doa, y=B6, col sep=comma] {\sample};
			\addplot[b7color, mark=*,mark size=.5pt] table [x=doa, y=B7, col sep=comma] {\sample};
			\addplot[b8color, mark=*,mark size=.5pt] table [x=doa, y=B8, col sep=comma] {\sample};
			\addplot[b8Acolor, mark=*,mark size=.5pt] table [x=doa, y=B8A, col sep=comma] {\sample};
			
			\addplot[b11color, mark=*,mark size=.5pt] table [x=doa, y=B11, col sep=comma] {\sample};
			\addplot[b12color, mark=*,mark size=.5pt] table [x=doa, y=B12, col sep=comma] {\sample};

			\node[inner sep=.5em, font=\sffamily\tiny, text=tumblue](annotground) at (axis cs:2018-08-25
			,1.2){ground};
			\coordinate(g6) at (axis cs:2018-04-05,.3);
			\coordinate(g7) at (axis cs:2018-09-20,.3);
			\coordinate(g8) at (axis cs:2018-07-25,.3);
			\coordinate(g9) at (axis cs:2018-08-20,.3);
			
			\draw[thin, tumblue] (annotground) -- (g6);
			\draw[thin, tumblue] (annotground) -- (g7);
			\draw[thin, tumblue] (annotground) -- (g8);
			\draw[thin, tumblue] (annotground) -- (g9);
			
			
			\coordinate(c8) at (axis cs:2018-04-17,1.1);
			\coordinate(c9) at (axis cs:2018-08-30,.8);
			\coordinate(c7) at (axis cs:2018-07-21,.9);
			
			\node[inner sep=.5em, font=\sffamily\tiny, text=tumblue](annotclouds) at (axis cs:2018-06-11,1.2){clouds};
			\draw[thin, draw=tumblue] (annotclouds) -- (c7);
			\draw[thin, draw=tumblue] (annotclouds) -- (c8);
			\draw[thin, draw=tumblue] (annotclouds) -- (c9);
%
			\legend{B02 (blue),B03 (green),B04 (red),B05,B06,B07,B08,B8A,B11,B12}
			
		\end{axis}
		
		\end{tikzpicture}	
		
}
\newcommand{\rawexamplecorn}{

\tikzsetnextfilename{rawexamplecorn}
\begin{tikzpicture}

\begin{axis}[
thin,
width=\textwidth,
enlarge x limits=0.1,
enlarge y limits=0,
height=3.5cm,
ymin=0, ymax=1.4,
draw opacity=.8,
smooth=0.01,
xlabel={time $t$},
ylabel={reflectance},
grid style={line width=.1pt, draw=gray!10}
]

\addplot[b11color, mark=*,mark size=.5pt] table [x=doa, y=B11, col sep=comma, forget plot] {images/example/\secondid_raw.csv};
\addplot[b12color, mark=*,mark size=.5pt] table [x=doa, y=B12, col sep=comma] {images/example/\secondid_raw.csv};

\addplot[b5color, mark=*,mark size=.5pt] table [x=doa, y=B05, col sep=comma, forget plot] {images/example/\secondid_raw.csv};
\addplot[b6color, mark=*,mark size=.5pt] table [x=doa, y=B06, col sep=comma, forget plot] {images/example/\secondid_raw.csv};
\addplot[b7color, mark=*,mark size=.5pt] table [x=doa, y=B07, col sep=comma, forget plot] {images/example/\secondid_raw.csv};
\addplot[b8color, mark=*,mark size=.5pt] table [x=doa, y=B08, col sep=comma, forget plot] {images/example/\secondid_raw.csv};
\addplot[b8Acolor, mark=*,mark size=.5pt] table [x=doa, y=B8A, col sep=comma] {images/example/\secondid_raw.csv};

\addplot[b2color, mark=*,mark size=.5pt] table [x=doa, y=B02, col sep=comma, forget plot] {images/example/\secondid_raw.csv};
\addplot[b3color, mark=*,mark size=.5pt] table [x=doa, y=B03, col sep=comma, forget plot] {images/example/\secondid_raw.csv};
\addplot[b4color, mark=*,mark size=.5pt] table [x=doa, y=B04, col sep=comma] {images/example/\secondid_raw.csv};

%
%
%


\end{axis}

\end{tikzpicture}	

}

%% file: tables/preraw/kappa.tex
\begin{tabular}{lcccccc}
\toprule
$\kappa$ & RF & \rnn & \transformer & \new{\duplo} & \msresnet & \tempcnn \\
\cmidrule(lr){2-2}\cmidrule(lr){3-3}\cmidrule(lr){4-4}\cmidrule(lr){5-5}\cmidrule(lr){6-6}\cmidrule(lr){7-7}
    	pre & 0.76 & 0.78$^{\pm0.01}$ & 0.79$^{\pm0.03}$ & \new{\textbf{0.81}} & 0.76$^{\pm0.03}$ & 0.80$^{\pm0.00}$ \\ 
	raw       & 0.53 & \textbf{0.71}$^{\pm0.01}$ & \textbf{0.71}$^{\pm0.03}$ & \new{0.68} & 0.69$^{\pm0.03}$ & 0.67$^{\pm0.02}$ \\ 
\bottomrule
\end{tabular}

%% file: tables/preraw2/kappa.tex
\begin{tabular}{lcccccc}
\toprule
$\kappa$ & RF & \rnn & \transformer & \new{\duplo} & \msresnet & \tempcnn \\
\cmidrule(lr){2-2}\cmidrule(lr){3-3}\cmidrule(lr){4-4}\cmidrule(lr){5-5}\cmidrule(lr){6-6}\cmidrule(lr){7-7}
    	pre & 0.86 & 0.87$^{\pm0.01}$ & 0.87$^{\pm0.04}$ & \new{\textbf{0.89}} & 0.85$^{\pm0.01}$ & 0.88$^{\pm0.03}$ \\ 
	raw       & 0.62 & \textbf{0.83}$^{\pm0.01}$ & 0.82$^{\pm0.02}$ & \new{0.78} & 0.78$^{\pm0.02}$ & 0.71$^{\pm0.07}$ \\ 
\bottomrule
\end{tabular}

%% file: tables/preraw/accuracy.tex
\begin{tabular}{lcccccc}
\toprule
acc. & RF & \rnn & \transformer & \new{\duplo} & \msresnet & \tempcnn \\
\cmidrule(lr){2-2}\cmidrule(lr){3-3}\cmidrule(lr){4-4}\cmidrule(lr){5-5}\cmidrule(lr){6-6}\cmidrule(lr){7-7}
    	pre & 0.83 & 0.85$^{\pm0.01}$ & 0.85$^{\pm0.02}$ & \new{\textbf{0.86}} & 0.83$^{\pm0.02}$ & \textbf{0.86}$^{\pm0.00}$ \\ 
	raw       & 0.71 & \textbf{0.81}$^{\pm0.01}$ & 0.80$^{\pm0.02}$ & \new{0.79} & 0.79$^{\pm0.03}$ & 0.79$^{\pm0.00}$ \\ 
\bottomrule
\end{tabular}

%% file: tables/preraw2/accuracy.tex
\begin{tabular}{lcccccc}
\toprule
acc. & RF & \rnn & \transformer & \new{\duplo} & \msresnet & \tempcnn \\
\cmidrule(lr){2-2}\cmidrule(lr){3-3}\cmidrule(lr){4-4}\cmidrule(lr){5-5}\cmidrule(lr){6-6}\cmidrule(lr){7-7}
    	pre & 0.91 & \textbf{0.92}$^{\pm0.01}$ & \textbf{0.92}$^{\pm0.03}$ & \new{\textbf{0.92}} & 0.91$^{\pm0.01}$ & \textbf{0.92}$^{\pm0.02}$ \\ 
	raw       & 0.80 & \textbf{0.90}$^{\pm0.00}$ & 0.89$^{\pm0.01}$ & \new{0.87} & 0.87$^{\pm0.01}$ & 0.83$^{\pm0.04}$ \\ 
\bottomrule
\end{tabular}

%% file: tables/preraw/f1.tex
\begin{tabular}{lcccccc}
\toprule
f1 & RF & \rnn & \transformer & \new{\duplo} & \msresnet & \tempcnn \\
\cmidrule(lr){2-2}\cmidrule(lr){3-3}\cmidrule(lr){4-4}\cmidrule(lr){5-5}\cmidrule(lr){6-6}\cmidrule(lr){7-7}
    	pre & 0.38 & 0.47$^{\pm0.02}$ & 0.50$^{\pm0.06}$ & \new{\textbf{0.55}} & 0.49$^{\pm0.03}$ & 0.50$^{\pm0.02}$ \\ 
	raw       & 0.18 & 0.43$^{\pm0.01}$ & \textbf{0.45}$^{\pm0.06}$ & \new{0.34} & 0.44$^{\pm0.01}$ & 0.36$^{\pm0.01}$ \\ 
\bottomrule
\end{tabular}

%% file: tables/preraw2/f1.tex
\begin{tabular}{lcccccc}
\toprule
f1 & RF & \rnn & \transformer & \new{\duplo} & \msresnet & \tempcnn \\
\cmidrule(lr){2-2}\cmidrule(lr){3-3}\cmidrule(lr){4-4}\cmidrule(lr){5-5}\cmidrule(lr){6-6}\cmidrule(lr){7-7}
    	pre & 0.55 & 0.60$^{\pm0.02}$ & \textbf{0.66}$^{\pm0.07}$ & \new{\textbf{0.66}} & 0.64$^{\pm0.02}$ & 0.60$^{\pm0.06}$ \\ 
	raw       & 0.34 & 0.63$^{\pm0.01}$ & \textbf{0.64}$^{\pm0.07}$ & \new{0.51} & 0.55$^{\pm0.04}$ & 0.41$^{\pm0.07}$ \\ 
\bottomrule
\end{tabular}

%% file: images/qualitative.tikz
\newcommand{\labellegend}{
	\begin{tikzpicture}[xscale=1.2]
		\tikzstyle{legendelement} = [minimum width=2.5em, minimum height=.5em, rounded corners=1pt]
		\tikzstyle{legendlabelelement} = [font=\tiny\sffamily,text height=1.5ex, text depth=0.25ex, label distance=0em]
		\node[legendelement, fill=tumblack,label={[legendlabelelement]below:fallow}] at (0,0){};	
		\node[legendelement, fill=tumgreen,label={[legendlabelelement]below:grassland}] at (1,0){};
		\node[legendelement, fill=tumredorange,label={[legendlabelelement]below:winter wheat}] at (2,0){};
		\node[legendelement, fill=tumred,label={[legendlabelelement]below:corn}] at (3,0){};
		\node[legendelement, fill=tumorange,label={[legendlabelelement]below:summer wheat}] at (4,0){};
		\node[legendelement, fill=tumbluedark,label={[legendlabelelement]below:winter spelt}] at (5,0){};
		\node[legendelement, fill=tumivory,label={[legendlabelelement]below:winter rye}] at (6,0){};
		\node[legendelement, fill=tumlimegreen,label={[legendlabelelement]below:winter barley}] at (7,0){};
		\node[legendelement, fill=tumturquoise,label={[legendlabelelement]below:summer barley}] at (8,0){};
		\node[legendelement, fill=tumbluelight,label={[legendlabelelement]below:summer oat}] at (9,0){};
		\node[legendelement, fill=tumdarkred,label={[legendlabelelement]below:winter triticale}] at (10,0){};
		\node[legendelement, fill=tumsand,label={[legendlabelelement]below:rapeseed}] at (11,0){};
	\end{tikzpicture}
	}

	\newcommand{\modellegend}{
		\definecolor{correctcolor}{rgb}{0,0.5,0}
		\definecolor{wrongcolor}{rgb}{1,0,0}
		\begin{tikzpicture}[xscale=.8]
		\tikzstyle{legendelement} = [minimum width=2em, minimum height=.5em, rounded corners=1pt]
		\tikzstyle{legendlabelelement} = [font=\tiny\sffamily,text height=1.5ex, text depth=0.25ex, label distance=0em]
		\node[legendelement, fill=correctcolor,label={[legendlabelelement]below:correct}] at (0,0){};	
		\node[legendelement, fill=wrongcolor,label={[legendlabelelement]below:wrong}] at (1,0){};
		\end{tikzpicture}
	}

	\begin{tikzpicture}[xscale=3.25]
		\tikzstyle{legendlabelelement} = [font=\small\sffamily, label distance=0em]
		
		\node[anchor=south west, label={[legendlabelelement]above:Labels}](first) at (0,0){\includegraphics[width=3.2cm]{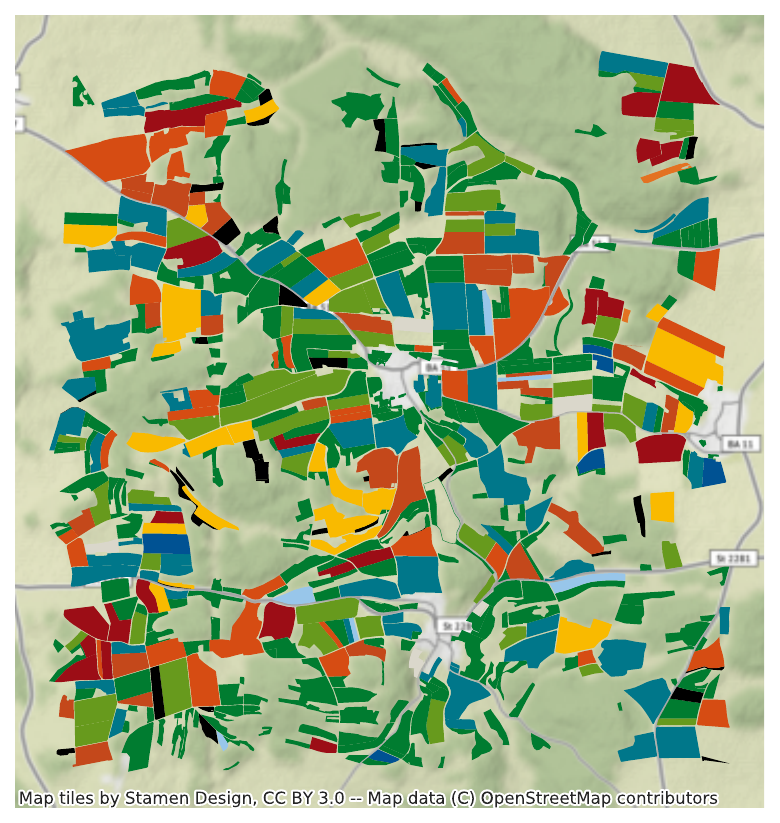}};
		\node[anchor=south west, label={[legendlabelelement]above:LSTM-RNN}] at (1,0){\includegraphics[width=3.2cm]{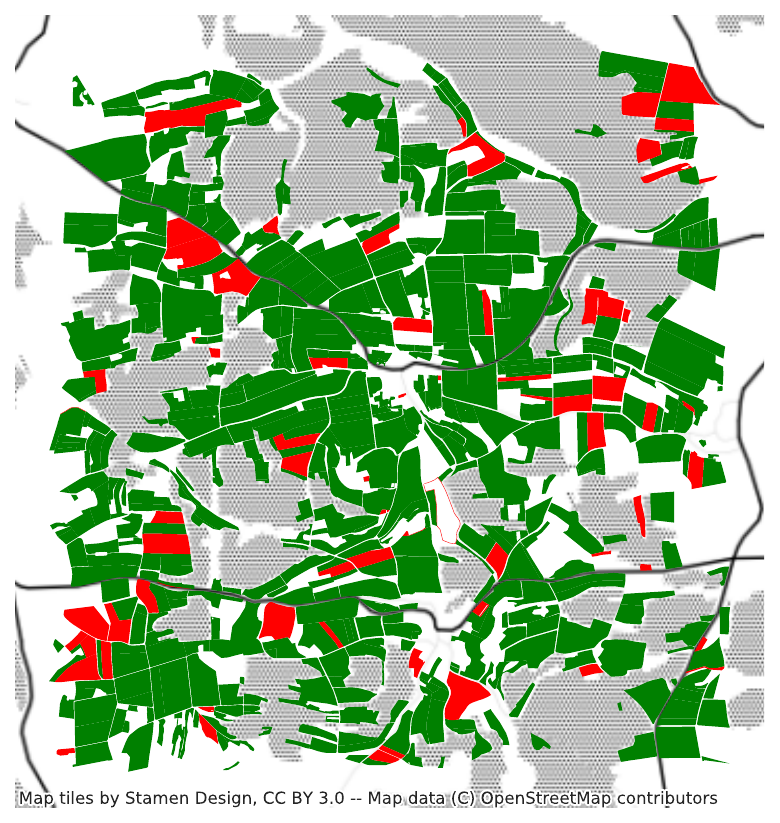}};
		\node[anchor=south west, label={[legendlabelelement]above:Transformer}] at (2,0){\includegraphics[width=3.2cm]{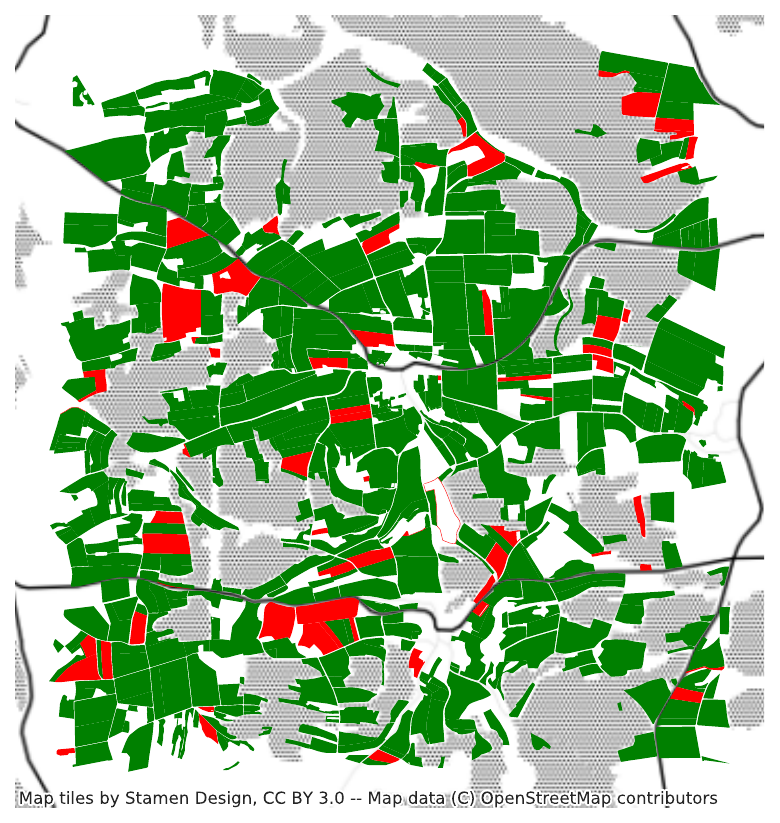}};
		\node[anchor=south west, label={[legendlabelelement]above:MSResNet}] at (3,0){\includegraphics[width=3.2cm]{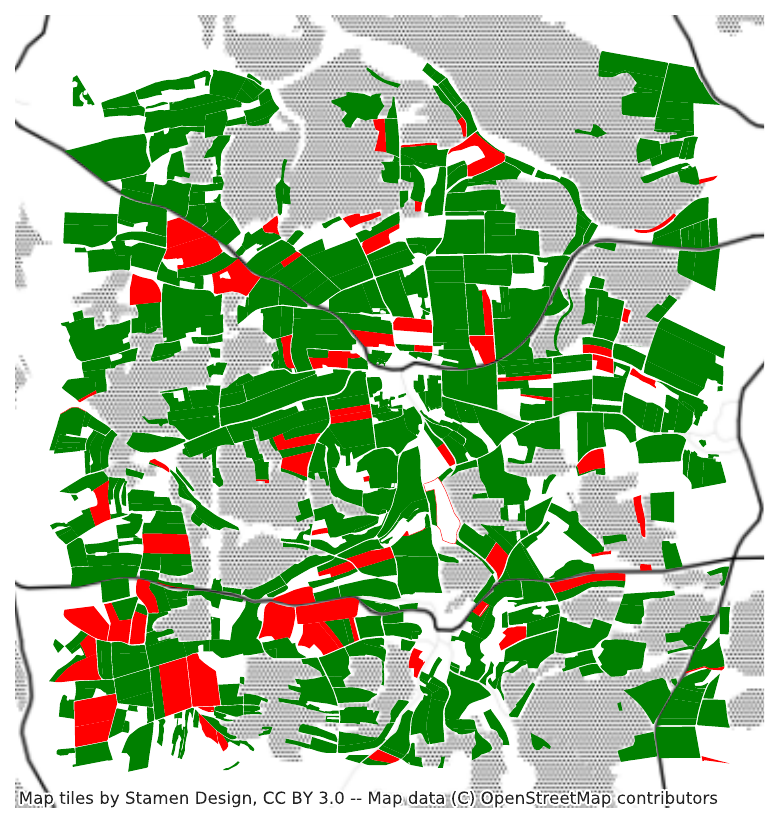}};
		\node[anchor=south west, label={[legendlabelelement]above:TempCNN}](last) at (4,0){\includegraphics[width=3.2cm]{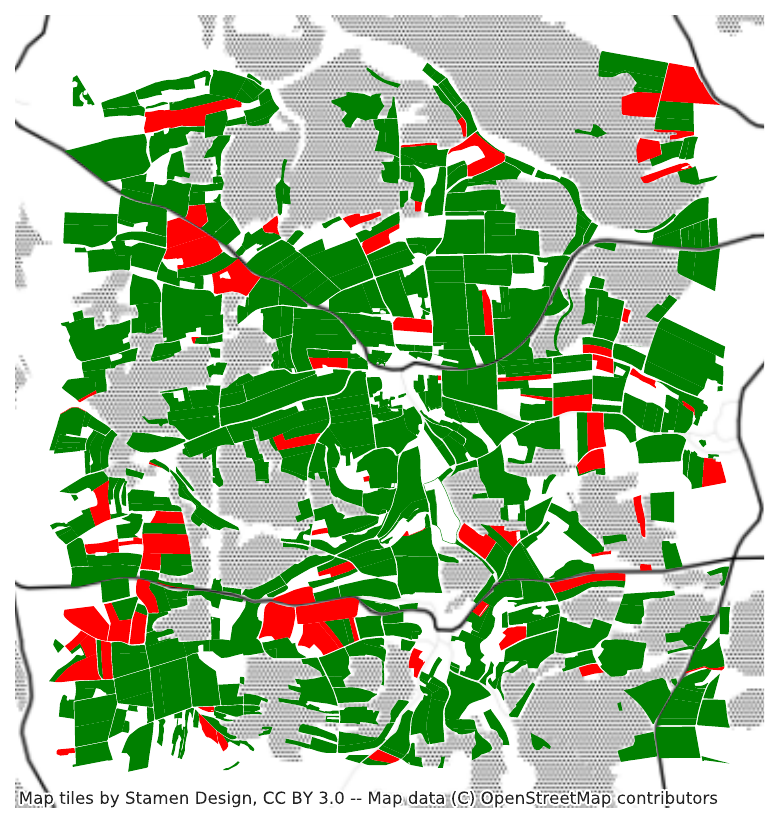}};
		
		\node[anchor=north west] at (first.south west)(labellegend){\labellegend};
		
		\node[anchor=north east] at (last.south east){\modellegend};
	\end{tikzpicture}

%% file: images/backprop.tikz
\tikzstyle{annot} = [inner sep=0, tumblue, font=\tiny, fill=white, fill opacity=.5, text opacity=1, rounded corners]
\tikzstyle{anno} = [-stealth, tumblue, thin]

\tikzstyle{annotg} = [inner sep=0, tumorange, font=\tiny, fill=white, fill opacity=.5, text opacity=1, rounded corners]
\tikzstyle{annog} = [-stealth, tumorange, thin]

\newcommand{\backpropexamplecorn}{
	
	\tikzsetnextfilename{rnn_gradients_71460046}
	\begin{tikzpicture}
	
	\def\meta{71460046}
	
	\tikzstyle{mymark}=[thin,mark=*,mark size=.5pt]
	
	\begin{groupplot}[
	group style = {
		group size = 1 by 5,
		xlabels at=edge bottom,
		xticklabels at=edge bottom,
		vertical sep=0pt,
	},
	width=.5\textwidth,
	enlarge x limits=0.0005,
	enlarge y limits=0.05,
	height=3cm,
	xlabel=observation time $t$,
	legend style={at={(1,1.8)},thick, line width=2pt, draw opacity=1, font=\tiny\sffamily},
	legend image post style={line width =1pt},
	legend columns=4,
	label style={font=\scriptsize},
	tick label style={font=\scriptsize},
  y label style={%
    at={(-0.15,0.5)}, 
    rotate=-90,
    anchor=center,
  },
	ylabel=$\frac{\partial y^\ast}{\partial \M{X}}$,
	title style={
		at={(0,-.2)},
		font=\sffamily\scriptsize,
		anchor=south west,
		fill=white,
		fill opacity=0.3,
		text opacity=1,
		rounded corners,
	},    
	]
	\nextgroupplot[draw opacity=.8, smooth=0.01, ylabel=reflectance $\M{X}$, y label style={rotate=90, anchor=center}, title={Sentinel 2 Time Series}, ymin=-.15,
	]
	
	\addplot[b2color, mymark] table [x=t, y=B2, col sep=comma] {images/backprop/rnn/\meta/x.csv};
	\addplot[b3color, mymark] table [x=t, y=B3, col sep=comma] {images/backprop/rnn/\meta/x.csv};
	\addplot[b4color, mymark] table [x=t, y=B4, col sep=comma] {images/backprop/rnn/\meta/x.csv};
	
	\addplot[b5color, mymark] table [x=t, y=B5, col sep=comma] {images/backprop/rnn/\meta/x.csv};
	\addplot[b6color, mymark] table [x=t, y=B6, col sep=comma] {images/backprop/rnn/\meta/x.csv};
	\addplot[b7color, mymark] table [x=t, y=B7, col sep=comma] {images/backprop/rnn/\meta/x.csv};
	\addplot[b8color, mymark] table [x=t, y=B8, col sep=comma] {images/backprop/rnn/\meta/x.csv};
	\addplot[b8Acolor, mymark] table [x=t, y=B8A, col sep=comma] {images/backprop/rnn/\meta/x.csv};
	
	\addplot[b11color, mymark] table [x=t, y=B11, col sep=comma] {images/backprop/rnn/\meta/x.csv};
	\addplot[b12color, mymark] table [x=t, y=B12, col sep=comma] {images/backprop/rnn/\meta/x.csv};
	
	\legend{B02 (blue),B03 (green),B04 (red),B05,B06,B07,B08,B8A,B11,B12}
	
	\node[annot](c1) at (axis cs:25,.95){clouds}; 
	\draw[anno] (c1) -- (axis cs:38,.35);
	\draw[anno] (c1) -- (axis cs:39,1);
	\draw[anno] (c1) -- (axis cs:37,.55);
	\draw[anno] (c1) -- (axis cs:42,.5);
	
	\node[annotg](c1) at (axis cs:50,.95){ground};
	\draw[annog] (c1) -- (axis cs:41,.2);
	
	%
	\nextgroupplot[draw opacity=.8, smooth=0.01, title=Transformer]
	\addplot[b11color, mymark] table [x=t, y=B11, col sep=comma, forget plot] {images/backprop/transformer/\meta/dydx.csv};
	\addplot[b12color, mymark] table [x=t, y=B12, col sep=comma] {images/backprop/transformer/\meta/dydx.csv};
	
	\addplot[b5color, mymark] table [x=t, y=B5, col sep=comma, forget plot] {images/backprop/transformer/\meta/dydx.csv};
	\addplot[b6color, mymark] table [x=t, y=B6, col sep=comma, forget plot] {images/backprop/transformer/\meta/dydx.csv};
	\addplot[b7color, mymark] table [x=t, y=B7, col sep=comma, forget plot] {images/backprop/transformer/\meta/dydx.csv};
	\addplot[b8color, mymark] table [x=t, y=B8, col sep=comma, forget plot] {images/backprop/transformer/\meta/dydx.csv};
	\addplot[b8Acolor, mymark] table [x=t, y=B8A, col sep=comma] {images/backprop/transformer/\meta/dydx.csv};
	
	\addplot[b2color, mymark] table [x=t, y=B2, col sep=comma, forget plot] {images/backprop/transformer/\meta/dydx.csv};
	\addplot[b3color, mymark] table [x=t, y=B3, col sep=comma, forget plot] {images/backprop/transformer/\meta/dydx.csv};
	\addplot[b4color, mymark] table [x=t, y=B4, col sep=comma] {images/backprop/transformer/\meta/dydx.csv};
	
	\node[annot](c1) at (axis cs:30,.4){no influence}; 
	\draw[anno] (c1) -- (axis cs:38,0);
	\draw[anno] (c1) -- (axis cs:42,0);
	
	\node[annotg](c1) at (axis cs:50,.4){influence};
	\draw[annog] (c1) -- (axis cs:41,.4);
	
	\nextgroupplot[draw opacity=.8, smooth=0.01, title=RNN (LSTM)]
	\addplot[b11color, mymark] table [x=t, y=B11, col sep=comma, forget plot] {images/backprop/rnn/\meta/dydx.csv};
	\addplot[b12color, mymark] table [x=t, y=B12, col sep=comma] {images/backprop/rnn/\meta/dydx.csv};
	
	\addplot[b5color, mymark] table [x=t, y=B5, col sep=comma, forget plot] {images/backprop/rnn/\meta/dydx.csv};
	\addplot[b6color, mymark] table [x=t, y=B6, col sep=comma, forget plot] {images/backprop/rnn/\meta/dydx.csv};
	\addplot[b7color, mymark] table [x=t, y=B7, col sep=comma, forget plot] {images/backprop/rnn/\meta/dydx.csv};
	\addplot[b8color, mymark] table [x=t, y=B8, col sep=comma, forget plot] {images/backprop/rnn/\meta/dydx.csv};
	\addplot[b8Acolor, mymark] table [x=t, y=B8A, col sep=comma] {images/backprop/rnn/\meta/dydx.csv};
	
	\addplot[b2color, mymark] table [x=t, y=B2, col sep=comma, forget plot] {images/backprop/rnn/\meta/dydx.csv};
	\addplot[b3color, mymark] table [x=t, y=B3, col sep=comma, forget plot] {images/backprop/rnn/\meta/dydx.csv};
	\addplot[b4color, mymark] table [x=t, y=B4, col sep=comma] {images/backprop/rnn/\meta/dydx.csv};
	
	\node[annot](c1) at (axis cs:30,.1){no influence}; 
	\draw[anno] (c1) -- (axis cs:38,0);
	\draw[anno] (c1) -- (axis cs:42,0);
	
	\node[annotg](c1) at (axis cs:50,.1){influence};
	\draw[annog] (c1) -- (axis cs:41,.05);
	
	\nextgroupplot[draw opacity=.8, smooth=0.01,title=MS-ResNet]
	\addplot[b11color, mymark] table [x=t, y=B11, col sep=comma, forget plot] {images/backprop/msresnet/\meta/dydx.csv};
	\addplot[b12color, mymark] table [x=t, y=B12, col sep=comma] {images/backprop/msresnet/\meta/dydx.csv};
	
	\addplot[b5color, mymark] table [x=t, y=B5, col sep=comma, forget plot] {images/backprop/msresnet/\meta/dydx.csv};
	\addplot[b6color, mymark] table [x=t, y=B6, col sep=comma, forget plot] {images/backprop/msresnet/\meta/dydx.csv};
	\addplot[b7color, mymark] table [x=t, y=B7, col sep=comma, forget plot] {images/backprop/msresnet/\meta/dydx.csv};
	\addplot[b8color, mymark] table [x=t, y=B8, col sep=comma, forget plot] {images/backprop/msresnet/\meta/dydx.csv};
	\addplot[b8Acolor, mymark] table [x=t, y=B8A, col sep=comma] {images/backprop/msresnet/\meta/dydx.csv};
	
	\addplot[b2color, mymark] table [x=t, y=B2, col sep=comma, forget plot] {images/backprop/msresnet/\meta/dydx.csv};
	\addplot[b3color, mymark] table [x=t, y=B3, col sep=comma, forget plot] {images/backprop/msresnet/\meta/dydx.csv};
	\addplot[b4color, mymark] table [x=t, y=B4, col sep=comma] {images/backprop/msresnet/\meta/dydx.csv};
	
	\node[annot](c1) at (axis cs:28,1){no influence}; 
	\draw[anno] (c1) -- (axis cs:38,0);
	
	\node[annot](c1) at (axis cs:40,1.1){some influence}; 
	\draw[anno] (c1) to[bend left, looseness=.8] (axis cs:42,0);
	
	\node[annotg](c1) at (axis cs:60,1){influence};
	\draw[annog] (c1) -- (axis cs:41,.3);
	
	\nextgroupplot[draw opacity=.8, smooth=0.01,title=TempCNN, ymin=-1.5, ymax=1.5]
	\addplot[b11color, mymark] table [x=t, y=B11, col sep=comma, forget plot] {images/backprop/tempcnn/\meta/dydx.csv};
	\addplot[b12color, mymark] table [x=t, y=B12, col sep=comma] {images/backprop/tempcnn/\meta/dydx.csv};
	
	\addplot[b5color, mymark] table [x=t, y=B5, col sep=comma, forget plot] {images/backprop/tempcnn/\meta/dydx.csv};
	\addplot[b6color, mymark] table [x=t, y=B6, col sep=comma, forget plot] {images/backprop/tempcnn/\meta/dydx.csv};
	\addplot[b7color, mymark] table [x=t, y=B7, col sep=comma, forget plot] {images/backprop/tempcnn/\meta/dydx.csv};
	\addplot[b8color, mymark] table [x=t, y=B8, col sep=comma, forget plot] {images/backprop/tempcnn/\meta/dydx.csv};
	\addplot[b8Acolor, mymark] table [x=t, y=B8A, col sep=comma] {images/backprop/tempcnn/\meta/dydx.csv};
	
	\addplot[b2color, mymark] table [x=t, y=B2, col sep=comma, forget plot] {images/backprop/tempcnn/\meta/dydx.csv};
	\addplot[b3color, mymark] table [x=t, y=B3, col sep=comma, forget plot] {images/backprop/tempcnn/\meta/dydx.csv};
	\addplot[b4color, mymark] table [x=t, y=B4, col sep=comma] {images/backprop/tempcnn/\meta/dydx.csv};
	
	\node[annot](c1) at (axis cs:28,1.3){still influence}; 
	\draw[anno] (c1) -- (axis cs:38,0);
	\draw[anno] (c1) -- (axis cs:42,0);
	
	\node[annotg](c1) at (axis cs:40,1.3){influence};
	\draw[annog] (c1) -- (axis cs:41,.3);
	
	\end{groupplot}
	\end{tikzpicture}
	
}

\newcommand{\backpropexamplesummerbarley}{
	
	\tikzsetnextfilename{rnn_gradients_71459842}
	\begin{tikzpicture}
	
	\def\meta{71459842}
	
	\tikzstyle{mymark}=[thin,mark=*,mark size=.5pt]
	
	\begin{groupplot}[
	group style = {
		group size = 1 by 5,
		xlabels at=edge bottom,
		xticklabels at=edge bottom,
		vertical sep=0pt,
	},
	width=.5\textwidth,
	enlarge x limits=0.0005,
	enlarge y limits=0.05,
	height=3cm,
	xlabel=observation time $t$,
	legend style={at={(1,1.8)},thick, line width=2pt, draw opacity=1, font=\tiny\sffamily},
	legend image post style={line width =1pt},
	legend columns=4,
	label style={font=\scriptsize},
	tick label style={font=\scriptsize},
	y label style={%
    at={(-0.15,0.5)}, 
    rotate=-90,
    anchor=center,
  },
	ylabel=$\frac{\partial y^\ast}{\partial \M{X}}$,
	title style={
		at={(0,-.2)},
		font=\sffamily\scriptsize,
		anchor=south west,
		fill=white,
		fill opacity=0.3,
		text opacity=1,
		rounded corners,
	},
	]
	\nextgroupplot[draw opacity=.8, smooth=0.01, ylabel=reflectance $\M{X}$, y label style={rotate=90, anchor=center}, title={Sentinel 2 Time Series}, ymin=-.15,ymax=1.1
	]
	
	\addplot[b2color, mymark] table [x=t, y=B2, col sep=comma] {images/backprop/rnn/\meta/x.csv};
	\addplot[b3color, mymark] table [x=t, y=B3, col sep=comma] {images/backprop/rnn/\meta/x.csv};
	\addplot[b4color, mymark] table [x=t, y=B4, col sep=comma] {images/backprop/rnn/\meta/x.csv};
	
	\addplot[b5color, mymark] table [x=t, y=B5, col sep=comma] {images/backprop/rnn/\meta/x.csv};
	\addplot[b6color, mymark] table [x=t, y=B6, col sep=comma] {images/backprop/rnn/\meta/x.csv};
	\addplot[b7color, mymark] table [x=t, y=B7, col sep=comma] {images/backprop/rnn/\meta/x.csv};
	\addplot[b8color, mymark] table [x=t, y=B8, col sep=comma] {images/backprop/rnn/\meta/x.csv};
	\addplot[b8Acolor, mymark] table [x=t, y=B8A, col sep=comma] {images/backprop/rnn/\meta/x.csv};
	
	\addplot[b11color, mymark] table [x=t, y=B11, col sep=comma] {images/backprop/rnn/\meta/x.csv};
	\addplot[b12color, mymark] table [x=t, y=B12, col sep=comma] {images/backprop/rnn/\meta/x.csv};
	
	\legend{B02 (blue),B03 (green),B04 (red),B05,B06,B07,B08,B8A,B11,B12}
	
	\node[annot](c1) at (axis cs:15,.95){clouds}; 
	\draw[anno] (c1) -- (axis cs:30,.5);
	\draw[anno] (c1) -- (axis cs:29,1);
	\draw[anno] (c1) -- (axis cs:31,.8);
	\draw[anno] (c1) -- (axis cs:33,.55);
	
	\node[annotg](c1) at (axis cs:45,.95){ground};
	\draw[annog] (c1) -- (axis cs:32,.3);
	\draw[annog] (c1) -- (axis cs:36,.38);

	%
	\nextgroupplot[draw opacity=.8, smooth=0.01, title=Transformer]
	\addplot[b11color, mymark] table [x=t, y=B11, col sep=comma, forget plot] {images/backprop/transformer/\meta/dydx.csv};
	\addplot[b12color, mymark] table [x=t, y=B12, col sep=comma] {images/backprop/transformer/\meta/dydx.csv};
	
	\addplot[b5color, mymark] table [x=t, y=B5, col sep=comma, forget plot] {images/backprop/transformer/\meta/dydx.csv};
	\addplot[b6color, mymark] table [x=t, y=B6, col sep=comma, forget plot] {images/backprop/transformer/\meta/dydx.csv};
	\addplot[b7color, mymark] table [x=t, y=B7, col sep=comma, forget plot] {images/backprop/transformer/\meta/dydx.csv};
	\addplot[b8color, mymark] table [x=t, y=B8, col sep=comma, forget plot] {images/backprop/transformer/\meta/dydx.csv};
	\addplot[b8Acolor, mymark] table [x=t, y=B8A, col sep=comma] {images/backprop/transformer/\meta/dydx.csv};
	
	\addplot[b2color, mymark] table [x=t, y=B2, col sep=comma, forget plot] {images/backprop/transformer/\meta/dydx.csv};
	\addplot[b3color, mymark] table [x=t, y=B3, col sep=comma, forget plot] {images/backprop/transformer/\meta/dydx.csv};
	\addplot[b4color, mymark] table [x=t, y=B4, col sep=comma] {images/backprop/transformer/\meta/dydx.csv};

	\node[annot](c1) at (axis cs:15,4){no influence}; 
	\draw[anno] (c1) -- (axis cs:30,0);
	\draw[anno] (c1) -- (axis cs:33,0);
	
	\node[annotg](c1) at (axis cs:45,4){influence};
	\draw[annog] (c1) -- (axis cs:32,4);
	\draw[annog] (c1) -- (axis cs:36,2.5);
	\nextgroupplot[draw opacity=.8, smooth=0.01, title=RNN (LSTM)]
	\addplot[b11color, mymark] table [x=t, y=B11, col sep=comma, forget plot] {images/backprop/rnn/\meta/dydx.csv};
	\addplot[b12color, mymark] table [x=t, y=B12, col sep=comma] {images/backprop/rnn/\meta/dydx.csv};
	
	\addplot[b5color, mymark] table [x=t, y=B5, col sep=comma, forget plot] {images/backprop/rnn/\meta/dydx.csv};
	\addplot[b6color, mymark] table [x=t, y=B6, col sep=comma, forget plot] {images/backprop/rnn/\meta/dydx.csv};
	\addplot[b7color, mymark] table [x=t, y=B7, col sep=comma, forget plot] {images/backprop/rnn/\meta/dydx.csv};
	\addplot[b8color, mymark] table [x=t, y=B8, col sep=comma, forget plot] {images/backprop/rnn/\meta/dydx.csv};
	\addplot[b8Acolor, mymark] table [x=t, y=B8A, col sep=comma] {images/backprop/rnn/\meta/dydx.csv};
	
	\addplot[b2color, mymark] table [x=t, y=B2, col sep=comma, forget plot] {images/backprop/rnn/\meta/dydx.csv};
	\addplot[b3color, mymark] table [x=t, y=B3, col sep=comma, forget plot] {images/backprop/rnn/\meta/dydx.csv};
	\addplot[b4color, mymark] table [x=t, y=B4, col sep=comma] {images/backprop/rnn/\meta/dydx.csv};

	\node[annot](c1) at (axis cs:15,28){no influence}; 
	\draw[anno] (c1) -- (axis cs:30,0);
	\draw[anno] (c1) -- (axis cs:33,0);
	
	\node[annotg](c1) at (axis cs:45,28){influence};
	\draw[annog] (c1) -- (axis cs:32,28);
	\draw[annog] (c1) -- (axis cs:36,10);

	\nextgroupplot[draw opacity=.8, smooth=0.01,title=MS-ResNet]
	\addplot[b11color, mymark] table [x=t, y=B11, col sep=comma, forget plot] {images/backprop/msresnet/\meta/dydx.csv};
	\addplot[b12color, mymark] table [x=t, y=B12, col sep=comma] {images/backprop/msresnet/\meta/dydx.csv};
	
	\addplot[b5color, mymark] table [x=t, y=B5, col sep=comma, forget plot] {images/backprop/msresnet/\meta/dydx.csv};
	\addplot[b6color, mymark] table [x=t, y=B6, col sep=comma, forget plot] {images/backprop/msresnet/\meta/dydx.csv};
	\addplot[b7color, mymark] table [x=t, y=B7, col sep=comma, forget plot] {images/backprop/msresnet/\meta/dydx.csv};
	\addplot[b8color, mymark] table [x=t, y=B8, col sep=comma, forget plot] {images/backprop/msresnet/\meta/dydx.csv};
	\addplot[b8Acolor, mymark] table [x=t, y=B8A, col sep=comma] {images/backprop/msresnet/\meta/dydx.csv};
	
	\addplot[b2color, mymark] table [x=t, y=B2, col sep=comma, forget plot] {images/backprop/msresnet/\meta/dydx.csv};
	\addplot[b3color, mymark] table [x=t, y=B3, col sep=comma, forget plot] {images/backprop/msresnet/\meta/dydx.csv};
	\addplot[b4color, mymark] table [x=t, y=B4, col sep=comma] {images/backprop/msresnet/\meta/dydx.csv};
	
	\node[annot](c1) at (axis cs:15,.4){no influence}; 
	\draw[anno] (c1) -- (axis cs:30,0);
	\node[annot](c1) at (axis cs:25,.4){some infl.};
	\draw[anno] (c1) -- (axis cs:33,.1);
	
	\node[annotg](c1) at (axis cs:47,.4){influence};
	\draw[annog] (c1) -- (axis cs:32,.3);
	\draw[annog] (c1) -- (axis cs:36,.2);
	
	\nextgroupplot[draw opacity=.8, smooth=0.01,title=TempCNN]
	\addplot[b11color, mymark] table [x=t, y=B11, col sep=comma, forget plot] {images/backprop/tempcnn/\meta/dydx.csv};
	\addplot[b12color, mymark] table [x=t, y=B12, col sep=comma] {images/backprop/tempcnn/\meta/dydx.csv};
	
	\addplot[b5color, mymark] table [x=t, y=B5, col sep=comma, forget plot] {images/backprop/tempcnn/\meta/dydx.csv};
	\addplot[b6color, mymark] table [x=t, y=B6, col sep=comma, forget plot] {images/backprop/tempcnn/\meta/dydx.csv};
	\addplot[b7color, mymark] table [x=t, y=B7, col sep=comma, forget plot] {images/backprop/tempcnn/\meta/dydx.csv};
	\addplot[b8color, mymark] table [x=t, y=B8, col sep=comma, forget plot] {images/backprop/tempcnn/\meta/dydx.csv};
	\addplot[b8Acolor, mymark] table [x=t, y=B8A, col sep=comma] {images/backprop/tempcnn/\meta/dydx.csv};
	
	\addplot[b2color, mymark] table [x=t, y=B2, col sep=comma, forget plot] {images/backprop/tempcnn/\meta/dydx.csv};
	\addplot[b3color, mymark] table [x=t, y=B3, col sep=comma, forget plot] {images/backprop/tempcnn/\meta/dydx.csv};
	\addplot[b4color, mymark] table [x=t, y=B4, col sep=comma] {images/backprop/tempcnn/\meta/dydx.csv};
	
	\node[annot](c1) at (axis cs:20,4){still influence}; 
	\draw[anno] (c1) -- (axis cs:30,2);
	\draw[anno] (c1) -- (axis cs:33,2);
	
	\node[annotg](c1) at (axis cs:45,4){wanted influence};
	\draw[annog] (c1) -- (axis cs:32,3);
	\draw[annog] (c1) -- (axis cs:36,2);

	\end{groupplot}
	\end{tikzpicture}
	
}

\newcommand{\backpropexample}[1]{

\tikzsetnextfilename{rnn_gradients_#1}
\begin{tikzpicture}

\def\meta{#1}

\tikzstyle{mymark}=[thin,mark=*,mark size=.5pt]

\begin{groupplot}[
group style = {
group size = 1 by 5,
xlabels at=edge bottom,
xticklabels at=edge bottom,
vertical sep=0pt,
},
width=.5\linewidth,
enlarge x limits=0.0005,
enlarge y limits=0.05,
height=3cm,
xlabel=observation time $t$,
legend style={at={(1,1.8)},thick, line width=2pt, draw opacity=1, font=\tiny\sffamily},
legend image post style={line width =1pt},
legend columns=4,
y label style={at={(-0.1,0.5)}},
ylabel=$\frac{\partial y^\ast}{\partial \M{X}}$,
title style={
	at={(0,-.2)},
	font=\sffamily\scriptsize,
	anchor=south west,
	fill=white,
	fill opacity=0.3,
	text opacity=1,
	rounded corners,
}
]
\nextgroupplot[draw opacity=.8, smooth=0.01, ylabel=reflectance $\M{X}$, title={Sentinel 2 Time Series}, ymin=-.15,
]

\addplot[b2color, mymark] table [x=t, y=B2, col sep=comma] {images/backprop/rnn/\meta/x.csv};
\addplot[b3color, mymark] table [x=t, y=B3, col sep=comma] {images/backprop/rnn/\meta/x.csv};
\addplot[b4color, mymark] table [x=t, y=B4, col sep=comma] {images/backprop/rnn/\meta/x.csv};

\addplot[b5color, mymark] table [x=t, y=B5, col sep=comma] {images/backprop/rnn/\meta/x.csv};
\addplot[b6color, mymark] table [x=t, y=B6, col sep=comma] {images/backprop/rnn/\meta/x.csv};
\addplot[b7color, mymark] table [x=t, y=B7, col sep=comma] {images/backprop/rnn/\meta/x.csv};
\addplot[b8color, mymark] table [x=t, y=B8, col sep=comma] {images/backprop/rnn/\meta/x.csv};
\addplot[b8Acolor, mymark] table [x=t, y=B8A, col sep=comma] {images/backprop/rnn/\meta/x.csv};

\addplot[b11color, mymark] table [x=t, y=B11, col sep=comma] {images/backprop/rnn/\meta/x.csv};
\addplot[b12color, mymark] table [x=t, y=B12, col sep=comma] {images/backprop/rnn/\meta/x.csv};

\legend{B02 (blue),B03 (green),B04 (red),B05,B06,B07,B08,B8A,B11,B12}

%
\nextgroupplot[draw opacity=.8, smooth=0.01, title=Transformer]
\addplot[b11color, mymark] table [x=t, y=B11, col sep=comma, forget plot] {images/backprop/transformer/\meta/dydx.csv};
\addplot[b12color, mymark] table [x=t, y=B12, col sep=comma] {images/backprop/transformer/\meta/dydx.csv};

\addplot[b5color, mymark] table [x=t, y=B5, col sep=comma, forget plot] {images/backprop/transformer/\meta/dydx.csv};
\addplot[b6color, mymark] table [x=t, y=B6, col sep=comma, forget plot] {images/backprop/transformer/\meta/dydx.csv};
\addplot[b7color, mymark] table [x=t, y=B7, col sep=comma, forget plot] {images/backprop/transformer/\meta/dydx.csv};
\addplot[b8color, mymark] table [x=t, y=B8, col sep=comma, forget plot] {images/backprop/transformer/\meta/dydx.csv};
\addplot[b8Acolor, mymark] table [x=t, y=B8A, col sep=comma] {images/backprop/transformer/\meta/dydx.csv};

\addplot[b2color, mymark] table [x=t, y=B2, col sep=comma, forget plot] {images/backprop/transformer/\meta/dydx.csv};
\addplot[b3color, mymark] table [x=t, y=B3, col sep=comma, forget plot] {images/backprop/transformer/\meta/dydx.csv};
\addplot[b4color, mymark] table [x=t, y=B4, col sep=comma] {images/backprop/transformer/\meta/dydx.csv};

\nextgroupplot[draw opacity=.8, smooth=0.01, title=RNN (LSTM)]
\addplot[b11color, mymark] table [x=t, y=B11, col sep=comma, forget plot] {images/backprop/rnn/\meta/dydx.csv};
\addplot[b12color, mymark] table [x=t, y=B12, col sep=comma] {images/backprop/rnn/\meta/dydx.csv};

\addplot[b5color, mymark] table [x=t, y=B5, col sep=comma, forget plot] {images/backprop/rnn/\meta/dydx.csv};
\addplot[b6color, mymark] table [x=t, y=B6, col sep=comma, forget plot] {images/backprop/rnn/\meta/dydx.csv};
\addplot[b7color, mymark] table [x=t, y=B7, col sep=comma, forget plot] {images/backprop/rnn/\meta/dydx.csv};
\addplot[b8color, mymark] table [x=t, y=B8, col sep=comma, forget plot] {images/backprop/rnn/\meta/dydx.csv};
\addplot[b8Acolor, mymark] table [x=t, y=B8A, col sep=comma] {images/backprop/rnn/\meta/dydx.csv};

\addplot[b2color, mymark] table [x=t, y=B2, col sep=comma, forget plot] {images/backprop/rnn/\meta/dydx.csv};
\addplot[b3color, mymark] table [x=t, y=B3, col sep=comma, forget plot] {images/backprop/rnn/\meta/dydx.csv};
\addplot[b4color, mymark] table [x=t, y=B4, col sep=comma] {images/backprop/rnn/\meta/dydx.csv};

\nextgroupplot[draw opacity=.8, smooth=0.01,title=MS-ResNet]
\addplot[b11color, mymark] table [x=t, y=B11, col sep=comma, forget plot] {images/backprop/msresnet/\meta/dydx.csv};
\addplot[b12color, mymark] table [x=t, y=B12, col sep=comma] {images/backprop/msresnet/\meta/dydx.csv};

\addplot[b5color, mymark] table [x=t, y=B5, col sep=comma, forget plot] {images/backprop/msresnet/\meta/dydx.csv};
\addplot[b6color, mymark] table [x=t, y=B6, col sep=comma, forget plot] {images/backprop/msresnet/\meta/dydx.csv};
\addplot[b7color, mymark] table [x=t, y=B7, col sep=comma, forget plot] {images/backprop/msresnet/\meta/dydx.csv};
\addplot[b8color, mymark] table [x=t, y=B8, col sep=comma, forget plot] {images/backprop/msresnet/\meta/dydx.csv};
\addplot[b8Acolor, mymark] table [x=t, y=B8A, col sep=comma] {images/backprop/msresnet/\meta/dydx.csv};

\addplot[b2color, mymark] table [x=t, y=B2, col sep=comma, forget plot] {images/backprop/msresnet/\meta/dydx.csv};
\addplot[b3color, mymark] table [x=t, y=B3, col sep=comma, forget plot] {images/backprop/msresnet/\meta/dydx.csv};
\addplot[b4color, mymark] table [x=t, y=B4, col sep=comma] {images/backprop/msresnet/\meta/dydx.csv};

\nextgroupplot[draw opacity=.8, smooth=0.01,title=TempCNN]
\addplot[b11color, mymark] table [x=t, y=B11, col sep=comma, forget plot] {images/backprop/tempcnn/\meta/dydx.csv};
\addplot[b12color, mymark] table [x=t, y=B12, col sep=comma] {images/backprop/tempcnn/\meta/dydx.csv};

\addplot[b5color, mymark] table [x=t, y=B5, col sep=comma, forget plot] {images/backprop/tempcnn/\meta/dydx.csv};
\addplot[b6color, mymark] table [x=t, y=B6, col sep=comma, forget plot] {images/backprop/tempcnn/\meta/dydx.csv};
\addplot[b7color, mymark] table [x=t, y=B7, col sep=comma, forget plot] {images/backprop/tempcnn/\meta/dydx.csv};
\addplot[b8color, mymark] table [x=t, y=B8, col sep=comma, forget plot] {images/backprop/tempcnn/\meta/dydx.csv};
\addplot[b8Acolor, mymark] table [x=t, y=B8A, col sep=comma] {images/backprop/tempcnn/\meta/dydx.csv};

\addplot[b2color, mymark] table [x=t, y=B2, col sep=comma, forget plot] {images/backprop/tempcnn/\meta/dydx.csv};
\addplot[b3color, mymark] table [x=t, y=B3, col sep=comma, forget plot] {images/backprop/tempcnn/\meta/dydx.csv};
\addplot[b4color, mymark] table [x=t, y=B4, col sep=comma] {images/backprop/tempcnn/\meta/dydx.csv};
%

\end{groupplot}
\end{tikzpicture}

}

%% file: images/embedding.tex
	\begin{minipage}{.83\linewidth}
	
			\begin{tikzpicture}[node distance=0em, inner sep=0]
		\node(umap0){\includegraphics[width=.33\textwidth]{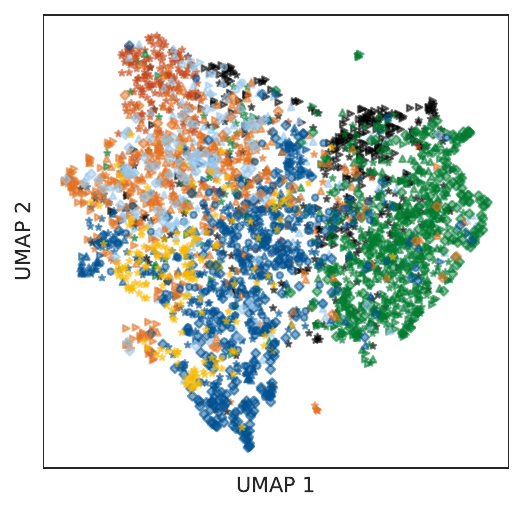}};
		\node[right=of umap0](umap1){\includegraphics[width=.33\textwidth]{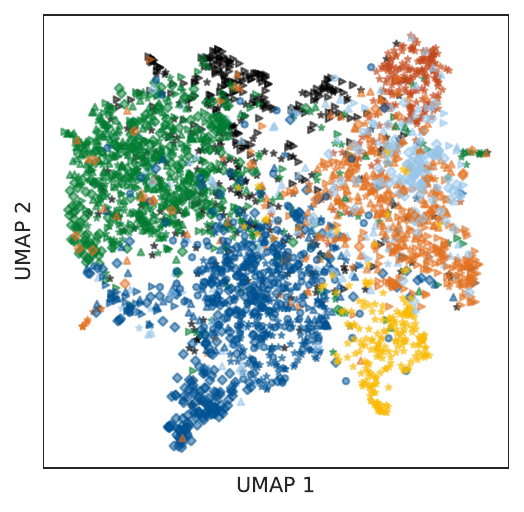}};
		\node[right=of umap1](umap2){\includegraphics[width=.33\textwidth]{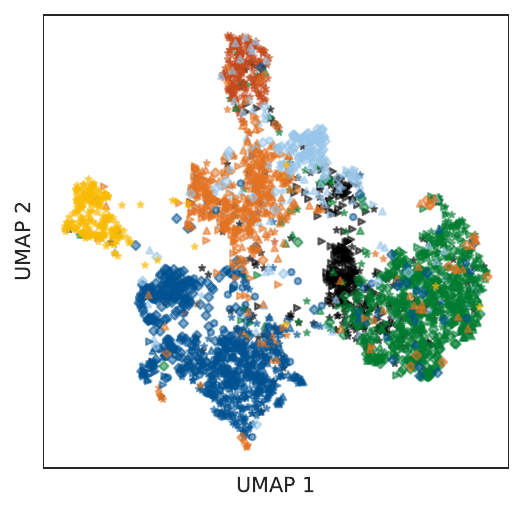}};
		
		\node[below=of umap0](tsne0){\includegraphics[width=.33\textwidth]{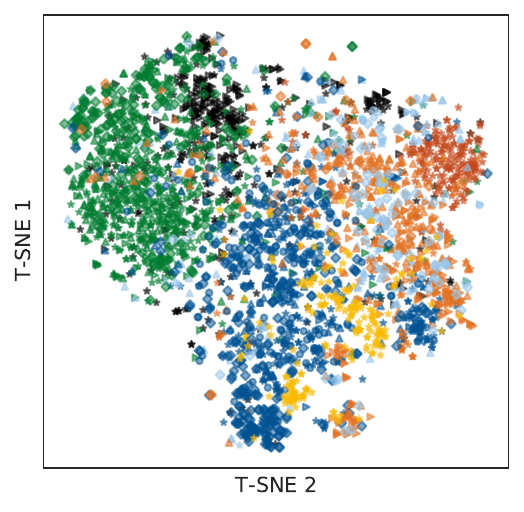}};
		\node[right=of tsne0](tsne1){\includegraphics[width=.33\textwidth]{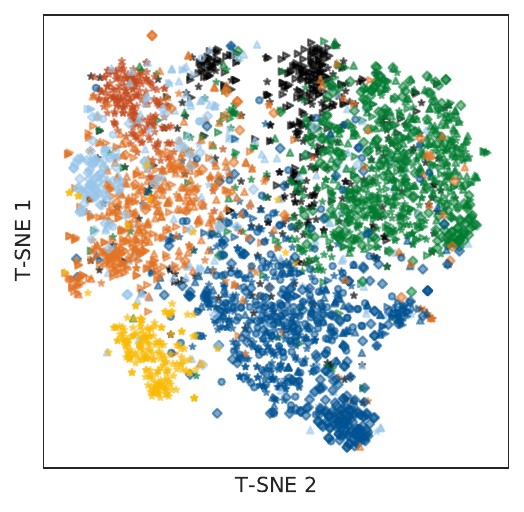}};
		\node[right=of tsne1](tsne2){\includegraphics[width=.33\textwidth]{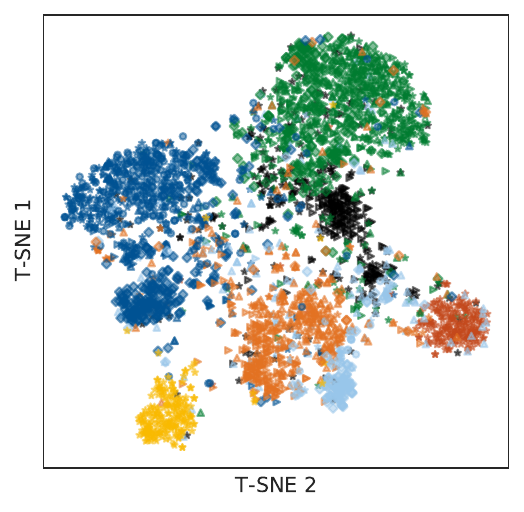}};
%
		\node[below=of tsne0](pca0){\includegraphics[width=.33\textwidth]{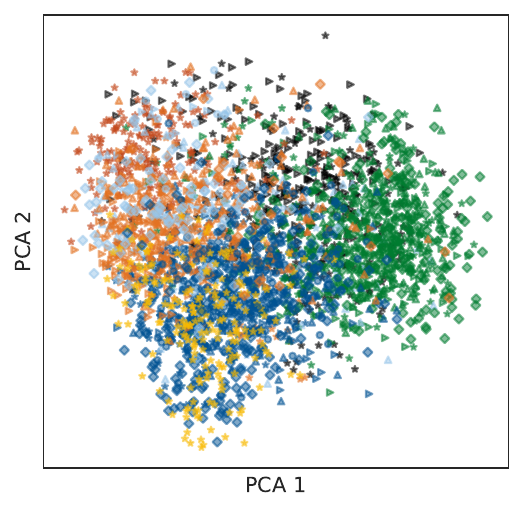}};
		\node[right=of pca0](pca1){\includegraphics[width=.33\textwidth]{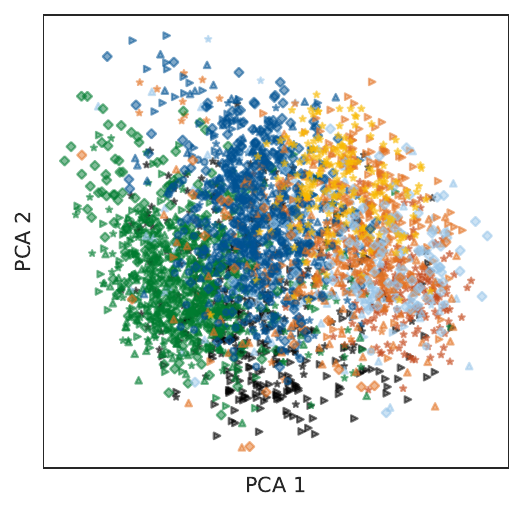}};
		\node[right=of pca1](pca2){\includegraphics[width=.33\textwidth]{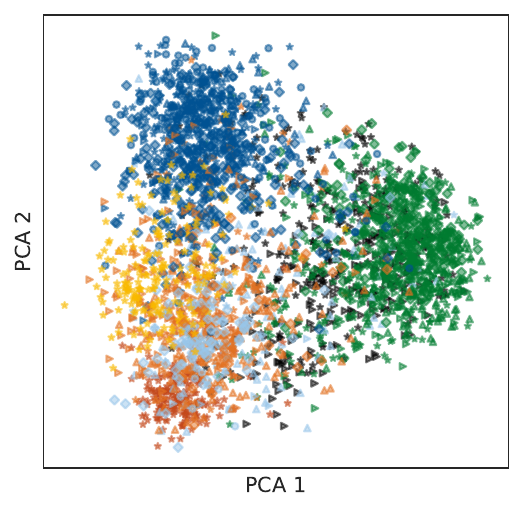}};
		
		\node[below=of pca0]{(a)};
		\node[below=of pca1]{(b)};
		\node[below=of pca2]{(c)};
		
		\node[anchor=north west, font=\sffamily\scriptsize, xshift=1.1em, yshift=-.5em] at (umap0.north west){UMAP};
		\node[anchor=north west, font=\sffamily\scriptsize, xshift=1.1em, yshift=-.5em] at (tsne0.north west){T-SNE};
		\node[anchor=north west, font=\sffamily\scriptsize, xshift=1.1em, yshift=-.5em] at (pca0.north west){PCA};
		
		\node[anchor=north west, font=\sffamily\scriptsize, xshift=1.1em, yshift=-.5em] at (umap1.north west){UMAP};
		\node[anchor=north west, font=\sffamily\scriptsize, xshift=1.1em, yshift=-.5em] at (tsne1.north west){T-SNE};
		\node[anchor=north west, font=\sffamily\scriptsize, xshift=1.1em, yshift=-.5em] at (pca1.north west){PCA};
		
		\node[anchor=north west, font=\sffamily\scriptsize, xshift=1.1em, yshift=-.5em] at (umap2.north west){UMAP};
		\node[anchor=north west, font=\sffamily\scriptsize, xshift=1.1em, yshift=-.5em] at (tsne2.north west){T-SNE};
		\node[anchor=north west, font=\sffamily\scriptsize, xshift=1.1em, yshift=-.5em] at (pca2.north west){PCA};
		
		\end{tikzpicture}
	\end{minipage}
	\begin{minipage}{.15\linewidth}
		\includegraphics[width=\linewidth]{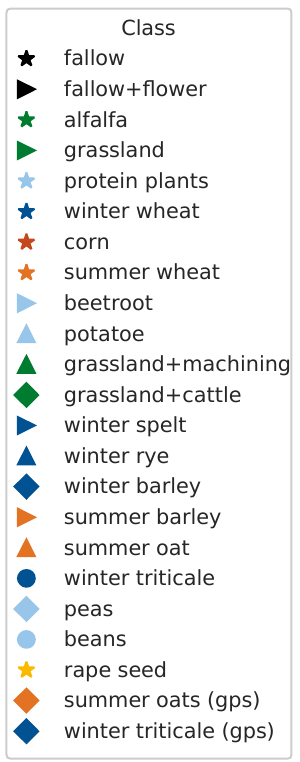}
		\input{images/models/tsnetransformer.tikz}
	\end{minipage}